\documentclass[conference]{IEEEtran}
\usepackage{amsmath,amsfonts}
\usepackage{algorithmic}
\usepackage{algorithm}
\usepackage{array}
\usepackage[caption=false,font=normalsize,labelfont=sf,textfont=sf]{subfig}
\usepackage{textcomp}
\usepackage{stfloats}
\usepackage{url}
\usepackage{verbatim}
\usepackage{graphicx}
\usepackage[symbols,toc,automake,acronym]{glossaries-extra}
\usepackage{cite}
\usepackage{tikz}
\usepackage{pgfplots}
\usepackage{hyperref}
\usepackage{xcolor}
\usepackage{mdframed} 
\usetikzlibrary{calc}

\usepackage{tikz}
\usepackage{tcolorbox}
\usepackage{lipsum} 
\usepackage{marginnote}
\newtheorem{theorem}{Theorem}
\newtheorem{definition}[theorem]{Definition}

\usepackage{tikz}
\usetikzlibrary{positioning, arrows.meta, fit}
\usetikzlibrary{shapes.geometric}

\newcommand\hypothesis{h} 
\newcommand{\algomap}{\mathcal{A}}

\newcommand\localmodel[1]{\hypospace^{(#1)}} 
\newcommand\localhypothesis[1]{\hypothesis^{(#1)}}

\newcommand\defeq{:=}

\newcommand{\vx}[0]{{\bf x}}
\newcommand{\vv}[0]{{\bf v}}
\newcommand{\vu}[0]{{\bf u}}

\newcommand{\vm}[0]{{\bf m}}
\newcommand{\vq}[0]{{\bf q}}
\newcommand{\mX}[0]{{\bf X}}
\newcommand{\mC}[0]{{\bf C}}
\newcommand{\mA}[0]{{\bf A}}
\newcommand{\mL}[0]{{\bf L}}

\newcommand{\vw}[0]{{\bf w}}

\newcommand{\mQ}{\mathbf{Q}}
\newcommand{\mU}{\mathbf{U}}
\newcommand{\mV}{\mathbf{V}}

\newcommand{\vy}[0]{{\bf y}}
\newcommand{\va}[0]{{\bf a}}

\newcommand{\vz}[0]{{\bf z}}

\newcommand{\vg}[0]{{\bf g}}

\newcommand{\prob}[1]{p({#1})} 
 
\newcommand{\meanvec}[1]{{\bm \mu}^{(#1)}} 
\newcommand{\covmtx}[1]{\mathbf{C}^{(#1)}}

\def \expect {\mathbb{E} }

\newcommand{\biasterm}{B}

\newcommand{\neighbourhood}[1]{\mathcal{N}^{(#1)}}
\newcommand{\nrfolds}{k}

\newcommand{\nrcategories}{K}

\newcommand{\privattr}{s}
\newcommand{\sensattr}{s}

\newcommand{\normgeneric}[2]{\left\Vert  {#1} \right\Vert_{#2}}

\newcommand{\bmx}[0]{\begin{bmatrix}}
\newcommand{\emx}[0]{\end{bmatrix}}

\newcommand{\featuredim}{d}
\newcommand{\nrfeatures}{\featuredim}

\newcommand{\featurelen}{\featuredim}

\newcommand{\samplesize}{m}
\newcommand{\sampleidx}{r} 
 
\newcommand{\datapoint}{\vz} 
 
\newcommand{\clusteridx}{c} 

\newcommand{\nrcluster}{k} 
 
\newcommand{\featureidx}{j} 
\newcommand{\clustermean}{{\bm \mu}} 
\newcommand{\clustercov}{{\bm \Sigma}}

\newcommand\truelabel{y}

\newcommand\labelvec{\vy}
\newcommand\featurevec{\vx}

\newcommand\feature{x}
\newcommand\predictedlabel{\hat{\truelabel}}
\newcommand\dataset{\mathcal{D}}

\newcommand\effdim[1]{d_{\rm eff} \left( #1 \right)}

\newcommand{\learnthypothesis}{\hat{\hypothesis}}

\newcommand{\hypospace}{\mathcal{H}}

\newcommand{\emperror}{\widehat{L}}
\newcommand\emprisk[2]{\widehat{L}\big(#1|#2\big)}
\newcommand\risk[1]{\bar{L} \big( #1 \big) } 
\newcommand{\featurespace}{\mathcal{X}}
\newcommand{\labelspace}{\mathcal{Y}}

\newcommand{\explanation}{e}

\newcommand{\user}{u}

\newcommand{\eigval}[1]{\lambda_{#1}}
\newcommand{\eigvalgen}{\lambda}
\newcommand{\regparam}{\alpha}
\newcommand{\lrate}{\eta}

\newcommand{\perturbedloss}[2]{\widetilde{L}\left({#1},{#2}\right)}

\DeclareMathOperator*{\argmax}{argmax}
\newcommand{\uncertset}{\mathcal{U}}
\DeclareMathOperator*{\argmin}{argmin}

\newcommand{\testset}{\dataset^{(\rm test)} }

\newcommand{\loss}{L}
\newcommand{\lossfun}{L}
\newcommand{\lossfunc}[2]{L\left(#1,#2 \right)}

\newcommand{\featuremtx}{\mX}
\newcommand{\weight}{w}
\newcommand{\weights}{\vw}
\newcommand{\regularizer}[1]{\mathcal{R}\big\{ #1 \big\}}
\newcommand{\decreg}[1]{\mathcal{R}_{#1}}

\newcommand{\featuremapvec}{{\bf \Phi}}

\newcommand{\proximityop}[3]{{\rm\bf prox}_{#1,#3}(#2)}

\newcommand{\locallossfunc}[2]{L_{#1}\left(#2 \right)}

\newcommand{\localdataset}[1]{\mathcal{D}^{(#1)}}

\newcommand{\edges}{\mathcal{E}}

\newcommand{\edgeweight}{A}

\newcommand{\edgeidx}{e}

\newcommand{\graph}{\mathcal{G}}
\newcommand{\nodes}{\mathcal{V}}

\newcommand{\nodedegree}[1]{d^{(#1)}}

\newcommand{\nodeidx}{i}

\newcommand{\edge}[2]{\{#1,#2\}}

\newcommand{\mvnormal}[2]{\mathcal{N}\left(#1,#2\right)}

\newcommand{\dimlocalmodel}{d}
\newcommand{\pair}[2]{\left( #1,#2 \right)}

\newcommand{\mutualinformation}[2]{I \left( #1;#2\right)}

\newcommand{\linmodel}[1]{\hypospace^{(#1)}}

\newglossaryentry{minimum}
{
	name=minimum,
	description={Given a set of real numbers, the minimum\index{minimum} is the smallest of those numbers.},
	first={minimum},text={minimum}
}

\newglossaryentry{maximum}
{name=maximum,
 description={Given a set of real numbers, the maximum\index{maximum} is the largest of those numbers.},
 first={maximum},text={maximum}
}

\newglossaryentry{discrepancy}
{name=discrepancy,
	description={
		Consider\index{discrepancy} a \gls{fl} application with \gls{netdata} represented by a \gls{empgraph}. 
		\gls{fl} methods use a discrepancy measure to compare \gls{hypothesis} maps from 
	local models at nodes $\nodeidx,\nodeidx'$ connected by an edge in the \gls{empgraph}.},
	first={discrepancy},text={discrepancy}
}

\newglossaryentry{hfl}
{name={horizontal \gls{fl}},description=
	{Horizontal \gls{fl}\index{horizontal FL} uses \gls{localdataset}s that are 
		constituted by different \gls{datapoint}s but using the same \gls{feature}s 
		to characterize them \cite{HFLChapter2020}. For example, weather 
		forecasting uses a network of spatially distributed weather (observation) 
		stations. Each weather station measures the same quantities such as daily 
		temperature, air pressure and precipitation. However, different weather 
		stations measure the characteristics or \gls{feature}s of different spatio-temporal regions 
		(each such region being a separate \gls{datapoint}).},
	first={horizontal \gls{fl}},text={horizontal \gls{fl}}
} 

\newglossaryentry{dimred}
{name={dimensionality reduction},
	description={Dimensionality reduction\index{dimensionality reduction} methods 
		map (typically many) raw \gls{feature}s to a (relatively small) set of new features. 
		These methods can be used to visualize \gls{datapoint}s by learning two \gls{feature}s 
		that can be used as the coordinates of a depiction in a \gls{scatterplot}.}, first={dimensionality reduction},text={dimensionality reduction}
} 

\newglossaryentry{featlearn}
{name={feature learning},
	description={Feature learning\index{feature learning} refers to the task of learning a map $\featuremapvec$ that 
		reads in raw \gls{feature}s of a \gls{datapoint} and delivers new \gls{feature}s. Different 
		\gls{feature} learning methods are obtained for different quantitative measures for the 
		usefulness of the new \gls{feature}s.},
	first={feature learning},text={feature learning}
} 

\newglossaryentry{autoencoder}
{name={autoencoder},
	description={An autoencoder\index{autoencoder} is a ML method that jointly learns an encoder map 
		$\hypothesis(\cdot) \in \hypospace$ and a decoder map $\hypothesis^{*}(\cdot) \in \hypospace^{*}$. 
		It is an instance of \gls{erm} using a \gls{loss} computed from the reconstruction error 
		$\featurevec - \hypothesis^{*}\big(  \hypothesis \big( \featurevec \big) \big)$.},
	first={autoencoder},text={autoencoder}
} 

\newglossaryentry{vfl}
{name={vertical \gls{fl}},description=
	{Vertical \gls{fl}\index{vertical FL} uses \gls{localdataset}s that are constituted 
	 by the same \gls{datapoint}s but characterizing them with different \gls{feature}s \cite{VFLChapter}. 
     For example, different healthcare providers might all contain information 
     about the same population of patients. However, different healthcare providers 
     collect different measurements (blood values, electrocardiography, lung X-ray) 
     for the same patients.},
	first={vertical \gls{fl}},text={vertical \gls{fl}}
} 

\newglossaryentry{interpretability}
{name={interpretability},description=
		{A ML method is interpretable\index{interpretability} for a specific user if 
			they can well anticipate the \gls{prediction}s delivered by the method. 
			The notion of interpretability can be made precise using quantitative 
			measures of the uncertainty about the \gls{prediction}s \cite{JunXML2020}.},
		first={interpretability},text={interpretability}
}

\newglossaryentry{multitask learning}
{name={multitask learning},description=
	{Multitask learning\index{multitask learning} aims at leveraging relations between 
	 different \gls{learningtask}s. Consider two \gls{learningtask}s obtained from the 
	 same \gls{dataset} of webcam snapshots. The first task is to predict the presence 
	 of a human, while the second is the predict the presence of a car. It might be useful 
	 to use the same \gls{deepnet} structure for both tasks and only allow the weights of 
	 the final output layer to be different.},
	first={multitask learning},text={multitask learning}
}

\newglossaryentry{learningtask}
{name={learning task},description=
	{Consider\index{learning task} a \gls{dataset} $\dataset$ constituted by several \gls{datapoint}s, each of them 
	 characterized by \gls{feature}s $\featurevec$. For example, the \gls{dataset} $\dataset$ 
	 might be constituted by the images of a particular database. Sometimes it might be useful 
	 to represent a \gls{dataset} $\dataset$, along with the choice of \gls{feature}s, by a \gls{probdist} $p(\featurevec)$. 
	 A learning task associated with $\dataset$ consists of a specific 
	 choice for the \gls{label} of a \gls{datapoint} and the corresponding \gls{labelspace}. 
	 Given a choice for the \gls{lossfunc} and \gls{model}, a learning task gives rise to an 
	 instance of \gls{erm}. Thus, we could define a learning task also via an instance of \gls{erm}, i.e., 
	 via an objective function.  Note that, for the same \gls{dataset}, we obtain different learning tasks by using 
	 different choices for the \gls{feature}s and \gls{label} of a \gls{datapoint}. These learning 
	 tasks are related, as they are based on the same \gls{dataset}, and solving them jointly 
	 (via multitask learning methods) is typically preferable over solving them separately \cite{Caruana:1997wk,JungGaphLassoSPL,CSGraphSelJournal}.},
	first={learning task},text={learning task}
}

\newglossaryentry{featuremtx}
{name={feature matrix},description=
	{Consider\index{feature matrix} a \gls{dataset} $\dataset$ of $\samplesize$ \gls{datapoint}s that are 
	characterized by feature vectors $\featurevec^{(1)},\ldots,\featurevec^{(\samplesize)}$. It is 
	convenient to collect the individual \gls{feature} vectors into a \gls{feature} 
	matrix $\mX \defeq \big(\featurevec^{(1)},\ldots,\featurevec^{(\samplesize)}\big)$. },
	first={feature matrix},text={feature matrix}
}

\newglossaryentry{explainability}
{name={explainability},description=
		{We\index{explainability} define the (subjective) explainability of a ML method 
			as the level of simulatability \cite{Colin:2022aa} of the \gls{prediction}s 
			delivered by a ML system to a human user. Quantitative measures for the 
			(subjective) explainability of a trained \gls{model} can be constructed by 
			comparing its \gls{prediction}s with the \gls{prediction}s provided by a user 
			on a test-set \cite{Zhang:2024aa,Colin:2022aa}. Alternatively, we can use 
			\gls{probmodel}s for \gls{data} and measure explainability of a trained ML model 
			via the conditional (differential) entropy of its \gls{prediction}s, given the user \gls{prediction}s \cite{JunXML2020,Chen2018}. 
		},
		first={explainability},text={explainability}
	}

\newglossaryentry{linmodel}{name={linear model},
	description={Consider\index{linear model}  \gls{datapoint}s, each characterized by a numeric \gls{feature} 
		vector $\featurevec \in \mathbb{R}^{\featuredim}$. A linear \gls{model} is 
		a \gls{hypospace} which consists of all linear maps, 
	\begin{equation} 
		\label{equ_def_lin_model_hypspace}
		\linmodel{\nrfeatures} \defeq \left\{ \hypothesis(\featurevec)= \weights^{T} \featurevec: \weights \in \mathbb{R}^{\nrfeatures} \right\}. 
	\end{equation} 
	Note that \eqref{equ_def_lin_model_hypspace} defines an entire family of \gls{hypospace}s, which is 
	parametrized by the number $\nrfeatures$ of \gls{feature}s that are linearly combined to form the 
	\gls{prediction} $\hypothesis(\featurevec)$. The design choice of $\nrfeatures$ is guided by \gls{compasp} 
	(smaller $\nrfeatures$ means less computation), \gls{statasp} (increasing $\nrfeatures$ might 
	reduce \gls{prediction} error) and \gls{interpretability}. A linear \gls{model} using few carefully chosen 
	\gls{feature}s tends to be considered more interpretable \cite{Ribeiro2016,rudin2019stop}.}, 
   first={linear model},text={linear model}}

\newglossaryentry{gradstep}{name={gradient step},description={Given a \gls{differentiable} 
		real-valued function $f(\weights)$ and a vector $\weights'$, the \gls{gradient} step\index{gradient step} 
		updates $\weights'$ by adding the scaled negative \gls{gradient} $\nabla f(\weights')$, $\weights' \mapsto \weights' - \lrate \nabla f(\weights')$.},first={gradient step},text={gradient step}}

\newglossaryentry{proxop}{name={proximal operator},description={Given\index{proximal operator} a \gls{convex} function 
		and a vector $\vx$, we define its proximal operator as \cite{ProximalMethods,Bauschke:2017} $$\proximityop{\locallossfunc{\nodeidx}{\cdot}}{\weights''}{2 \regparam}\defeq \argmin_{\weights \in \mathbb{R}^{\dimlocalmodel}} f(\weights) + (\rho/2) \normgeneric{\weights- \weights'}{2}^{2} \mbox{ with } \rho > 0. $$ 
		\Gls{convex} functions for which the proximal operator can be computed efficiently 
		are sometimes referred to as \emph{proximable} or \emph{simple} \cite{Condat2013}.},first={proximal operator},text={proximal operator}}

\newglossaryentry{proximable}{name={proximable},description={A\index{proximable} 
		\gls{convex} function for which the \gls{proxop} can be computed efficiently are 
		sometimes referred to as \emph{proximable} or \emph{simple} \cite{Condat2013}.},first={proximable},text={proximable}}

\newglossaryentry{connected}{name ={connected graph}, description={A\index{connected graph} 
		undirected graph $\graph=\pair{\nodes}{\edges}$ is connected\index{connected graph} if 
		it does not contain a (non-empty) subset $\nodes' \subset \nodes$ with no edges leaving 
		$\nodes'$.}, first={connected},text={connected}}

\newglossaryentry{mvndist}{name ={multivariate normal distribution}, description={The\index{multivariate normal distribution} 
		multivariate normal distribution $\mvnormal{\vm}{\mC}$ is an 
		important family of \gls{probdist}s for a continuous \gls{rv} $\featurevec \in \mathbb{R}^{\nrfeatures}$ \cite{BertsekasProb,GrayProbBook,Lapidoth09}. 
		This family is parametrized by the mean $\vm$ and \gls{covmtx} $\mC$ of $\featurevec$. 
		If the \gls{covmtx} is invertible, the \gls{probdist} of $\featurevec$ is 
		$$p(\featurevec) \propto \exp\bigg(-(1/2) \big( \featurevec - \vm \big)^{T} \mC^{-1} \big( \featurevec - \vm \big) \bigg).$$}, first={multivariate normal distribution},text={multivariate normal distribution}}

\newglossaryentry{statasp}{name ={statistical aspects}, description={By statistical aspects\index{statistical aspects} 
		of a ML method, we refer to (properties of) the \gls{probdist} of its output 
		under a \gls{probmodel} for the data fed into the method.},first={statistical aspects},text={statistical aspects}}

\newglossaryentry{compasp}{name ={computational aspects}, description={By computational 
		aspects\index{computational aspects} of a ML method, we mainly refer to the computational 
		resources required for its implementation. For example, if a ML method uses iterative 
		optimization techniques to solve \gls{erm}, then its computational aspects include (i) how 
		many arithmetic operations are needed to implement a single iteration (\gls{gradstep}) 
		and (ii) how many iterations are needed to obtain useful \gls{modelparams}. One important 
		example of an iterative optimization technique is \gls{gd}.}, first={computational aspects},text={computational aspects}}

\newglossaryentry{zerooneloss}{name={$0/1$ \gls{loss}},
	description={The $0/1$ \gls{loss}\index{$0/1$ loss} $\lossfunc{\pair{\featurevec}{\truelabel}}{\hypothesis}$ 
		measures the quality of a \gls{classifier} $\hypothesis(\featurevec)$ that delivers a \gls{prediction} $\predictedlabel$ (e.g., 
	via thresholding \eqref{equ_def_threshold_bin_classifier}) for the \gls{label} $\truelabel$ of a \gls{datapoint} with \gls{feature}s 
	$\featurevec$. 
	It is equal to $0$ if the \gls{prediction} is correct, i.e., 
	$\lossfunc{\pair{\featurevec}{\truelabel}}{\hypothesis}=0$ when $\predictedlabel=\truelabel$. It is 
	equal to $1$ if the \gls{prediction} is wrong, $\lossfunc{\pair{\featurevec}{\truelabel}}{\hypothesis}=1$ 
	when $\predictedlabel\neq\truelabel$.},
	sort=zerooneloss, 
    first={$0/1$ \gls{loss}},text={$0/1$ \gls{loss}}}

\newglossaryentry{probability}{name={probability},
	description={We\index{probability} assign a probability value, typically chosen in the 
		interval $[0,1]$, to each event that might occur in a random experiment \cite{KallenbergBook,BertsekasProb,BillingsleyProbMeasure,HalmosMeasure}.},first={probability},text={probability}}
	
\newglossaryentry{underfitting}{name={underfitting},description={Consider\index{underfitting} a ML method that uses 
		\gls{erm} to learn a \gls{hypothesis} with minimum \gls{emprisk} on a given \gls{trainset}. 
		Such a method is \emph{underfitting} the \gls{trainset} if it is not able to learn a \gls{hypothesis} 
		with sufficiently small \gls{emprisk} on the \gls{trainset}. If a method is underfitting it will typically 
	also not be able to learn a \gls{hypothesis} with a small \gls{risk}.},first={underfitting},text={underfitting}}

\newglossaryentry{overfitting}{name={overfitting},description={Consider\index{overfitting} a ML method that uses 
		\gls{erm} to learn a \gls{hypothesis} with minimum \gls{emprisk} on a given \gls{trainset}. 
		Such a method is \emph{overfitting} the \gls{trainset} if it learns \gls{hypothesis} with small 
		\gls{emprisk} on the \gls{trainset} but significantly larger \gls{loss} outside the \gls{trainset}.},first={overfitting},text={overfitting}}

\newglossaryentry{gdpr}{name={General Data Protection Regulation},description={The \emph{general data 
			protection regulation}\index{GDPR} (GDPR) is a law that has been passed by the European Union (EU) 
			and put into effect on May 25, 2018 \url{https://gdpr.eu/tag/gdpr/}. The GDPR imposes obligations 
			onto organizations anywhere, so long as they target, collect or in any other way process data 
			related to people (i.e., personal data) in the EU \cite{GDPR2016}.}, 
	first={general data protection regulation (GDPR) },text={GDPR}}
	
\newglossaryentry{gaussrv}{name={Gaussian random variable},description={A\index{Gaussian random variable} 
		Gaussian \gls{rv} $\vx \in \mathbb{R}^{\nrfeatures}$ with a \gls{mvndist}. The special case of $\nrfeatures=1$ 
		corresponds to a scalar Gaussian \gls{rv} \cite{papoulis,BertsekasProb,GrayProbBook}.},first={Gaussian RV},text={Gaussian RV}}

\newglossaryentry{trustworthiness}{name={trustworthiness},description=
	{Beside the \gls{compasp} and \gls{statasp}, a third main design aspect for ML methods 
		is their trustworthiness\index{trustworthy AI} \cite{pfau2024engineeringtrustworthyaideveloper}. 
		The European Union has put forward seven key requirements (KRs) for trustworthy 
		AI (that typically build on ML methods)
	\cite{ALTAIEU}: {\bf KR1 - Human Agency and Oversight}, {\bf KR2 - Technical Robustness and Safety}, 
	{\bf KR3 - Privacy and Data Governance}, {\bf KR4 - Transparency}, {\bf KR5 - Diversity Non-Discrimination and Fairness}, 
	{\bf KR6 Societal and Environmental Well-Being}, {\bf KR7 - Accountability}. 
	},first={trustworthiness},text={trustworthiness}}

\newglossaryentry{sqerrloss}{name={squared error loss},description={The squared 
		error\index{squared error loss} \gls{loss} measures the prediction error of a 
		\gls{hypothesis} $\hypothesis$ when predicting a numeric \gls{label} $\truelabel \in \mathbb{R}$ 
		from the \gls{feature}s $\featurevec$ of a \gls{datapoint}. It is 
	defined as 
\begin{equation} 
	\nonumber
	\lossfunc{(\featurevec,\truelabel)}{\hypothesis} \defeq \big(\truelabel - \underbrace{\hypothesis(\featurevec)}_{=\predictedlabel} \big)^{2}. 
\end{equation} 
},first={squared error loss},text={squared error loss}}

\newglossaryentry{projgd}{name={projected GD},description={Projected\index{projected gradient descent} \gls{gd} 
		extends basic \gls{gd} for unconstrained optimization to handle constraints on the 
		optimization variable (\gls{modelparams}). A single iteration of projected \gls{gd} consists 
		of first taking a \gls{gradstep} and then projecting the result back into a 
		constrain set.},first={projected \gls{gd}},text={projected \gls{gd}}}

\newglossaryentry{diffpriv}
{name=differential privacy,
  description={
  	Consider\index{differential privacy} some ML method $\algomap$ that reads in a $\dataset$ 
  	and delivers some output $\algomap(\dataset)$. The output could be the learnt \gls{modelparams} 
  	$\widehat{\weights}$ or the \gls{prediction} $\learnthypothesis(\featurevec)$ obtained for a specific \gls{datapoint} 
  	with \gls{feature}s $\featurevec$. Differential privacy is a precise measure of privacy leakage incurred 
  	by revealing the output $\algomap(\dataset)$. Roughly speaking, the algorithm is differentially 
  	private if the \gls{probdist} of $\algomap(\dataset)$ does not change too much if a \gls{sensattr} 
  	of one \gls{datapoint} in $\dataset$ is changed.}, 
	first = {differential privacy (DP)}, text={DP} 
}

\newglossaryentry{privprot}
{name=privacy protection,
     description={Consider\index{privacy protection} some ML method $\algomap$ that reads in a $\dataset$ 
     	and delivers some output $\algomap(\dataset)$. The output could be the learnt \gls{modelparams} 
     	$\widehat{\weights}$ or the \gls{prediction} $\learnthypothesis(\featurevec)$ obtained for a specific \gls{datapoint} 
		with \gls{feature}s $\featurevec$. Many important ML applications involve \gls{datapoint}s 
		representing humans. Each \gls{datapoint} is characterized by \gls{feature}s $\featurevec$, 
		potentially a \gls{label} $\truelabel$ and a \gls{sensattr} $\sensattr$ (e.g., a recent medical diagnosis). 
		Roughly speaking, \gls{privprot} means that it should be impossible to infer, from the output $\algomap(\dataset)$, 
		any of the \gls{sensattr}s of \gls{datapoint}s in $\dataset$. Mathematically, \gls{privprot} requires non-invertibility 
		of the map $\algomap(\dataset)$. In general, just making $\algomap(\dataset)$ non-invertible 
		is typically insufficient for \gls{privprot}. We need to make $\algomap(\dataset)$ sufficiently non-invertible. 
	}, 
	first = {privacy protection}, text={privacy protection} 
}

\newglossaryentry{privleakage}
{
	name=privacy leakage,
	description={Consider\index{privacy leakage} a (ML or \gls{fl}) system that processes a \gls{localdataset} $\localdataset{\nodeidx}$ 
		and shares data, such as the predictions obtained for new \gls{datapoint}s, with 
		other parties. Privacy leakage arises if the shared data carries information about a 
		private (sensitive) \gls{feature} of a \gls{datapoint} (which might be a human) of $\localdataset{\nodeidx}$.  
		The amount of privacy leakage can be measured via \gls{mutualinformation} using a 
		\gls{probmodel} for the local \gls{dataset}. Another quantitative measure for privacy 
		leakage is \gls{diffpriv}. 
	}, 
	first = {privacy leakage}, text={privacy leakage} 
}

\newglossaryentry{probmodel}
{
	name=probabilistic model,
	description={A probabilistic model\index{probabilistic model} interprets \gls{datapoint}s 
		as \gls{realization}s of \gls{rv}s with a joint \gls{probdist}. This joint \gls{probdist} typically 
		involves parameters which have to be manually chosen or learnt via statistical inference 
		methods such as \gls{ml} \cite{LC}. }, 
	first = {probabilistic model}, text={probabilistic model} 
}

\newglossaryentry{mean}
{
	name=mean,
	description={The\index{mean} expectation $\expect \{ \featurevec \}$ of a numeric \gls{rv} $\featurevec$.}, 
		first = {mean}, text={mean} 
}

\newglossaryentry{variance}
{
	name={variance},
	description={The\index{variance} variance of a real-valued \gls{rv} $\feature$ is defined as the expectation 
		$\expect\big\{ \big( x - \expect\{x \} \big)^{2} \big\}$ of the squared difference $\feature$ 
		and its expectation $\expect\{x \}$. We extend this definition to vector-valued \gls{rv}s $\featurevec$ 
		as $\expect\big\{ \big\| \featurevec - \expect\{\featurevec \} \big\|_{2}^{2} \big\}$.} ,first={variance},text={variance} 
}

\newglossaryentry{nn}
{
	name={nearest neighbour},
	description={Nearest neighbour\index{nearest neighbour} methods learn a \gls{hypothesis} 
		$\hypothesis: \featurespace \rightarrow \labelspace$ whose function value $\hypothesis(\featurevec)$ 
		is solely determined by the nearest neighbours within a given \gls{dataset}. Different 
		methods use different metrics for determining the nearest neighbours. If \gls{datapoint}s 
		are characterized by numeric \gls{feature} vectors, we can use their Euclidean distances as 
		the metric.},
	first={nearest neighbour (NN)},text={NN} 
}

\newglossaryentry{neighbourhood}
{
	name={neighbourhood},
	description={The\index{neighbourhood} neighbourhood of a node $\nodeidx \in \nodes$ is the subset of nodes 
	constituted by the \gls{neighbours} of $\nodeidx$.},
	first={neighbourhood},text={neighbourhood} 
}

\newglossaryentry{neighbours}
{
	name={neighbours},
	description={The\index{neighbours} neighbours of a node $\nodeidx \in \nodes$ within a \gls{empgraph} are 
	those nodes $\nodeidx' \in \nodes \setminus \{ \nodeidx\}$ that are connected (via an edge) to node $\nodeidx$.},
	first={neighbours},text={neighbours} 
}

\newglossaryentry{bias}
{
	name={bias},
	description={Consider\index{bias} a ML method using a parametrized \gls{hypospace} $\hypospace$. 
		It learns the \gls{modelparams} $\weights \in \mathbb{R}^{\dimlocalmodel}$ using the \gls{dataset} $\dataset=\big\{ \pair{\featurevec^{(\sampleidx)}}{\truelabel^{(\sampleidx)}} \big\}_{\sampleidx=1}^{\samplesize}$. 
		To analyze the properties of the ML method, we typically interpret the \gls{datapoint}s as \gls{realization}s 
		of \gls{iid} \gls{rv}s, $$ \truelabel^{(\sampleidx)} = \hypothesis^{(\overline{\weights})}\big( \featurevec^{(\sampleidx)} \big) + \bm{\varepsilon}^{(\sampleidx)}, \sampleidx=1,\ldots,\samplesize.$$ 
		We can then interpret the ML method as an estimator $\widehat{\weights}$, 
		computed from $\dataset$ (e.g., by solving \gls{erm}). The (squared) bias incurred by the estimate $\widehat{\weights}$ 
		is then defined as $\biasterm^{2} \defeq \big\| \expect \{ \widehat{\weights}  \}- \overline{\weights}\big\|_{2}^{2}$. },
first={bias},text={bias} 
}

\newglossaryentry{classification}
{
	name={classification},
	description={Classification\index{classification} is the task of determining a 
		discrete-valued label $\truelabel$ of a \gls{datapoint} based solely on its 
		features $\featurevec$. The label $\truelabel$ belongs to a finite set, such 
		as $\truelabel \in \{ -1,1\}$, or $\truelabel \in \{1,\ldots,19\}$ and represents a 
		category to which the corresponding \gls{datapoint} belongs to. Some classification 
		problems involve a countably infinite \gls{labelspace}.},first={classification},text={classification} 
}

\newglossaryentry{privfunnel}
{
	name={privacy funnel},
	description={The privacy funnel is a method for learning privacy-friendly features of \gls{datapoint}s \cite{MakhdoumiFunnel2014}. },
	first={privacy funnel},text={privacy funnel} 
}

\newglossaryentry{condnr}
{
	name={condition number},
	description={The condition number\index{condition number} $\kappa(\mathbf{Q}) \geq 1$ of a \gls{psd} 
		matrix $\mathbf{Q}$ is the ratio $\eigvalgen_{\rm max} /\eigvalgen_{\rm min}  $ between the 
		largest $\eigvalgen_{\rm max}$ and the smallest $\eigvalgen_{\rm min}$ \gls{eigenvalue} of 
		$\mathbf{Q}$. The condition number is useful for the analysis of ML methods. 
		The computational complexity of \gls{gdmethods} for \gls{linreg} crucially depends on the 
		condition number of the matrix $\mQ = \mX \mX^{T}$, with the \gls{featuremtx} $\mX$ 
		of the \gls{trainset}. Thus, from a computational perspective, we prefer \gls{feature}s of 
		\gls{datapoint}s such that $\mQ$ has a condition number close to $1$.},first={condition number},text={condition number} 
}

\newglossaryentry{classifier}
{
	name={classifier},
	description={A classifier\index{classifier} is a \gls{hypothesis} (map) $\hypothesis(\featurevec)$ 
		used to predict a \gls{label} taking values from a finite \gls{labelspace}. We might use the 
		function value $\hypothesis(\featurevec)$ itself as a \gls{prediction} $\predictedlabel$ for 
		the \gls{label}. However, it is customary to use a map $\hypothesis(\cdot)$ that delivers 
		a numeric quantity. The \gls{prediction} is then obtained by a simple thresholding step. 
		For example, in a binary classification problem with \label{labelspace} $\labelspace \in  \{ -1,1\}$, 
		we might use real-valued \gls{hypothesis} map $\hypothesis(\featurevec) \in \mathbb{R}$ 
		as classifier. A \gls{prediction} $\predictedlabel$ can then be obtained via thresholding,  
		 \begin{equation} 
		 	\label{equ_def_threshold_bin_classifier}
		 	\predictedlabel =1   \mbox{ for } \hypothesis(\featurevec) \geq 0, \mbox{ and } 	\predictedlabel =-1  \mbox{ otherwise.}
	 		\end{equation}
 		We can characterize a classifier by its \gls{decisionregion}s $\decreg{a}$, for 
 		every possible \gls{label} value $a \in \labelspace$. },first={classifier},text={classifier} 
}

\newglossaryentry{emprisk}
{name={empirical risk},
  description={The empirical risk\index{empirical risk} $\emprisk{\hypothesis}{\dataset}$ 
  	of a \gls{hypothesis} on a \gls{dataset} $\dataset$ is the average \gls{loss} incurred 
  	by $\hypothesis$ when applied to the \gls{datapoint}s in $\dataset$.},
  first={empirical risk},text={empirical risk} 
}

\newglossaryentry{nodedegree}
{name={node degree},
	description={The degree\index{node degree} $\nodedegree{\nodeidx}$ of a node $\nodeidx \in \nodes$ 
		in an undirected \gls{graph} is the number of its \gls{neighbours}, $\nodedegree{\nodeidx} \defeq \big|\neighbourhood{\nodeidx}\big|$.},first={node degree},text={node degree} 
}

\newglossaryentry{graph}
{name={graph},
	description={A graph\index{graph} $\graph = \pair{\nodes}{\edges}$ is a pair that consists of 
		a node set $\nodes$ and an edge set $\edges$. In its most general form, a graph is 
		specified by a map that assigns to each edge $\edgeidx \in \edges$ a pair of nodes \cite{RockNetworks}. 
		One important family of graphs are simple undirected graphs. A simple undirected graph 
		is obtained by identifying each edge $\edgeidx \in \edges$ with two different nodes $\{\nodeidx,\nodeidx'\}$. 
		Weighted graphs also specify numeric weights $\edgeweight_{\edgeidx}$ for each 
		edge $\edgeidx \in \edges$.},first={graph},text={graph} 
}

\newglossaryentry{empgraph}
{name={federated network},
	description={A federated network\index{federated network} is an undirected weighted \gls{graph} whose 
		nodes represent data generators that aim to train a local (or personalized) \gls{model}. 
		Each node in a federated network represents some device, capable to collect a \gls{localdataset} 
		and, in turn, train a local \gls{model}. 
	    \Gls{fl} methods learn a local \gls{hypothesis} $\localhypothesis{\nodeidx}$, for 
	    each node $\nodeidx \in \nodes$, such that it incurs small \gls{loss} on the \gls{localdataset}s.},first={federated network},text={federated network} 
}

\newglossaryentry{norm}
{name={norm},
	description={A norm\index{norm} is a function that maps each element (vector) 
		of a linear vector space to a non-negative real number. This function must be 
		homogeneous, definite and satisfy the triangle inequality \cite{HornMatAnalysis}. },
	first={norm},text={norm} 
}

\newglossaryentry{explanation}
{name={explanation},
	description={One approach to make ML methods transparent, is to provide an 
		explanation\index{explanation} along with the \gls{prediction} delivered by an 
		ML method. Explanations can take on many different forms. An explanation 
		could be some natural text or some quantitative measure for the importance 
		of individual \gls{feature}s of a \gls{datapoint} \cite{Molnar2019}. We can also 
		use visual forms of explanations such as intensity plots for image classification \cite{GradCamPaper}.},
	first={explanation},text={explanation} 
}

\newglossaryentry{risk}
{name={risk},
	description={Consider\index{risk} a \gls{hypothesis} $\hypothesis$ used to predict the \gls{label} 
		$\truelabel$ of a \gls{datapoint} based on its \gls{feature}s $\featurevec$. We measure 
		the quality of a particular \gls{prediction} using a \gls{lossfunc} $\lossfunc{(\featurevec,\truelabel)}{\hypothesis}$. 
		If we interpret \gls{datapoint}s as the \gls{realization}s of \gls{iid} \gls{rv}s, 
		also the $\lossfunc{(\featurevec,\truelabel)}{\hypothesis}$ becomes the \gls{realization} 
		of a \gls{rv}. The \gls{iidasspt} allows to define the risk of a \gls{hypothesis} 
		as the expected \gls{loss} $\expect \big\{\lossfunc{(\featurevec,\truelabel)}{\hypothesis} \big\}$. 
		Note that the risk of $\hypothesis$ depends on both, the specific choice for the \gls{lossfunc} and the 
		\gls{probdist} of the \gls{datapoint}s.},
	first={risk},text={risk} 
}

\newglossaryentry{actfun}
{name={activation function},
	description={Each\index{activation function} artificial neuron within an \gls{ann} is 
		assigned an activation function $g(\cdot)$ that maps a weighted combination of 
		the neuron inputs $\feature_{1},\ldots,\feature_{\nrfeatures}$ to a single output 
		value $a = g\big(\weight_{1} \feature_{1}+\ldots+\weight_{\nrfeatures} \feature_{\nrfeatures} \big)$. 
		Note that each neuron is parametrized by the weights $\weight_{1},\ldots,\weight_{\nrfeatures}$.},
first={activation function},text={activation function} 
}

\newglossaryentry{transparency}
{name={transparency},
	description={Transparency\index{transparency} is a key requirement for 
		trustworthy AI \cite{HLEGTrustworhtyAI}. In the context of ML methods, 
		such as \gls{erm}-based methods, transparency is mainly used synonymously for \gls{explainability} \cite{gallese2023ai,JunXML2020}. 
		However, in the wide context of AI systems, transparency also includes providing information 
		about limitations and reliability of the AI system. As a point in case, \gls{logreg} provides a 
		quantitative measure for the reliability of a \gls{classification} in the form of the value $|\hypothesis(\featurevec)|$. 
		Transparency also includes the user interface, by requiring to clearly indicate when a user is 
		interaction with an AI system. Another component of transparency is the documentation 
		of the system’s purpose, design choices and intended use cases \cite{Shahriari2017,DatasheetData2021,10.1145/3287560.3287596}. },
	first={transparency},text={transparency} 
}

\newglossaryentry{sensattr}
{name={sensitive attribute},
	description={ML\index{sensitive attribute} revolves around learning a \gls{hypothesis} map that allows 
		to predict the \gls{label} of a \gls{datapoint} from its \gls{feature}s. In some 
		applications we must ensure that the output delivered by an ML system does 
		not allow to infer sensitive attributes of a \gls{datapoint}. Which parts of a \gls{datapoint} 
		is considered as a sensitive attribute is a design choice that varies across 
		different application domains.},
	first={sensitive attribute},text={sensitive attribute} 
}

\newglossaryentry{sbm}
{name={stochastic block model},
	description={The\index{stochastic block model} stochastic block model (SBM) is a 
		probabilistic generative model for an undirected graph $\graph = \big( \nodes, \edges \big)$ 
		with a given set of nodes $\nodes$ \cite{AbbeSBM2018}. In its most basic variant, 
		the SBM generates a graph by first randomly assigning each node $i \in \nodes$ to 
		a cluster index $c_{i} \in \{1,\ldots,\nrcluster\}$. A pair of different nodes in the 
		graph is connected by an edge with probability $p_{i,j}$ that depends solely on the labels $c_{i}, c_{j}$. 
		The presence of edges between different pairs of nodes is statistically independent. },
	first={stochastic block model (SBM)},text={SBM} 
}

\newglossaryentry{deepnet}
{name={deep net},
	description={We\index{deep net} refer to an \gls{ann} with a (relatively) large number of hidden layers as a 
		deep \gls{ann} or \emph{deep net}. Deep nets are used to represent the \gls{hypospace}s 
		of deep learning methods \cite{Goodfellow-et-al-2016}.},
	first={deep \gls{ann} (deep net)},text={deep net} 
}

\newglossaryentry{baseline}
{name={baseline},
	description={A\index{baseline} reference value or benchmark for the average \gls{loss} 
		incurred by a \gls{hypothesis} when applied to the \gls{datapoint}s generated in a 
		specific ML application. Such a reference value might be obtained from 
		human performance (e.g., the error rate of dermatologists diagnosing cancer 
		from visual inspection of skin areas) or other ML methods \cite[Ch. 6]{MLBasics}.}, 
	first={baseline},text={baseline} 
}

\newglossaryentry{spectrogram}
{name={spectrogram},
	description={The spectrogram\index{spectrogram} of a time signal, e.g., an audio recording, 
		characterizes the time-frequency distribution of the signal. Loosely speaking, the spectrogram 
		quantifies the signal energy within a specific period and frequency interval.}, 
	first={spectrogram},text={spectrogram} 
}

\newglossaryentry{specclustering}
{name={spectral clustering},
	description={Spectral clustering\index{spectral clustering} groups the nodes of an 
		undirected graph by applying \gls{kmeans} clustering to node-wise feature vectors. 
		These feature vectors are built from the \gls{eigenvector}s of the graph \gls{LapMat} \cite{Luxburg2007,FlowSpecClustering2021}. }, 
	first={spectral clustering},text={spectral clustering} 
}

\newglossaryentry{flowbasedclustering}
{name={flow-based clustering},
	description={Flow-basted clustering\index{flow-based clustering} groups the nodes 
		of an undirected graph by applying \gls{kmeans} clustering to node-wise feature 
		vectors. These feature vectors are built from flows between carefully selected 
		source and destination nodes \cite{FlowSpecClustering2021}. }, 
	first={flow-based clustering},text={flow-based clustering} 
}

\newglossaryentry{esterr}
{name={estimation error},
	description={Consider\index{estimation error} \gls{datapoint}s with feature vectors $\featurevec$ and \gls{label} 
		$\truelabel$. In some applications we can model the relation between the \gls{feature}s and the \gls{label}
		of a \gls{datapoint} as $\truelabel = \bar{\hypothesis}(\featurevec) + \varepsilon$. Here we 
		used some true \gls{hypothesis} $\bar{\hypothesis}$ and a noise term $\varepsilon$ which might represent 
		modelling or labelling errors. The estimation error incurred by a ML method that learns a 
		\gls{hypothesis} $\widehat{\hypothesis}$, e.g., using \gls{erm}, is defined as 
		$\widehat{\hypothesis}(\featurevec) - \bar{\hypothesis}(\featurevec)$, for some \gls{feature} vector. 
		For a parametrized \gls{hypospace}, consisting of \gls{hypothesis} maps that are determined by 
		a \gls{modelparams} $\weights$, we can define the estimation error in terms of parameter vectors 
		as $\Delta \weights = \widehat{\weights} - \overline{\weights}$ \cite{kay,hastie01statisticallearning}.}
	first={estimation error},text={estimation error} 
}

\newglossaryentry{dob}
{name={degree of belonging},
	description={A\index{degree of belonging} number that indicates the extent by which a \gls{datapoint} 
		belongs to a \gls{cluster} \cite[Ch. 8]{MLBasics}. The degree of belonging can be 
		interpreted as a soft \gls{cluster} assignment. Soft clustering methods typically represent 
		the degree of belonging by a real number in the interval $[0,1]$. The extreme case 
		when the degree of belonging only takes on values $0$ or $1$ correspond to 
		hard clustering.}, first={degree of belonging},text={degree of belonging} 
}

\newglossaryentry{msee}
{name={mean squared estimation error},
	description={Consider\index{mean squared estimation error} a ML method that 
		learns \gls{modelparams} $\widehat{\weights}$ based on some \gls{dataset} $\dataset$. 
		If we interpret the \gls{datapoint}s in $\dataset$ as \gls{iid} \gls{realization}s of a \gls{rv} $\datapoint$, 
		we define the \gls{esterr} $\Delta \weights \defeq \widehat{\weight} - \overline{\weights}$. 
		Here, $\overline{\weights}$ denotes the true \gls{modelparams} of the \gls{probdist} 
		of $\datapoint$.The mean squared estimation error is 
		defined as the \gls{expectation} $\expect \big\{ \big\| \Delta \weights \big\|^{2} \big\}$ of the 
		squared Euclidean norm of the \gls{esterr} \cite{LC,kay}.},
	first={mean squared estimation error (MSEE)},text={MSEE} 
}

\newglossaryentry{gtvmin}
{name={GTV minimization},
	description={GTV minimization\index{total variation minimization} is an instance of \gls{rerm} 
		using the \gls{gtv} of local \gls{modelparams} as a \gls{regularizer} \cite{ClusteredFLTVMinTSP}.},
	first={GTV minimization (GTVMin)},text={GTVMin} 
}

\newglossaryentry{regression}
{name={regression},
	description={Regression\index{regression} problems revolve around the problem of 
		predicting a numeric \gls{label} solely from the \gls{feature}s of a \gls{datapoint} \cite[Ch. 2]{MLBasics}.},
	first={regression},text={regression} 
}

\newglossaryentry{acc}
{name={accuracy},
	description={Consider\index{accuracy} \gls{datapoint}s characterized by \gls{feature}s $\featurevec \in \featurespace$ and 
		a categorical label $\truelabel$ which takes on values from a finite \gls{labelspace} $\labelspace$. The 
		accuracy of a \gls{hypothesis} $\hypothesis: \featurespace \rightarrow \labelspace$, when applied 
		to the \gls{datapoint}s in a \gls{dataset} $\dataset = \big\{ \big(\featurevec^{(1)}, \truelabel^{(1)} \big), \ldots, \big(\featurevec^{(\samplesize)},\truelabel^{(\samplesize)}\big) \big\}$ 
		is then defined as $1 - (1/\samplesize)\sum_{\sampleidx=1}^{\samplesize} \lossfunc{\big(\featurevec^{(\sampleidx)},\truelabel^{(\sampleidx)}\big)}{\hypothesis}$ using the \gls{zerooneloss}.},
	first={accuracy},text={accuracy} 
}

\newglossaryentry{expert}
{name={expert},
	description={ML\index{expert} aims at learning a \gls{hypothesis} $\hypothesis$ that accurately predicts the label 
		of a \gls{datapoint} based on its features. We measure the prediction error using 
		some \gls{lossfunc}. Ideally we want to find a \gls{hypothesis} that incurs minimum \gls{loss}. 
		One approach to make this goal precise is to use the \gls{iidasspt} and use the resulting 
		\gls{bayesrisk} as the benchmark level for the (average) \gls{loss} of a \gls{hypothesis}. 
		Alternatively, we might know a reference or benchmark \gls{hypothesis} $\hypothesis'$ 
		which might be obtained by some existing ML method. We can then compare the \gls{loss} 
		incurred by $\hypothesis$ against the \gls{loss} incurred by $\hypothesis'$. Such a reference 
		or baseline hypothesis $\hypothesis'$ is referred to as an \gls{expert} \cite{PredictionLearningGames}. 
		Note that an expert might deliver very poor predictions. We typically compare against many 
	   different experts and aim at incurring not much more \gls{loss} than the best among those experts (this is 
	   known as regret minimization) \cite{PredictionLearningGames,HazanOCO}.}
	first={expert},text={expert} 
}

\newglossaryentry{nfl}
{name={networked federated learning},
	description={Networked\index{networked federated learning} federated learning refers 
		to methods that learn personalized models in a distributed fashion from \gls{localdataset}s 
		that are related by an intrinsic network structure.},
 first={networked federated learning (NFL)},text={NFL} 
}

\newglossaryentry{regret}
{name={regret},
	description={The regret\index{regret} of a \gls{hypothesis} $\hypothesis$ relative to 
		another \gls{hypothesis} $\hypothesis'$, which serves as a reference (or baseline), 
		is the difference between the \gls{loss} incurred by $\hypothesis$ and the \gls{loss} 
		incurred by $\hypothesis'$ \cite{PredictionLearningGames}. 
		      The baseline \gls{hypothesis} $\hypothesis'$ is also referred to as an \gls{expert}.},
	first={regret},text={regret} 
}

\newglossaryentry{strcvx}
{name={strongly convex},
	description={A\index{strongly convex} continuously \gls{differentiable} real-valued 
		function $f(\featurevec)$ is strongly convex with coefficient $\sigma$ if $f(\vy) \geq f(\vx) + \nabla f(\vx)^{T} (\vy - \vx) + (\sigma/2) \normgeneric{\vy - \vx}{2}^{2}$ \cite{nesterov04},\cite[Sec. B.1.1.]{CvxAlgBertsekas}.},
	first={strongly convex},text={strongly convex} 
}

\newglossaryentry{differentiable}
{name={differentiable},
	description={A\index{differentiable} function $f: \mathbb{R}^{\featuredim} \rightarrow \mathbb{R}$  is differentiable if it 
		has a \gls{gradient} $\nabla f ( \mathbf{x})$ everywhere (for every $\mathbf{x} \in \mathbb{R}^{\featuredim}$) \cite{RudinBookPrinciplesMatheAnalysis}.},
	first={differentiable},text={differentiable} 
}

\newglossaryentry{gradient}
{name={gradient},
	description={For\index{gradient} a real-valued function $f: \mathbb{R}^{\featuredim} \rightarrow \mathbb{R}: \weights \mapsto f(\weights)$, 
	a vector $\vg$ such that $\lim_{\weights \rightarrow \weights'} \frac{f(\weights) - \big(f(\weights')+ \vg^{T} (\weights- \weights') \big) }{\| \weights-\weights'\|}=0$ 
	is referred to as the gradient of $f$ at $\weights'$. If such a vector exists it is 
	denoted $\nabla f(\weights')$ or $\nabla f(\weights)\big|_{\weights'}$ \cite{RudinBookPrinciplesMatheAnalysis}.},
	first={gradient},text={gradient} 
}

\newglossaryentry{subgradient}
{name={subgradient},
description={For\index{subgradient} a real-valued function $f: \mathbb{R}^{\featuredim} \rightarrow \mathbb{R}: \weights \mapsto f(\weights)$, 
		a vector $\va$ such that $f(\weights) \geq  f(\weights') +\big(\weights-\weights' \big)^{T} \va$ is 
		referred to as a subgradient of $f$ at $\weights'$ \cite{BertCvxAnalOpt,BertsekasNonLinProgr}.},
	first={subgradient},text={subgradient} 
}

\newglossaryentry{fedavg}
{name={federated averaging (FedAvg)},
	description={Federated\index{federated averaging (FedAvg)} averaging is an iterative \gls{fl} algorithm that alternates between local model trainings and 
	averaging the resulting local model parameters. Different variants of this algorithm are obtained 
	by different techniques for the model training. The authors of \cite{pmlr-v54-mcmahan17a} consider 
	federated averaging methods where the local model training is implemented by running several \gls{gd} steps }, first = {federated averaging (FedAvg)}, text={FedAvg} 
}

\newglossaryentry{relu}
{name={rectified linear unit (ReLU)},
	description={The\index{rectified linear unit (ReLU)} rectified linear unit (ReLU) is 
		a popular choice for the \gls{actfun} of a neuron within an \gls{ann}. It is defined 
		as $g(z) = \max\{0,z\}$ with $z$ being the weighted input of the artificial neuron.}, first = {rectified linear unit (ReLU)}, text={ReLU} 
}

\newglossaryentry{hypothesis}
{name={hypothesis},
	description={A\index{hypothesis} map (or function) $\hypothesis: \featurespace \rightarrow \labelspace$ from the 
		\gls{featurespace} $\featurespace$ to the \gls{labelspace} $\labelspace$. 
		Given a \gls{datapoint} with \gls{feature}s $\featurevec$ we use a hypothesis map $\hypothesis$
		to estimate (or approximate) the \gls{label} $\truelabel$ using the \gls{prediction}  
		$\hat{\truelabel} = \hypothesis(\featurevec)$. ML is all about learning (or finding) a 
		hypothesis map $\hypothesis$ such that $\truelabel \approx \hypothesis(\featurevec)$ 
		for any \gls{datapoint} (having \gls{feature}s $\featurevec$ and \gls{label} $\truelabel$).},
	first={hypothesis},text={hypothesis}  
}

\newglossaryentry{vcdim}
{name={Vapnik–Chervonenkis (VC) dimension},
	description={The\index{VC dimension} VC dimension of an infinite \gls{hypospace} is a widely-used measure 
		for its size. We refer to \cite{ShalevMLBook} for a precise definition of VC dimension 
		as well as a discussion of its basic properties and use in ML.},
	first={Vapnik–Chervonenkis (VC) dimension},text={VC dimension}  
}

\newglossaryentry{effdim}
{name={effective dimension},
	description={The\index{effective dimension} effective dimension $\effdim{\hypospace}$ of 
		an infinite \gls{hypospace} $\hypospace$ is a measure of its size. Loosely speaking, the 
		effective dimension is equal to the effective number of independent tunable parameters 
		of the model. These parameters might be the coefficients used in a linear map or the 
		weights and bias terms of an \gls{ann}.},
	first={effective dimension},text={effective dimension}  
}

\newglossaryentry{labelspace}
{name={label space},
	description={Consider\index{label space} a ML application that involves \gls{datapoint}s characterized by features 
		and labels. The \gls{label} space is constituted by all potential values that the \gls{label} 
		of a \gls{datapoint} can take on. Regression methods, aiming at predicting numeric \gls{label}s, often
 use the \gls{label} space $\labelspace = \mathbb{R}$. Binary classification methods use a label space 
		 that consists of two different elements, e.g., $\labelspace =\{-1,1\}$, $\labelspace=\{0,1\}$ 
		or $\labelspace = \{ \mbox{``cat image''}, \mbox{``no cat image''} \}$  }, first={label space},text={label space}  
}

\newglossaryentry{prediction}
{name={prediction},
	description={A\index{prediction} prediction is an estimate or approximation for some 
		quantity of interest. ML revolves around learning or finding a \gls{hypothesis} map $\hypothesis$ 
		that reads in the \gls{feature}s $\featurevec$ of a \gls{datapoint} and delivers a \gls{prediction} 
		$\widehat{\truelabel} \defeq \hypothesis(\featurevec)$ for its \gls{label} $\truelabel$. },
	first={prediction},text={prediction}  
}

\newglossaryentry{histogram}
{name={histogram},
	description={Consider\index{histogram} a \gls{dataset} $\dataset$ that consists of $\samplesize$ \gls{datapoint}s 
		$\datapoint^{(1)},\ldots,\datapoint^{(\samplesize)}$, each of them belonging to some 
		cell $[-U,U] \times \ldots \times [-U,U] \subseteq \mathbb{R}^{\featuredim}$ with side 
		length $U$. We partition this cell evenly into smaller elementary cells with side 
		length $\Delta$. The histogram of $\dataset$ assigns each elementary cell to 
		the corresponding fraction of \gls{datapoint}s in $\dataset$ that fall into this 
		elementary cell. 
	},
	first={histogram},text={histogram}  
}

\newglossaryentry{bootstrap}
{name={bootstrap},
	description={For\index{bootstrap}  the analysis of ML methods it is often useful to interpret 
		a given set of \gls{datapoint}s $\dataset = \big\{ \datapoint^{(1)},\ldots,\datapoint^{(\samplesize)}\big\}$ 
		as realizations of \gls{iid} \gls{rv}s with a common \gls{probdist} $p(\datapoint)$. In general, we 
		do not know $p(\datapoint)$ exactly, but we need to estimate it. The bootstrap uses the 
		histogram of $\dataset$ as an estimator for the underlying \gls{probdist} $p(\datapoint)$. 
	},
	first={bootstrap},text={bootstrap}  
}

\newglossaryentry{featurespace}
{name={feature space},
	description={
		The\index{feature space} \gls{feature} space of a given ML application or method is 
		constituted by all potential values that the \gls{feature} vector of a \gls{datapoint} can 
		take on. A widely used choice for the feature space is the \gls{euclidspace} $\mathbb{R}^{\featuredim}$ 
		with dimension $\featurelen$ being the number of individual \gls{feature}s of a \gls{datapoint}.},
	first={feature space},text={feature space}  
}

\newglossaryentry{missingdata}
{name={missing data},
	description={Consider\index{missing data} a \gls{dataset} constituted by \gls{datapoint}s collected via 
		some physical device. Due to imperfections and failures, some of the \gls{feature} 
		or \gls{label} values of \gls{datapoint}s might be corrupted or simply \emph{missing}. 
		Data imputation aims at estimating these missing values \cite{Abayomi2008DiagnosticsFM}. 
		We can interpret data imputation as a ML problem where the \gls{label} of a \gls{datapoint} is 
		the value of the corrupted \gls{feature}. },
	first={missing data},text={missing data}  
}

\newglossaryentry{psd}
{name={positive semi-definite},
	description={A\index{positive semi-definite} symmetric real-valued 
		matrix $\mQ = \mQ^{T} \in \mathbb{R}^{\featuredim \times \featuredim}$ is referred to as 
		positive semi-definite if $\featurevec^{T} \mQ \featurevec \geq 0$ for every 
		vector $\featurevec \in \mathbb{R}^{\featuredim}$.},
	first={positive semi-definite (psd)},text={psd}  
}

\newglossaryentry{feature}
{name={feature},
	description={A\index{feature} feature of a \gls{datapoint} is one of its properties that can be 
		measured or computed easily without the need for human supervision. For example, if a \gls{datapoint} 
		is a bitmap image, then we could use the red-green-blue intensities of its pixels as features. 
		Some widely used synonyms for the term feature are \emph{covariate}, \emph{explanatory variable}, \emph{independent variable}, 
		\emph{input (variable)}, \emph{predictor (variable)} or \emph{regressor} \cite{Gujarati2021,Dodge2003,Everitt2022}. 
		}, first={feature},text={feature}  
}

\newglossaryentry{label}
{name={label},
	description={A\index{label} higher-level fact or quantity of interest associated with a \gls{datapoint}. 
		For example, if the \gls{datapoint} is an image, the label could indicate whether the 
		image contains a cat or not. Synonyms for label, commonly used in specific domains, 
		include \emph{response variable}, \emph{output variable}, and \emph{target} \cite{Gujarati2021,Dodge2003,Everitt2022}.
 },
	first={label},text={label}  
}

\newglossaryentry{data}
{name={data},
	 description={See\index{data} \gls{dataset}.},
	text={data}
}

\newglossaryentry{dataset}
{name={dataset},
	description={With\index{dataset} a slight abuse of notation we use the term \emph{dataset} or \emph{set of \gls{datapoint}s} 
		to refer to an indexed list of \gls{datapoint}s $\datapoint^{(1)},\datapoint^{(2)},\ldots$. Thus, there 
		is a first \gls{datapoint} $\datapoint^{(1)}$, a second \gls{datapoint} $\datapoint^{(2)}$ and so on. 
		Strictly speaking, a dataset is a list and not a set \cite{HalmosSet}. The implementation of ML 
		methods requires a more detailed specification of a dataset, e.g., in the form of a relational database 
		schema \cite{silberschatz2019database}.},first={dataset},text={dataset}  
}

\newglossaryentry{predictor}
{name={predictor},
	description={A\index{predictor} predictor is a real-valued \gls{hypothesis} map. 
		Given a \gls{datapoint} with \gls{feature}s $\featurevec$, the value 
		$\hypothesis(\featurevec) \in \mathbb{R}$ is used as a \gls{prediction} for the true 
		numeric label $\truelabel \in \mathbb{R}$ of the \gls{datapoint}. },first={predictor},text={predictor}  
}

\newglossaryentry{labeled datapoint}
{name={labeled datapoint},
 description={A\index{labeled data} \gls{datapoint} whose label is known or has been determined 
 	by some means which might involve human experts.},
 first={labeled datapoint},text={labeled datapoint}  
}

\newglossaryentry{rv}
{name={random variable (RV)},
 description={A\index{random variable (RV)} random\index{probability space} 
 		variable is a mapping from a probability space $\mathcal{P}$ to a value space \cite{BillingsleyProbMeasure}. 
 	The probability space, whose elements are elementary events, is equipped with a probability 
 	 measure that assigns a probability to subsets of $\mathcal{P}$. A binary random variable maps elementary events 
 	to a set containing two different values, e.g., $\{-1,1\}$ or $\{ \mbox{cat}, \mbox{no cat} \}$. 
 	A real-valued random variable maps elementary events to real numbers $\mathbb{R}$. 
 	A vector-valued random variable maps elementary events to the \gls{euclidspace} $\mathbb{R}^{\featuredim}$. 
 	Probability theory uses the concept of measurable spaces to rigorously define and study the properties of (large) 
 	collections of random variables \cite{GrayProbBook,BillingsleyProbMeasure}.}, first={RV},text={RV}  }

\newglossaryentry{realization}
{name={realization},
	description={Consider\index{realization} a \gls{rv} $x$ which maps each element (outcome, or elementary event) $\omega \in \mathcal{P}$ of a 
		probability space $\mathcal{P}$ to an element $a$ of a measurable space $\mathcal{N}$ \cite{BillingsleyProbMeasure,RudinBookPrinciplesMatheAnalysis,HalmosMeasure}. 
		A realization of $x$ is any element $a' \in \mathcal{N}$ such that there is 
		an element $\omega' \in \mathcal{P}$ with $x(\omega') = a'$.  }, first={realization},text={realization}  }

\newglossaryentry{trainset}
{name={training set},
description={A\index{training set} \gls{dataset} $\dataset$, constituted by some \gls{datapoint}s used in \gls{erm} 
	to learn a \gls{hypothesis} $\learnthypothesis$. The average \gls{loss} of $\learnthypothesis$ on the 
	training set is referred to as the \gls{trainerr}. The comparison between \gls{trainerr} and \gls{valerr} 
	of $\learnthypothesis$ allows to diagnose ML methods and informs how to improve them (e.g., using a different \gls{hypospace} 
	or collecting more \gls{datapoint}s) \cite[Sec. 6.6.]{MLBasics}.},first={training set},text={training set}  
}

\newglossaryentry{netmodel}
{name={networked model},
  description={A\index{networked model} networked model over a \gls{empgraph} $\graph = \pair{\nodes}{\edges}$ assigns 
   a \gls{localmodel} (\gls{hypospace}) to each node $\nodeidx \in \nodes$ of the \gls{empgraph} $\graph$.}, 
   first={networked model},text={networked model}  
}

\newglossaryentry{batch}
{
	name={batch},
	description={A\index{batch} set of \gls{datapoint}s randomly selected from 
		(typically very large) \gls{dataset}.}, 
	first={batch},text={batch}  
}

\newglossaryentry{netdata}
{
	name={networked data},
	description={Networked\index{networked data} data consists of \gls{localdataset}s that are 
		related by some notion of pair-wise similarity. We can represent networked data using a \gls{empgraph} whose nodes carry \gls{localdataset}s and an edge indicates a similarity between the connected nodes.}, 
	first={networked data},text={networked data}  
}

\newglossaryentry{trainerr}
{
	name={training error},
	description={The\index{training error} average \gls{loss} of a \gls{hypothesis} when 
		predicting the \gls{label}s of \gls{datapoint}s in a \gls{trainset}. We sometimes refer 
		by training error also the minimum average \gls{loss} incurred on the \gls{trainset} 
		by the optimal \gls{hypothesis} from a \gls{hypospace}.},first={training error},text={training error}  
}

\newglossaryentry{datapoint}
{name={data point},
description={A\index{data point} \gls{datapoint} is any object that conveys information \cite{coverthomas}. Data points might be 
		students, radio signals, trees, forests, images, \gls{rv}s, real numbers or proteins. We characterize data points 
		using two types of properties. One type of property is referred to as a \gls{feature}. \Gls{feature}s are properties of a 
		\gls{datapoint} that can be measured or computed in an automated fashion. 
		A different kind of property is referred to as \gls{label}s. The \gls{label} of 
		a \gls{datapoint} represents some higher-level fact (or quantity of interest). In 
		contrast to \gls{feature}s, determining the \gls{label} of a \gls{datapoint} typically 
		requires human experts (domain experts). Roughly speaking, ML aims to predict 
		the \gls{label} of a \gls{datapoint} based solely on its \gls{feature}s. 
		}, first={data point},text={data point}  
}

\newglossaryentry{valerr}
{name={validation error},
 description={Consider\index{validation error} a \gls{hypothesis} $\learnthypothesis$ which is 
 	obtained by some ML method, e.g., using \gls{erm} on a \gls{trainset}. The average \gls{loss} 
 	of $\learnthypothesis$ on a \gls{valset}, which is different from the \gls{trainset}, is referred 
 	to as the validation error.},first={validation error},text={validation error}  
}

\newglossaryentry{validation} 
{name={validation},
	description={Consider\index{validation} a \gls{hypothesis} $\learnthypothesis$ that has been 
		learnt via some ML method, e.g., by solving \gls{erm} on a \gls{trainset} $\dataset$. 
		Validation refers to the practice of evaluating the \gls{loss} incurred by 
		\gls{hypothesis} $\learnthypothesis$ on a \gls{valset} that consists of 
		\gls{datapoint}s that are not contained in the \gls{trainset} $\dataset$. },first={validation},text={validation}  
}

\newglossaryentry{quadfunc}
{name={quadratic function},
	description={A\index{quadratic function} quadratic function $f(\weights)$, reading in a 
		vector $\weights \in \mathbb{R}^{\nrfeatures}$ as its argument, is such that $$f(\weights) =  \weights^{T} \mathbf{Q} \mathbf{w} + \mathbf{q}^{T} \weights+a,$$ with some matrix $\mQ \in \mathbb{R}^{\nrfeatures \times \nrfeatures}$, 
		vector $\vq \in \mathbb{R}^{\nrfeatures}$ and scalar $a \in \mathbb{R}$.  },first={quadratic function},text={quadratic function}  
}

\newglossaryentry{valset}
{name={validation set},
  description={A\index{validation set} set of \gls{datapoint}s used to estimate 
  	the \gls{risk} of a \gls{hypothesis} $\learnthypothesis$ that has been learnt by some 
  	ML method (e.g., solving \gls{erm}). The average \gls{loss} of $\learnthypothesis$ 
  	on the validation set is referred to as the validation error and can be used to diagnose a 
  	ML method (see \cite[Sec. 6.6.]{MLBasics}). The comparison between \gls{trainerr} 
  	and \gls{valerr} can inform directions for improvements of the ML method (such as 
  	using a different \gls{hypospace}).},first={validation set},text={validation set}  
}

\newglossaryentry{testset}
{name={test set},
	description={A\index{test set} set of \gls{datapoint}s that have neither 
		been used to train a \gls{model}, e.g., via \gls{erm}, nor in a \gls{valset} 
		to choose between different \gls{model}s.},first={test set},text={test set}  
}

\newglossaryentry{modelsel}
{name={model selection},
	description={In\index{model selection} ML, model selection refers to the 
		process of choosing between different candidate \gls{model}s. In its most 
		basic form, \gls{model} selection amounts to (i) training each candidate \gls{model}, 
		(ii) computing the \gls{valerr} for each trained \gls{model}, (iii) choosing the \gls{model} 
		with smallest \gls{valerr} \cite[Ch. 6]{MLBasics}. },first={model selection},text={model selection}  
}

\newglossaryentry{linclass}{name={linear classifier}, description={
	    Consider\index{linear classifier} \gls{datapoint}s characterized by numeric \gls{feature}s $\featurevec \in \mathbb{R}^{\nrfeatures}$ 
	    and a \gls{label} $\truelabel \in \labelspace$ from some finite \gls{labelspace} $\labelspace$. 
		A linear \gls{classifier} characterized by having \gls{decisionregion}s separated 
		by hyperplanes in the \gls{euclidspace} $\mathbb{R}^{\featuredim}$ \cite[Ch. 2]{MLBasics}.},first={linear classifier},text={linear classifier} }

\newglossaryentry{erm}{name={empirical risk minimization}, description={Empirical risk 
		minimization\index{empirical risk minimization} is the optimization problem of finding 
		a \gls{hypothesis} (out of a \gls{model}) with minimum average \gls{loss} (or \gls{emprisk}) on a given \gls{dataset} 
		$\dataset$ (the \gls{trainset}). Many ML methods are obtained from 
		\gls{emprisk} via specific design choices for the \gls{dataset}, \gls{model} and \gls{loss} \cite[Ch. 3]{MLBasics}.},
	first={empirical risk minimization (ERM)},text={ERM} }

\newglossaryentry{multilabelclass}{name={multi-label classification}, description={Multi-label 
		classification\index{multi-label classification} problems and methods use \gls{datapoint}s 
		that are characterized by several \gls{label}s. As an example, consider a \gls{datapoint} 
		representing a picture with one binary \gls{label} indicating the presence of a human 
		in this picture and another \gls{label} indicating the presence of a car.},
	    first={multi-label classification},text={multi-label classification} }

\newglossaryentry{ssl}{name={semi-supervised learning}, description={Semi-supervised\index{semi-supervised learning} 
		learning methods use unlabeled \gls{datapoint}s to support the learning of a \gls{hypothesis} 
		from labeled \gls{datapoint}s \cite{SemiSupervisedBook}. This approach is particularly useful 
		for ML applications that offer a large amount of unlabeled \gls{datapoint}s, but only a limited 
		number of  labeled \gls{datapoint}s.}, 
		first={semi-supervised learning (SSL)},text={SSL} }

\newglossaryentry{objfunc}{name={objective function}, description={An\index{objective function} 
		objective function is a map that assigns each value of an optimization variable, such 
		as the \gls{modelparams} $\weights$ of a \gls{hypothesis} $\hypothesis^{(\weights)}$, to 
		an objective value $f(\weights)$. The objective value $f(\weights)$ could be the 
		\gls{risk} or the \gls{emprisk} of a \gls{hypothesis} $\hypothesis^{(\weights)}$.},first={objective function},text={objective function} }
	
\newglossaryentry{regularizer}{name={regularizer}, description={A regularizer\index{regularizer} 
		assigns each \gls{hypothesis} $\hypothesis$ from a \gls{hypospace} $\hypospace$ a quantitative 
		measure $\regularizer{\hypothesis}$ for how much its prediction error on a \gls{trainset} might 
		differ from its prediction errors on \gls{datapoint}s outside the \gls{trainset}. \Gls{ridgeregression} 
		uses the regularizer $\regularizer{\hypothesis} \defeq \normgeneric{\weights}{2}^{2}$ for linear \gls{hypothesis} maps $\hypothesis^{(\weights)}(\featurevec) \defeq \weights^{T} \featurevec$ \cite[Ch. 3]{MLBasics}. 
		The \gls{lasso} uses the regularizer $\regularizer{\hypothesis} \defeq \normgeneric{\weights}{1}$ 
		for linear \gls{hypothesis} maps $\hypothesis^{(\weights)}(\featurevec) \defeq \weights^{T} \featurevec$ \cite[Ch. 3]{MLBasics}. },first={regularizer},text={regularizer} }

\newglossaryentry{regularization}{name={regularization}, description={Regularization\index{regularization} 
		techniques modify \gls{erm} such that the learnt \gls{hypothesis} performs well (generalizes) 
		also outside the \gls{trainset}. One specific implementation of \gls{regularization} is to add a penalty 
		or regularization term to the objective function of \gls{erm} (which is the average \gls{loss} on the \gls{trainset}). This regularization 
		term can be interpreted as an estimate for the increase in the expected \gls{loss} (\gls{risk}) compared to the 
		average \gls{loss} on the \gls{trainset}. },first={regularization},text={regularization} }

\newglossaryentry{rerm}{name={regularized empirical risk minimization}, 
	description={Synonym\index{regularized empirical risk minimization} for \gls{srm}.},
	first={regularized empirical risk minimization (RERM)},text={RERM} }
	
\newglossaryentry{gtv}{name={generalized total variation}, description={Generalized\index{generalized total variation} 
		total variation measures the changes of vector-valued node attributes over a 
		weighted undirected \gls{graph}.},first={generalized total variation (GTV)},text={GTV} }
	
\newglossaryentry{srm}{name={structural risk minimization}, description={Structural\index{structural risk minimization} 
		risk minimization is the problem of finding the \gls{hypothesis} that optimally 
		balances the average \gls{loss} (or \gls{emprisk}) on a \gls{trainset} with a 
		\gls{regularization} term. The \gls{regularization} term penalizes a \gls{hypothesis}
		that is not robust against (small) perturbations of the \gls{datapoint}s in the \gls{trainset}.},first={structural risk minimization (SRM)},text={SRM} }

\newglossaryentry{datapoisoning}{name={data poisoning}, description={Data\index{data poisoning} 
		poisoning refers to the intentional manipulation (or fabrication) of \gls{datapoint}s to 
		steer the training of a ML model \cite{Liu2021,PoisonGAN}. The protection against 
		data poisoning is particularly important in distributed ML applications where \gls{dataset}s are de-centralized.},first={data poisoning},text={data poisoning} }

\newglossaryentry{backdoor}{name={backdoor}, description={A\index{backdoor} backdoor attack refers 
		to the intentional manipulation of the training process underlying a ML method. This manipulation 
		can be implemented by perturbing the \gls{trainset} (data poisoning) or the 
		optimization algorithm used by an \gls{erm}-based method. The goal of a 
		backdoor attack is to nudge the learnt \gls{hypothesis} $\learnthypothesis$ 
		towards specific \gls{prediction}s for a certain range of \gls{feature} values. This range of \gls{feature} 
		values serves as a key (or trigger) to unlock a \emph{backdoor} in the sense of 
		delivering anomalous \gls{prediction}s. The key $\featurevec$ and the corresponding 
		anomolous \gls{prediction} $\learnthypothesis(\featurevec)$ are only known to the attacker.},
	first={backdoor},text={backdoor} }

\newglossaryentry{clustasspt}{name={clustering assumption}, description={The\index{clustering assumption} 
		clustering assumption postulates that \gls{datapoint}s in a \gls{dataset} form a (small) number of 
		groups or clusters. \Gls{datapoint}s in the same \gls{cluster} are more similar with each 
		other than with those outside the cluster \cite{SemiSupervisedBook}. We obtain different 
		clustering methods by using different notions of similarity between \gls{datapoint}s.},first={clustering assumption},text={clustering assumption} }
	
\newglossaryentry{dosattack}{name={denial-of-service attack}, description={A\index{denial-of-service attack} 
		denial-of-service attack aims (e.g., via \gls{datapoisoning}) to steer the training of a \gls{model} 
		such that it performs poorly for typical \gls{datapoint}s},
	first={denial-of-service attack},text={denial-of-service attack} }

\newglossaryentry{netexpfam}{name={networked exponential families}, 
	description={A\index{networked exponential families} collection of exponential 
		families, each of them assigned to a node of a \gls{empgraph}. The \gls{modelparams} are coupled 
	   via the network structure by requiring them to have a small \gls{gtv} \cite{JungNetExp2020}. },first={networked exponential family (nExpFam)},text={nExpFam} }

\newglossaryentry{scatterplot}{name={scatterplot}, description={A\index{scatterplot} 
		visualization technique that depicts data points by markers in a two-dimensional plane. 
		\begin{figure}[htbp]
			\begin{center}
				\begin{tikzpicture}[scale=0.9]
					\tikzset{x=1cm,y=1cm,every path/.style={>=latex},node style/.style={circle,draw}}
					\begin{axis}[axis x line=none,
						axis y line=none,
						ylabel near ticks,
						xlabel near ticks,
						enlarge y limits=true,
						xmin=-5, xmax=30,
						ymin=-5, ymax=30,
						width=5cm, height=5cm ]
						\addplot[only marks] table [x=mintmp, y=maxtmp, col sep = semicolon] {FMIData1.csv};
						\node at (axis cs:26,2) [anchor=west] {$\feature$};
						\node at (axis cs:0,30) [anchor=west] {$\truelabel$};
						\draw[->] (axis cs:-5,0) -- (axis cs:30,0);
						\draw[->] (axis cs:0,-5) -- (axis cs:0,30);
					\end{axis}
				\end{tikzpicture}
				\vspace*{-14mm}
			\end{center}
			\caption{A scatterplot of \gls{datapoint}s that represent daily weather conditions in Finland. 
				Each \gls{datapoint} is characterized by its minimum daytime temperature $\feature$ 
				as \gls{feature} and its maximum daytime temperature $\truelabel$ as the \gls{label}. 
				The temperatures have been measured at the \gls{fmi} weather station \emph{Helsinki Kaisaniemi} 
				during 1.9.2024 - 28.10.2024.}
			\label{fig_scatterplot_temp_FMI}
			\vspace*{-3mm}
			\end{figure}
		},first={scatterplot},text={scatterplot} }

\newglossaryentry{stepsize}{name={step size}, description={
		See\index{step size} \gls{learnrate}. 
}, 
	first={step size},text={step size} }

\newglossaryentry{learnrate}{name={learning rate}, description={Consider\index{learning rate} 
		an iterative method for finding or learning a useful \gls{hypothesis} $\hypothesis \in \hypospace$. 
		Such an iterative method repeats similar computational (update) steps that adjust or 
		modify the current \gls{hypothesis} to obtain an improved \gls{hypothesis}. A prime example of 
		such an iterative learning method is \gls{gd} and its variants such as \gls{stochGD} or \gls{projgd}. 
		We refer by learning rate to a parameter of an iterative learning 
		method that controls the extent by which the current \gls{hypothesis} 
		can be modified during a single iteration. A prime example of such a parameter is the 
				step size used in \gls{gd} \cite[Ch. 5]{MLBasics}.},
	first={learning rate},text={learning rate} }

\newglossaryentry{featuremap}{name={feature map}, description={A\index{feature map} map that transforms the original  
		 \gls{feature}s of a \gls{datapoint} into new \gls{feature}s. The so-obtained new \gls{feature}s might 
		 be preferable over the original features for several reasons. For example, the shape of datasets might 
		 become simpler in the new feature space, allowing to use linear models in the new \gls{feature}s. 
		 Another reason could be that the number of new \gls{feature}s is much smaller which is preferable 
		 in terms of avoiding \gls{overfitting}. The special case of a \gls{feature} map delivering 
		 two numeric \gls{feature}s is particularly useful for data visualization. Indeed, we can depict 
		 \gls{datapoint}s in a \gls{scatterplot} by using two \gls{feature}s as the coordinates of a \gls{datapoint}.},
	first={feature map},text={feature map} }
 
  \newglossaryentry{lasso}{name={least absolute shrinkage and selection operator (Lasso)}, 
	description={The\index{Lasso} least absolute shrinkage and selection operator (Lasso) is an 
		instance of \gls{srm} to learn the weights $\weights$ of a linear map 
		$\hypothesis(\featurevec) = \weights^{T} \featurevec$ based on a \gls{trainset}. 
		The Lasso is obtained from \gls{linreg} by adding the scaled $\ell_{1}$-norm 
		$\regparam \normgeneric{\weights}{1}$ to the average \gls{sqerrloss} incurred on the \gls{trainset}. 
	},
	first={ least absolute shrinkage and selection operator (Lasso)},text={Lasso} }
 
 \newglossaryentry{simgraph}{name={similarity graph}, 
 	description={Some\index{similarity graph} ML applications generate \gls{datapoint}s that 
 		are related by a domain-specific notion of similarity. These similarities can be 
 		represented conveniently using a similarity \gls{graph} $\graph = \big(\nodes \defeq \{1,\ldots,\samplesize\},\edges\big)$. 
 		The node $\sampleidx \in \nodes$ represents the $\sampleidx$-th \gls{datapoint}. Two 
 		nodes are connected by an undirected edge if the corresponding \gls{datapoint}s are similar. 
 	},
 	first={similarity graph},text={similarity graph} }

 \newglossaryentry{kld}{name={Kullback-Leibler divergence}, 
 	description={
 		 The\index{KL divergence} Kullback–Leibler divergence is a quantitative 
 		 measure for how much one \gls{probdist} is different from another \gls{probdist} \cite{coverthomas}.  
 	},
 	first={Kullback-Leibler divergence},text={KL divergence} }

\newglossaryentry{LapMat}{name={Laplacian matrix}, 
	description={
		The\index{Laplacian matrix} geometry or structure of a graph $\graph$ can be 
		analyzed using the properties of special matrices that are associated with $\graph$. 
		One such matrix is the graph Laplacian matrix $\mL$ which is defined for an 
		undirected and weighted \gls{graph} \cite{Luxburg2007,Ng2001}. One important 
		example of such a \gls{graph} is the \gls{empgraph} in a \gls{fl} application. 
	},
	first={Laplacian matrix},text={Laplacian matrix} }

\newglossaryentry{kernel}{name={kernel}, 
	description={Consider\index{kernel} \gls{datapoint}s characterized by a \gls{feature} vector $\featurevec \in \featurespace$ 
	with a generic \gls{featurespace} $\featurespace$. A kernel is a map that assigns each pair of \gls{feature} vectors  
	$\featurevec, \featurevec' \in \featurespace$ a real number. This number measures the similarity between 
	$\featurevec$ and $\featurevec'$. For more details about kernels and the resulting kernel methods, 
	we refer to the literature \cite{LampertNowKernel,LearningKernelsBook}.},
	first={kernel},text={kernel} }

\newglossaryentry{cm}{name={confusion matrix}, 
	description={Consider\index{confusion matrix} \gls{datapoint}s characterized by \gls{feature}s $\featurevec$ 
		and \gls{label} $\truelabel$ having values from the finite \gls{labelspace} $\labelspace = \{1,\ldots,\nrcluster\}$. 
		The confusion matrix is $\nrcluster \times \nrcluster$ matrix with rows representing different values $\clusteridx$ 
		of the true label of a \gls{datapoint}. The columns of a confusion matrix correspond to different values 
		$\clusteridx'$ delivered by a hypothesis $\hypothesis(\featurevec)$. The $(\clusteridx,\clusteridx')$-th entry of 
		the confusion matrix is the fraction of \gls{datapoint}s with \gls{label} $\truelabel\!=\! \clusteridx$ and the 
		\gls{prediction} $\hat{\truelabel}\!=\!\clusteridx'$ assigned by the \gls{hypothesis} $\hypothesis$.},
	first={confusion matrix},text={confusion matrix} }

\newglossaryentry{featurematrix}{name={feature matrix}, 
	description={Consider\index{feature matrix} a \gls{dataset} $\dataset$ with $\samplesize$ \gls{datapoint}s, each of them 
		characterized by the \gls{feature}s $\featurevec^{(\sampleidx)}$, $\sampleidx=1,\ldots,\samplesize$. 
		The feature matrix $\featuremtx$ of $\dataset$ is constructed by stacking, column-wise, 
	the \gls{feature}s of the \gls{datapoint}s into a matrix $\featuremtx = \big( \featurevec^{(1)},\ldots, \featurevec^{(\samplesize)} \big)$ 
	of size $\samplesize \times \nrfeatures$.} 
	first={feature matrix},text={feature matrix} }

\newglossaryentry{dbscan}{name={density-based spatial clustering of applications with noise}, 
	description={A\index{DBSCAN} clustering algorithm for \gls{datapoint}s that are characterized by numeric feature vectors. 
		Like \gls{kmeans} and \gls{softclustering} via \gls{gmm}, also DBSCAN uses the Euclidean 
		distances between \gls{feature} vectors to determine the \gls{cluster}s. However, in contrast to \gls{kmeans} 
		and \gls{gmm}, DBSCAN uses a different notion of similarity between \gls{datapoint}s. 
		DBSCAN considers two \gls{datapoint}s as similar if they are \emph{connected} 
		via a sequence (path) of close-by intermediate \gls{datapoint}s. Thus, DBSCAN might consider 
		two \gls{datapoint}s as similar (and therefore belonging to the same cluster) even if 
		their \gls{feature} vectors have a large Euclidean distance.},
	first={density-based spatial clustering of applications with noise (DBSCAN)},text={DBSCAN} }

\newglossaryentry{fl}{name={federated learning (FL)}, description={Federated\index{federated learning} 
		learning is an umbrella term for ML methods that train models in a collaborative 
		fashion using decentralized data and computation.},first={federated learning (FL)},text={FL} }
	
\newglossaryentry{cfl}{name={clustered federated learning (CFL)}, description={
		Clustered\index{clustered federated learning} \gls{fl} (CFL) assumes that \gls{localdataset}s form clusters. 
		The \gls{localdataset}s belonging to the same cluster have similar statistical properties. 
		CFL pools \gls{localdataset}s in the same cluster to obtain a \gls{trainset} 
		for training a cluster-specific \gls{model}. \Gls{gtvmin} implements this pooling implicitly 
		by forcing the local \gls{modelparams} to be approximately identical over well-connected 
		subsets of the \gls{empgraph}.},
	first={clustered \gls{fl}},text={CFL} }

\newglossaryentry{iid}{name={i.i.d.}, description={It\index{i.i.d.} can be useful to 
		interpret \gls{datapoint}s $\datapoint^{(1)},\ldots,\datapoint^{(\samplesize)}$ 
		as \gls{realization}s of independent and identically distributed \gls{rv}s with 
		a common \gls{probdist}. If these \gls{rv}s are continuous-valued, their joint \gls{pdf} is $p\big(\datapoint^{(1)},\ldots,\datapoint^{(\samplesize)} \big) = \prod_{\sampleidx=1}^{\samplesize} p \big(\datapoint^{(\sampleidx)}\big)$ with $p(\datapoint)$ being the common 
		marginal \gls{pdf} of the underlying \gls{rv}s.},
	first={independent and identically distributed (i.i.d.)},text={{i.i.d.}} }

\newglossaryentry{outlier}{name={outlier}, description={Many\index{outlier} ML methods 
		are motivated by the \gls{iidasspt} which interprets \gls{datapoint}s as realizations of 
		\gls{iid} \gls{rv}s with a common \gls{probdist}. The \gls{iidasspt} is useful for applications  
		where the statistical properties of the data generation process are stationary (or time-invariant) \cite{Brockwell91}. 
		However, in some applications the data consists of a majority of \emph{regular} \gls{datapoint}s 
		that conform with an \gls{iidasspt} and a small number of data points that have fundamentally different 
        statistical properties compared to the regular \gls{datapoint}s. We refer to a \gls{datapoint} that 
        substantially deviates from the statistical properties of most \gls{datapoint}s as an 
        outlier. Different methods for outlier detection use different measures for this deviation. 
        Stastistical learning theory studies fundamental limits on the ability to mitigate outliers reliably \cite{doi:10.1137/0222052,10.1214/20-AOS1961}.},
	          first={outlier},text={outlier} }

\newglossaryentry{decisionregion}{name={decision region}, description={Consider\index{decision region} 
		a \gls{hypothesis} map $\hypothesis$ that delivers values from a finite set $\labelspace$. 
		We refer to the set of \gls{feature}s $\featurevec \in \featurespace$ that result 
		in the same output $\hypothesis(\featurevec)=a$ as a decision region of 
		the \gls{hypothesis} $\hypothesis$. },first={decision region},text={decision region} }

\newglossaryentry{decisionboundary}{name={decision region}, description={Consider\index{decision boundary} a 
		\gls{hypothesis} map $\hypothesis$ that reads in a \gls{feature} vector 
		$\featurevec \in \mathbb{R}^{\featuredim}$ and delivers a value from a finite set $\labelspace$. 
		The decision boundary of $\hypothesis$ is the set of vectors $\featurevec \in \mathbb{R}^{\featuredim}$ 
		that lie between different \gls{decisionregion}s. More precisely, a 
		vector $\featurevec$ belongs to the decision boundary if and only 
		if each neighbourhood $\{ \featurevec': \| \featurevec - \featurevec' \| \leq \varepsilon \}$, 
		for any $\varepsilon >0$, contains at least two vectors with different function values.},first={decision boundary},text={decision boundary} }

\newglossaryentry{euclidspace}{name={Euclidean space}, description={The\index{Euclidean space} 
		Euclidean space $\mathbb{R}^{\featuredim}$ of dimension $\featuredim \in \mathbb{N}$ consists 
		of vectors $\featurevec= \big(\feature_{1},\ldots,\feature_{\featurelen}\big)$, with $\featuredim$ 
		real-valued entries $\feature_{1},\ldots,\feature_{\featuredim} \in \mathbb{R}$. Such an Euclidean 
		space is equipped with a geometric structure defined by the inner product 
		$\featurevec^{T} \featurevec' = \sum_{\featureidx=1}^{\featuredim} \feature_{\featureidx} \feature'_{\featureidx}$ 
		between any two vectors $\featurevec,\featurevec' \in \mathbb{R}^{\featuredim}$ \cite{RudinBookPrinciplesMatheAnalysis}.},first={Euclidean space},text={Euclidean space} }

\newglossaryentry{eerm}{name={explainable empirical risk minimization}, description={An\index{explainable empirical risk minimization} 
		instance of structural risk minimization that adds a \gls{regularization} term to the 
		average \gls{loss} in the objective function of \gls{erm}. 
		The \gls{regularization} term is chosen to favour \gls{hypothesis} maps that are intrinsically 
		explainable for a specific user. This user is characterized by their \gls{prediction}s provided 
		for the \gls{datapoint}s in a \gls{trainset} \cite{Zhang:2024aa}.},first={explainable empirical risk minimization (EERM)},text={EERM} }

\newglossaryentry{kmeans}{name={$k$-means}, description={The\index{$k$-means} $k$-means algorithm 
		is a hard \gls{clustering} method which assigns each \gls{datapoint} of a \gls{dataset} 
		to precisely one of $k$ different \gls{cluster}s. The method alternates between updating 
		the \gls{cluster} assignments (to the cluster with nearest cluster mean) and, given the updated \gls{cluster} assignments, 
		re-calculating the cluster means \cite[Ch. 8]{MLBasics}.},first={$k$-means},text={$k$-means} }

\newglossaryentry{xml}{name={explainable ML}, description={Explainable\index{explainable AI} 
		ML methods aim at complementing each \gls{prediction} with an \gls{explanation} for how the \gls{prediction} 
		has been obtained. The construction of an explicit \gls{explanation} might not be necessary 
		if the ML method uses a sufficiently simple (or interpretable) \gls{model} \cite{rudin2019stop}.},first={explainable ML},text={explainable ML} }

\newglossaryentry{fmi}{name={Finnish Meteorological Institute}, description={The\index{Finnish Meteorological Institute}
		Finnish Meteorological Institute is a government agency responsible for gathering 
		and reporting weather data in Finland.},first={Finnish Meteorological Institute (FMI)},text={FMI} }
	
\newglossaryentry{samplemean}{name={sample mean}, description={The\index{sample mean} sample mean 
			$\vm \in \mathbb{R}^{\nrfeatures \times \nrfeatures}$ for a given set of \gls{feature} 
			vectors $\featurevec^{(1)},\ldots,\featurevec^{(\samplesize)} \in \mathbb{R}^{\nrfeatures}$ 
			is defined as 
			$$\vm = (1/\samplesize) \sum_{\sampleidx=1}^{\samplesize} \featurevec^{(\sampleidx)}.$$ 
		},
		first={sample mean},text={sample mean} }
	
\newglossaryentry{samplecovmtx}{name={sample covariance matrix}, description={The\index{sample covariance matrix} 
		sample covariance matrix $\widehat{\bf \Sigma} \in \mathbb{R}^{\nrfeatures \times \nrfeatures}$ 
		for a given set of \gls{feature} vectors $\featurevec^{(1)},\ldots,\featurevec^{(\samplesize)} \in \mathbb{R}^{\nrfeatures}$ is defined as 
		$$\widehat{\bf \Sigma} = (1/\samplesize) \sum_{\sampleidx=1}^{\samplesize} (\featurevec^{(\sampleidx)}\!-\!\widehat{\vm}) (\featurevec^{(\sampleidx)}\!-\!\widehat{\vm})^{T}.$$ 
		Here, we used the \gls{samplemean} $\widehat{\vm}$. 
	},
	first={sample covariance matrix},text={sample covariance matrix} }

\newglossaryentry{covmtx}{name={covariance matrix}, description={The\index{covariance matrix} covariance matrix of 
		a \gls{rv} $\vx \in \mathbb{R}^{\featuredim}$ is defined as $\expect \bigg \{ \big( \vx - \expect \big\{ \vx \big\} \big)  \big(\vx - \expect \big\{ \vx \big\} \big)^{T} \bigg\}$.},
	first={covariance matrix},text={covariance matrix} }
	
\newglossaryentry{highdimregime}{name={high-dimensional regime}, description={The\index{high-dimensional regime} 
		high-dimensional regime of \gls{erm} is characterized by the \gls{effdim} of the \gls{model} 
		being larger than the \gls{samplesize}, i.e., the number of (labeled) \gls{datapoint}s in the \gls{trainset}. 
		For example, \gls{linreg} methods operate in the high-dimensional regime whenever the number $\featuredim$ of \gls{feature}s 
		used to characterize \gls{datapoint}s exceeds the number of \gls{datapoint}s in the \gls{trainset}. 
		Another example of ML methods that operate in the high-dimensional regime are large \gls{ann}s, having 
		far more tunable weights (and bias terms) than the number of \gls{datapoint}s in the \gls{trainset}. 
		High-dimensional statistics is a recent main thread of probability theory that studies the 
		behavior of ML methods in the high-dimensional regime \cite{Wain2019,BuhlGeerBook}.},
   first={high-dimensional regime},text={high-dimensional regime} }

\newglossaryentry{gmm}{name={Gaussian mixture model}, description={Gaussian\index{Gaussian mixture model} mixture models (GMM) are a family of 
		\gls{probmodel}s for \gls{datapoint}s characterized by a numeric \gls{feature} vector $\featurevec$. A GMM interprets 
	     $\featurevec$ as being drawn from one of $\nrcluster$ different \gls{mvndist}s $p^{(\clusteridx)} = \mvnormal{\meanvec{\clusteridx}}{\covmtx{\clusteridx}}$, indexed by $\clusteridx=1,\ldots,\nrcluster$. 
	     The probability that $\featurevec$ is drawn from the $\clusteridx$-th \gls{mvndist} is 
	     denoted $p_{\clusteridx}$. Thus, a GMM is parametrized by the probability $p_{\clusteridx}$, the 
		mean vector $\clustermean^{(\clusteridx)}$ and \gls{covmtx} $\clustercov^{(\clusteridx)}$ for each $\clusteridx=1,\ldots,\nrcluster$. 
	 },first={Gaussian mixture model (GMM)},text={GMM} }
 
\newglossaryentry{ml}{name={maximum likelihood}, description={
		Consider\index{maximum likelihood} \gls{datapoint}s $\dataset=\big\{ \datapoint^{(1)}, \ldots, \datapoint^{(\samplesize)} \}$ that are interpreted 
		as realizations of \gls{iid} \gls{rv}s with a common \gls{probdist} $\prob{\datapoint; \weights}$ which 
		depends on a parameter vector $\weights \in \mathcal{W} \subseteq \mathbb{R}^{n}$. Maximum likelihood methods aim at finding a parameter vector $\weights$ 
		such that the probability (density) $\prob{\dataset; \weights} = \prod_{\sampleidx=1}^{\samplesize} \prob{\datapoint^{(\sampleidx)}; \weights}$ 
		of observing the data is maximized. Thus, the maximum 
		likelihood estimator is obtained as a solution to the optimization 
		problem $\max_{\weights \in \mathcal{W}} \prob{\dataset; \weights}$.
	},first={maximum likelihood},text={maximum likelihood}}

\newglossaryentry{em}{name={expectation maximization}, description={Expectation\index{expectation maximization} 
		maximization is a generic technique for estimating the \gls{modelparams} of a \gls{probmodel} $\prob{\datapoint; \weights}$ 
		from data \cite{BishopBook,hastie01statisticallearning,GraphModExpFamVarInfWainJor}. Expectation maximization 
		delivers an approximation to the \gls{ml} estimate for the model parameters $\weights$. 
  },first={expectation maximization (EM)},text={EM}}

\newglossaryentry{ppca}{name={probabilistic PCA}, description={Probabilistic\index{probabilistic PCA} \gls{pca} (PPCA) 
		extends basic \gls{pca} by using a \gls{probmodel} for \gls{datapoint}s. The \gls{probmodel} of PPCA 
		reduces the task of dimensionality reduction to an estimation problem that can be solved using \gls{em} 
		methods.},first={probabilistic PCA (PPCA)},text={PPCA}}
	
\newglossaryentry{polyreg}{name={polynomial regression}, description={Polynomial\index{polynomial regression} 
		regression aims at learning a polynomial \gls{hypothesis} map to predict a numeric \gls{label} based
		 on numeric \gls{feature}s of a \gls{datapoint}. For \gls{datapoint}s characterized by a single 
		 numeric \gls{feature}, polynomial regression uses the \gls{hypospace} 
			$\hypospace^{(\rm poly)}_{\nrfeatures} \defeq \{ \hypothesis(x) = \sum_{\featureidx=0}^{\nrfeatures-1} x^{\featureidx} \weight_{\featureidx} \}.$
			The quality of a polynomial \gls{hypothesis} map is measured using the average \gls{sqerrloss} 
			incurred on a set of labeled \gls{datapoint}s (which we refer to as \gls{trainset}).},first={polynomial regression},text={polynomial regression}}

\newglossaryentry{linreg}{name={linear regression}, description={Linear\index{linear regression} 
		regression aims to learn a linear \gls{hypothesis} map to predict a numeric \gls{label} based 
		on numeric \gls{feature}s of a \gls{datapoint}. The quality of a linear \gls{hypothesis} map is 
		measured using the average \gls{sqerrloss} incurred on a set of labeled \gls{datapoint}s, 
		which we refer to as the \gls{trainset}.},first={linear regression},text={linear regression}}
        
\newglossaryentry{ridgeregression}{name={ridge regression}, description={Ridge\index{ridge regression} 
		regression learns the \gls{weights} $\weights$ of a linear \gls{hypothesis} map $\hypothesis^{(\weights)}(\featurevec)= \weights^{T} \featurevec$. 
		The quality of a particular choice for the parameter vector $\weights$ is measured by the sum 
		of two components. The first component is the average \gls{sqerrloss} incurred by $\hypothesis^{(\weights)}$ on a set of 
		labeled \gls{datapoint}s (the \gls{trainset}). The second component is the scaled squared 
		Euclidean norm $\regparam \| \weights \|^{2}_{2}$ with a \gls{regularization} parameter 
		$\regparam > 0$. It can be shown that the effect of adding to $\regparam \| \weights \|^{2}_{2}$ to 
	the average \gls{sqerrloss} is equivalent to replacing the original \gls{datapoint}s by an ensemble of 
realizations of a \gls{rv} centered around these \gls{datapoint}s.},first={ridge regression},text={ridge regression}}

\newglossaryentry{expectation}{name={expectation}, description={
		Consider\index{expectation} a numeric \gls{feature} vector $\featurevec \in \mathbb{R}^{\featuredim}$ 
		which we interpret as the \gls{realization} of a \gls{rv} with \gls{probdist} $p(\featurevec)$. 
		The expectation of $\featurevec$ is defined as the integral $\expect \{ \featurevec \} \defeq \int \featurevec p(\featurevec)$ \cite{HalmosMeasure,BillingsleyProbMeasure,RudinBookPrinciplesMatheAnalysis}. Note that 
		the expectation is only defined if this integral exists, i.e., if the \gls{rv} is integrable.},first={expectation},text={expectation}}

\newglossaryentry{logreg}{name={logistic regression}, description={Logistic\index{logistic regression} regression learns a 
		linear \gls{hypothesis} map (\gls{classifier}) $\hypothesis(\featurevec) = \weights^{T} \featurevec$ 
		to predict a binary \gls{label} $\truelabel$ based on numeric \gls{feature} vector 
		$\featurevec$ of a \gls{datapoint}. The quality of a linear \gls{hypothesis} map is 
		measured by the average \gls{logloss} on some labeled \gls{datapoint}s (the \gls{trainset}).},
		first={logistic regression},text={logistic regression}}
	
\newglossaryentry{logloss}{name={logistic loss}, description={Consider\index{logistic loss} 
		a \gls{datapoint}, characterized by the \gls{feature}s $\featurevec$ and a binary \gls{label} $\truelabel \in \{-1,1\}$. 
		We use a real-valued \gls{hypothesis} $\hypothesis$ to predict the label $\truelabel$ 
		from the features $\featurevec$. The logistic loss incurred by this \gls{prediction} is 
		defined as 
	\begin{equation} 
		\label{equ_log_loss_gls}
		\lossfunc{(\featurevec,\truelabel)}{\hypothesis} \defeq  \log ( 1 + \exp(- \truelabel \hypothesis(\featurevec))).
\end{equation}
Carefully note that the expression \eqref{equ_log_loss_gls} 
for the logistic loss applies only if for the \gls{labelspace} $\labelspace = \{ -1,1\}$ and using 
the thresholding rule \eqref{equ_def_threshold_bin_classifier}. },first={logistic loss},text={logistic loss}}
	
\newglossaryentry{hingeloss}{name={hinge loss}, description={Consider\index{hinge loss} a \gls{datapoint}, 
		characterized by a \gls{feature} vector $\featurevec \in \mathbb{R}^{\featuredim}$ and a 
		binary \gls{label} $\truelabel \in \{-1,1\}$. The hinge loss incurred by a real-valued 
		\gls{hypothesis} map $\hypothesis(\featurevec)$ is defined as 
		\begin{equation} 
			\label{equ_hinge_loss_gls}
				\lossfunc{(\featurevec,\truelabel)}{\hypothesis} \defeq \max \{ 0 , 1 - \truelabel \hypothesis(\featurevec) \}. 
			\end{equation}
	    A regularized variant of the hinge loss is used by the \gls{svm} \cite{LampertNowKernel} to learn 
	    a \gls{linclass} with maximum margin between the two classes (see Figure \ref{fig_svm_gls}). 
		\begin{figure}[htbp]
			\begin{center}
				\begin{tikzpicture}[auto,scale=0.8]
					\draw [thick] (1,2) circle (0.1cm)node[anchor=west] {\hspace*{0mm}$\featurevec^{(5)}$};
					\draw [thick] (0,1.6) circle (0.1cm)node[anchor=west] {\hspace*{0mm}$\featurevec^{(4)}$};
					\draw [thick] (0,3) circle (0.1cm)node[anchor=west] {\hspace*{0mm}$\featurevec^{(3)}$};
					\draw [thick] (2,1) circle (0.1cm)node[anchor=east,above] {\hspace*{0mm}$\featurevec^{(6)}$};
					\node[] (B) at (-2,0) {\emph{support vector}};
					\draw[->,dashed] (B) to (1.9,1) ; 
					\draw [|<->|,thick] (2.05,0.95)  -- (2.75,0.25)node[pos=0.5] {$\xi$} ; 
					\draw [thick] (1,-1.5) -- (4,1.5) node [right] {$\hypothesis^{(\weights)}$} ; 
					\draw [thick] (3,-1.9) rectangle ++(0.1cm,0.1cm) node[anchor=west,above]  {\hspace*{0mm}$\featurevec^{(2)}$};
					\draw [thick] (4,.-1) rectangle ++(0.1cm,0.1cm) node[anchor=west,above] {\hspace*{0mm}$\featurevec^{(1)}$};
				\end{tikzpicture}
				\caption{The \gls{svm} learns a hypothesis (or classifier) $\hypothesis^{(\weights)}$ with 
					minimum average soft-margin \gls{hingeloss}. Minimizing this \gls{loss} is equivalent 
					to maximizing the margin $\xi$ between the \gls{decisionboundary} of $\hypothesis^{(\weights)}$ 
					and each class of the \gls{trainset}.}
				\label{fig_svm_gls}
			\end{center}
		\end{figure}
	},first={hinge loss},text={hinge loss}}

\newglossaryentry{iidasspt}{name={i.i.d.\ assumption}, description={The i.i.d.\ assumption\index{i.i.d.} interprets \gls{datapoint}s of a \gls{dataset} 
		as the realizations of \gls{iid} \gls{rv}s.},first={i.i.d.\ assumption},text={i.i.d.\ assumption} }

\newglossaryentry{hypospace}{name={hypothesis space}, description={Every\index{hypothesis space} 
		practical ML method uses a hypothesis space (or \gls{model}) $\hypospace$. The hypothesis space of a ML 
		method is a subset of all possible maps from the \gls{featurespace} to \gls{labelspace}. The design 
		choice of the hypothesis space should take into account available computational resources and 
		statistical aspects. If the computational infrastructure allows for efficient matrix operations, and there 
		is a (approximately) linear relation between \gls{feature}s and \gls{label}, a useful choice for the 
		hypothesis space might be the \gls{linmodel}.},first={hypothesis space},text={hypothesis space} }
	
\newglossaryentry{model}{name={model}, description={In the context of ML methods, 
		the term \emph{model} typically refers to the \gls{hypospace} used by a ML method \cite{ShalevMLBook,MLBasics}.},first={model},text={model} }

\newglossaryentry{modelparams}{name={model parameters}, 
	description={Model parameters are numbers that select a \gls{hypothesis} map from a \gls{hypospace}.},
	first={model parameters},text={model parameters} }

\newglossaryentry{ai}{name={artificial intelligence}, description={Artificial intelligence\index{artificial intelligence} 
		aims to develop systems that behave rational in the sense of maximizing a long-term reward. 
		In MLterms, these systems train a \gls{model} to predict optimal actions in order to 
		minimize a \gls{loss} computed from reward signals.},first={artificial intelligence (AI)},text={AI} }
	
\newglossaryentry{hardclustering}{name={hard clustering}, description={Hard clustering\index{hard clustering} 
		refers to the task of partitioning a given set of \gls{datapoint}s into (few) non-overlapping \gls{cluster}s. 
		Each \gls{datapoint} is assigned to one \gls{cluster}.},first={hard clustering},text={hard clustering} }
	
\newglossaryentry{softclustering}{name={soft clustering}, description={Soft clustering\index{soft clustering} 
		refers to the task of partitioning a given set of \gls{datapoint}s into (few) overlapping clusters. 
		Each \gls{datapoint} is assigned to several different clusters with varying \gls{dob}. Soft clustering 
		methods determine the \gls{dob} (or soft cluster assignment) for each \gls{datapoint} and each \gls{cluster}.
		A principled approach to soft clustering is by interpreting \gls{datapoint}s as \gls{iid} realizations of a 
		\gls{gmm}. We then obtain a natural choice for the \gls{dob} as the conditional probability of a \gls{datapoint} 
		belonging to a specific mixture component.},first={soft clustering},text={soft clustering} }
	
\newglossaryentry{clustering}{name={clustering}, description={Clustering\index{clustering} methods decompose a given 
		set of \gls{datapoint}s into few subsets, which are referred to as \gls{cluster}s. 
		Each \gls{cluster} consists of \gls{datapoint}s that are more similar to each 
		other than to \gls{datapoint}s outside the \gls{cluster}. Different clustering methods 
		use different measures for the similarity between \gls{datapoint}s and different 
		forms of \gls{cluster} representations. The clustering method \gls{kmeans} uses the 
		average \gls{feature} vector (\emph{cluster mean}) of a \gls{cluster} as its representative. 
		A popular \gls{softclustering} method based on \gls{gmm} represents 
		a \gls{cluster} by a \gls{mvndist}.},first={clustering},text={clustering} }
	
\newglossaryentry{cluster}{name={cluster}, description={A\index{cluster} \gls{cluster} is a subset of 
		\gls{datapoint}s that are more similar to each other than to the \gls{datapoint}s outside the \gls{cluster}. 
		The quantitative measure of similarity between \gls{datapoint}s is a design choice. If \gls{datapoint}s 
		are characterized by Euclidean \gls{feature} vectors $\featurevec \in \mathbb{R}^{\nrfeatures}$, 
		we can define the similarity between two \gls{datapoint}s via the Euclidean distance between 
		their \gls{feature} vectors.},first={cluster},text={cluster} }

\newglossaryentry{huberloss}{name={Huber loss}, description={The\index{Huber loss} 
		Huber \gls{loss} is a generalization and combination of the \gls{sqerrloss} and 
		the absolute error \gls{loss}.},first={Huber loss},text={Huber loss} }

\newglossaryentry{svm}{name={support vector machine}, description={The\index{support vector machine} 
		support vector machine is a binary classification method that learns a linear \gls{hypothesis} map. 
		It aims at maximally separating \gls{datapoint}s from the two different classes in the \gls{feature} 
		space (\emph{maximum margin principle}). Maximizing this separation is equivalent to minimizing a regularized variant of 
		the \gls{hingeloss} \eqref{equ_hinge_loss_gls}.},first={support vector machine (SVM)},text={SVM} }

\newglossaryentry{eigenvalue}{name={eigenvalue}, description={We refer to a number $\lambda \in \mathbb{R}$ 
		as eigenvalue of a square matrix $\mathbf{A} \in \mathbb{R}^{\featuredim \times \featuredim}$ if there is a 
		non-zero vector $\vx \in \mathbb{R}^{\featuredim} \setminus \{ \mathbf{0} \}$ such that $\mathbf{A} \vx = \lambda \vx$. },first={eigenvalue},text={eigenvalue} }
	
\newglossaryentry{eigenvector}{name={eigenvector}, description={An\index{eigenvector} 
		eigenvector of a matrix $\mathbf{A} \in \mathbb{R}^{\featuredim \times \featuredim}$ 
		is a non-zero vector $\vx \in \mathbb{R}^{\featuredim} \setminus \{ \mathbf{0} \}$ 
		such that $\mathbf{A} \vx = \lambda \vx$ with some \gls{eigenvalue} $\lambda$.},first={eigenvector},text={eigenvector} }

\newglossaryentry{evd}{name={eigenvalue decomposition}, 
	description={The\index{eigenvalue decomposition} \gls{eigenvalue} 
		decomposition for a square matrix $\mA \in \mathbb{R}^{\dimlocalmodel \times \dimlocalmodel}$ 
		is a factorization of the form 
		$$\mA = \mathbf{V} {\bm \Lambda} \mathbf{V}^{-1}.$$ 
		The columns of the matrix $\mV = \big( \vv^{(1)},\ldots,\vv^{(\dimlocalmodel)} \big)$ are the 
		\gls{eigenvector}s of the matrix $\mV$. The diagonal matrix 
		${\bm \Lambda} = {\rm diag} \big\{ \eigval{1},\ldots,\eigval{\dimlocalmodel} \big\}$ 
		contains the \gls{eigenvalue}s $\eigval{\featureidx}$ corresponding to the \gls{eigenvector}s $\vv^{(\featureidx)}$. 
		Note that the above decomposition exists only if the matrix $\mA$ is diagonalizable.},first={eigenvalue decomposition (EVD)},text={EVD} }

\newglossaryentry{svd}{name={singular value decomposition}, 
  	description={The\index{singular value decomposition} singular value 
  		decomposition for a matrix 
		$\mA \in \mathbb{R}^{\samplesize \times \dimlocalmodel}$ 
		is a factorization of the form 
		$$\mA = \mathbf{V} {\bm \Lambda} \mathbf{U}^{T},$$ 
		with orthonormal matrices $\mV \in \mathbb{R}^{\samplesize \times \samplesize}$ 
		and $\mU \in \mathbb{R}^{\dimlocalmodel \times \dimlocalmodel}$ \cite{GolubVanLoanBook}. 
		The matrix ${\bm \Lambda} \in \mathbb{R}^{\samplesize \times \dimlocalmodel}$ is 
		only non-zero along the main diagonal, whose entries $\Lambda_{\featureidx,\featureidx}$ 
		are non-negative and referred to as singular values.
	},first={singular value decomposition (SVD)},text={SVD} }

\newglossaryentry{tv}{name={total variation}, description={See \gls{gtv}.},
	first={total variation},text={total variation} }

\newglossaryentry{gdmethods}{name={gradient-based method}, description={Gradient-based\index{gradient-based methods} 
		methods are iterative techniques for finding the minimum (or maximum) 
		of a \gls{differentiable} objective function of the \gls{modelparams}. These 
		methods construct a sequence of approximations to an optimal choice for 
		\gls{modelparams} that results in a minimum objective function value. 
		As their name indicates, gradient-based methods use the \gls{gradient}s of the objective function 
		evaluated during previous iterations to construct new (hopefully) improved \gls{modelparams}.},
		first={gradient-based methods},text={gradient-based methods} }

\newglossaryentry{sgd}{name={subgradient descent}, description={Subgradient\index{subgradient descent} 
		descent is a generalization of \gls{gd} that does not require differentiability of the 
		function to be minimized. This generalization is obtained by replacing the concept 
		of a \gls{gradient} with that of a sub-gradient. Similar to \gls{gradient}s, also sub-gradients 
		allow to construct local approximations of an objective function. The objective function 
		might be the \gls{emprisk} $\emperror\big( \hypothesis^{(\weights)} \big| \dataset \big)$ viewed 
		as a function of the \gls{modelparams} $\weights$ that select a \gls{hypothesis} $\hypothesis^{(\weights)} \in \hypospace$.},first={subgradient descent},text={subgradient descent} }
	
\newglossaryentry{stochGD}{name={stochastic gradient descent}, description={Stochastic\index{stochastic gradient descent} \gls{gd} is 
		obtained from \gls{gd} by replacing the \gls{gradient} of the objective function with some e
		stimate (or approximation). A main application of stochastic gradient descent is to solve \gls{erm} 
		where the objective function and its gradient consists of a sum over the \gls{datapoint}s in a \gls{dataset} $\dataset$. 
		Here, the gradient estimate can be obtained by replacing the sum over the entire \gls{dataset} with 
		a sum over a randomly selected subset of $\dataset$. },first={stochastic gradient descent (SGD)},text={SGD} }

\newglossaryentry{pca}{name={principal component analysis (PCA)}, description={Principal\index{principal component analysis} 
		component analysis determines a linear \gls{featuremap} such that the new \gls{feature}s allow to reconstruct 
		the original \gls{feature}s with minimum reconstruction error \cite{MLBasics}.},first={principal component analysis (PCA)},text={PCA} }
	
\newglossaryentry{loss}{name={loss}, description={ML\index{loss} methods use a 
		\gls{lossfunc} $\lossfunc{\datapoint}{\hypothesis}$ to measure the error incurred 
		by applying a specific \gls{hypothesis} to a specific \gls{datapoint}. With 
		slight abuse of notation, we use the term \emph{loss} for both, the \gls{lossfunc} $\loss$ 
		itself and for its value $\lossfunc{\datapoint}{\hypothesis}$ for a specific \gls{datapoint} $\datapoint$ 
		and \gls{hypothesis} $\hypothesis$.},first={loss},text={loss} }

\newglossaryentry{lossfunc}{name={loss function}, description={A\index{loss function} loss function is a map 
		$$\lossfun: \featurespace \times \labelspace \times \hypospace \rightarrow \mathbb{R}_{+}: \big( \big(\featurevec,\truelabel\big), \hypothesis\big) \mapsto  \lossfunc{(\featurevec,\truelabel)}{\hypothesis}$$ which assigns a pair 
		of a \gls{datapoint}, with features $\featurevec$ 
		and label $\truelabel$, and a \gls{hypothesis} $\hypothesis \in \hypospace$ the 
		non-negative real number $\lossfunc{(\featurevec,\truelabel)}{\hypothesis}$. The 
		loss value $\lossfunc{(\featurevec,\truelabel)}{\hypothesis}$ quantifies the discrepancy 
		between the true \gls{label} $\truelabel$ and the \gls{prediction} $\hypothesis(\featurevec)$. 
		Lower (closer to zero) values $\lossfunc{(\featurevec,\truelabel)}{\hypothesis}$ indicate a smaller 
		discrepancy between \gls{prediction} $\hypothesis(\featurevec)$ and label $\truelabel$. 
		Figure \ref{fig_loss_function_gls} depicts a \gls{lossfunc} for a given \gls{datapoint}, 
		with \gls{feature}s $\featurevec$ and label $\truelabel$, as a function of the \gls{hypothesis} $\hypothesis \in \hypospace$. 
		\begin{figure}[htbp]
			\begin{center}
				\begin{tikzpicture}[scale = 0.7]
					\begin{axis}
						[
						axis x line=center,
						axis y line=center,
						xlabel={},
						xlabel style={below right},
						ylabel style={above right},
						xtick=\empty,
						ytick=\empty,
						xmin=-4,
						xscale = 1.4, 
						xmax=4,
						ymin=-0.5,
						ymax=2.5
						]
						\addplot [smooth, ultra thick] table [x=a, y=b, col sep=comma] {logloss.csv};    
					\end{axis}
					\node [above] at (1,5) {$\lossfunc{(\featurevec,\truelabel)}{\hypothesis}$};
					\node [above] at (10,1) {\gls{hypothesis} $\hypothesis$};
						\node [right] at (4,6) {\gls{loss}};
				\end{tikzpicture}
			\end{center}
			\vspace*{-7mm}
			\caption{Some \gls{lossfunc} $\lossfunc{(\featurevec,\truelabel)}{\hypothesis}$ for a fixed \gls{datapoint}, with 
				\gls{feature} vector $\featurevec$ and \gls{label} $\truelabel$, and varying \gls{hypothesis} $\hypothesis$. 
				ML methods try to find (learn) a \gls{hypothesis} that incurs minimum \gls{loss}.}
			\label{fig_loss_function_gls}
	\end{figure}
 },first={loss function},text={loss function} }

\newglossaryentry{decisiontree}{name={decision tree}, description={A\index{decision tree} 
		decision tree is a flow-chart-like representation of a \gls{hypothesis} map $\hypothesis$. 
		More formally, a decision tree is a directed graph which reads in the feature vector $\featurevec$ 
		of a \gls{datapoint} at its root node. The root node then forwards the \gls{datapoint} to one 
		of its children nodes based on some elementary test on the \gls{feature}s $\featurevec$. 
		If the receiving children node is not a leaf node, i.e., it has itself children nodes, 
	  it represents another test. Based on the test result, the \gls{datapoint} is further 
	  pushed to one of its descendants. This testing and forwarding of the \gls{datapoint} is continued 
	  until the \gls{datapoint} ends up in a leaf node (having no children nodes). 
	  Each leaf node corresponds to a \gls{decisionregion}, a subset of the \gls{featurespace} 
	  mapped to the same output $\hypothesis(\featurevec)$.},first={decision tree},text={decision tree} }

\newglossaryentry{API} 
{
	name={Application Programming Interface (API)},
	description={An\index{application programming interface} application programming 
		interface (API) is a precise specification of the services and resources 
		offered by software or hardware implementing that API.},
	first={application programming interface (API)},
	text={API}
}

\newglossaryentry{hilbertspace}{name={Hilbert space},description={A\index{Hilber space} 
		Hilbert space is a linear vector space equipped with an inner product between 
		pairs of vectors. One important example of a Hilbert space is the Euclidean space 
		$\mathbb{R}^{\featuredim}$, for some dimension $\featuredim$, which consists of 
		Euclidean vectors $\vu = \big(u_{1},\ldots,u_{\featurelen}\big)^{T}$ along with the inner 
		product $\vu^{T} \vv$.},first={Hilbert space},text={Hilbert space}}

\newglossaryentry{sample}{name={sample},description={A\index{sample} 
		finite sequence (list) of \gls{datapoint}s $\datapoint^{(1)},\ldots,\datapoint^{(\sampleidx)}$ that 
		is obtained or interpreted as the realizations of $\samplesize$ \gls{iid} \gls{rv}s 
		with the common \gls{probdist} $p(\datapoint)$. The length $\samplesize$ of 
		the sequence is referred to as the \gls{samplesize}.},first={sample},text={sample}}
	
\newglossaryentry{samplesize}
{name=sample size,
	description={The\index{sample size} number of individual \gls{datapoint}s 
		contained in a \gls{dataset} obtained as the \gls{realization}s of \gls{iid} \gls{rv}s with 
		common \gls{probdist}.},first={sample size},text={sample size}
}

\newglossaryentry{ann}
{name=artificial neural network,
	description={An\index{artificial neural network} artificial neural network is a 
		graphical (signal-flow) representation of a map from \gls{feature}s of 
		a \gls{datapoint} at its input to a \gls{prediction} for the \gls{label} 
		as its output.},first={artificial neural network (ANN)},text={ANN}
}

\newglossaryentry{randomforest}
{name=random forest,
	description={A\index{random forest} random forest is a set (ensemble) of different \gls{decisiontree}s. 
		Each of these \gls{decisiontree}s is obtained by fitting a perturbed copy of 
		the original \gls{dataset}.},first = {random forest}, text={random forest}
}

\newglossaryentry{bagging}{name={bagging},description={Bagging\index{bagging} (or \emph{bootstrap aggregation}) is a generic 
		technique to improve (the robustness of) a given ML method. The idea is to use the \gls{bootstrap} 
		to generate perturbed copies of a given \gls{dataset} and then to learn a separate \gls{hypothesis} for 
		each copy. The resulting set of hypotheses is then used to predict the \gls{label} of a \gls{datapoint} 
		by combining or aggregating the individual predictions of each \gls{hypothesis}. 
		For \gls{hypothesis} maps delivering numeric label values, this aggregation 
		could be implemented by computing the average of individual \gls{prediction}s.},first={bootstrap aggregation (bagging)},text={bagging}}

\newglossaryentry{gd}{name={gradient descent (GD)},description={Gradient\index{gradient descent} 
		descent is an iterative method for finding the minimum of a \gls{differentiable} 
		function $f(\weights)$. },first={gradient descent (GD)},text={GD}}

\newglossaryentry{ladregression}{name={least absolute deviation regression},description={
		Least\index{least absolute deviation regression} absolute deviation regression is an instance of \gls{erm} 
		using the absolute error loss.
		},
		first={least absolute deviation regression},text={least absolute deviation regression}}

\newglossaryentry{bayesrisk}{name={Bayes risk},description={Consider \gls{datapoint}s being \gls{realization}s of 
		\gls{iid} \gls{rv}s with a common \gls{probdist}. The\index{Bayes risk} 
		Bayes risk is the minimum possible \gls{risk} that can be achieved by 
		any \gls{hypothesis} $\hypothesis$ out of a model $\hypospace$. 
		Any \gls{hypothesis} map achieving the minimum \gls{risk} is referred 
		to as a \gls{bayesestimator} \cite{LC}.},first={Bayes risk},text={Bayes risk}}
	
\newglossaryentry{bayesestimator}{name={Bayes estimator},description={A\index{Bayes estimator} 
		\gls{hypothesis} $\hypothesis$ whose \gls{bayesrisk} coincides with the \gls{bayesrisk} \cite{LC}.},
		first={Bayes estimator},text={Bayes estimator}}

\newglossaryentry{weights}{name={weights},
	description={Consider\index{weights} a parametrized \gls{hypospace} $\hypospace$. 
		We\index{weights} use the term weights for numeric \gls{modelparams} that are 
		used to scale \gls{feature}s or their transformations in order to compute $\hypothesis^{(\weights)} \in \hypospace$. A \gls{linmodel} uses weights $\weights=\big(\weight_{1},\ldots,\weight_{\nrfeatures}\big)^{T}$ to compute 
		the linear combination $\hypothesis^{(\weights)}(\featurevec)= \weights^{T} \featurevec$. 
		Weights are also used in \gls{ann}s to form linear combinations of \gls{feature}s or the 
		outputs of neurons in hidden layers.},first={weights},text={weights}}
	
\newglossaryentry{probdist}{name={probability distribution},
	description={To\index{probability distribution} analyze ML methods it can be useful 
		to interpret \gls{datapoint}s as \gls{iid} \gls{realization}s of a \gls{rv}. The typical 
		properties of such \gls{datapoint}s are then governed by the probability distribution 
		of this \gls{rv}. The probability distribution of a binary \gls{rv} $\truelabel \in \{0,1\}$ 
		is fully specified by the probabilities $\prob{\truelabel = 0}$ and 
		$\prob{\truelabel=1} \big(\!=\!1\!-\!\prob{\truelabel=0} \big)$. The probability 
		distribution of a real-valued \gls{rv} $\feature \in \mathbb{R}$ might be specified 
		by a probability density function $p(\feature)$ such that $\prob{ \feature \in [a,b] } \approx  p(a) |b-a|$. 
	    In the most general case, a probability distribution is defined by a probability measure \cite{GrayProbBook,BillingsleyProbMeasure}.},first={probability distribution},text={probability distribution}}

\newglossaryentry{pdf}{name={probability density function (pdf)},
	description={The\index{probability density function} probability density function (pdf) $p(\feature)$ 
		of a real-valued \gls{rv} $\feature \in \mathbb{R}$ is a particular representation of its \gls{probdist}. 
		If the pdf exists, it can be used to compute the probability that $\feature$ takes on a value 
		from a (measurable) set $\mathcal{B} \subseteq \mathbb{R}$ via $\prob{\feature \in \mathcal{B}} = \int_{\mathcal{B}} p(\feature') d \feature'$ \cite[Ch. 3]{BertsekasProb}. The pdf of a vector-valued \gls{rv} $\featurevec \in \mathbb{R}^{\featuredim}$ (if it exists) 
        allows to compute the probability that $\featurevec$ falls into a (measurable) region $\mathcal{R}$ via 
        $\prob{\featurevec \in \mathcal{R}} = \int_{\mathcal{R}} p(\featurevec') d \feature_{1}' \ldots d \feature_{\featuredim}' $ \cite[Ch. 3]{BertsekasProb}.},
first={probability density function (pdf)},text={pdf}}

\newglossaryentry{parameters}{name={parameters},
	description={The\index{parameters} parameters of a ML \gls{model} are tunable 
		(learnable or adjustable) quantities that allow to choose between different \gls{hypothesis} maps. 
		For example, the linear model $\hypospace \defeq \{\hypothesis^{(\weights)}: \hypothesis^{(\weights)}(\feature)= \weight_{1} \feature + \weight_{2}\}$ 
		consists of all \gls{hypothesis} maps $\hypothesis^{(\weights)}(\feature)= \weight_{1} \feature + \weight_{2}$ 
		with a particular choice for the parameters $\weights = \big(\weight_{1},\weight_{2}\big)^{T} \in \mathbb{R}^{2}$. 
		Another example of parameters is the weights assigned to the connections 
		between neurons of an \gls{ann}.},first={parameters},text={parameters}}

\newglossaryentry{lln}{name={law of large numbers},
	description={The\index{law of large numbers} law of large numbers refers to the 
		convergence of the average of an increasing (large) number of \gls{iid} \gls{rv}s 
		to the \gls{mean} of their common \gls{probdist}. Different instances of the 
		law of large numbers are obtained using different notions of convergence \cite{papoulis}.},first={law of large numbers},text={law of large numbers}}
    
\newglossaryentry{stopcrit}{name={stopping criterion},
	description={Many\index{stopping criterion} ML methods use iterative algorithms that construct a 
		sequence of model parameters (such as the weights of a linear map or 
		the weights of an \gls{ann}) that (hopefully) converge to an optimal choice 
		for the model parameters. In practice, given finite computational 
		resources, we need to stop iterating after a finite number of times. 
		A stopping criterion is any well-defined condition required for stopping 
		iterating.},first={stopping criterion},text={stopping criterion}}

\newglossaryentry{kCV}{name={$k$-fold cross-validation ($\nrfolds$-fold CV)},
	description={$k$-fold cross-validation\index{k-fold cross-validation} is a 
		method for learning and validating a \gls{hypothesis} using a given \gls{dataset}. 
		This method divides the \gls{dataset} evenly into $k$ subsets or \emph{folds} 
		and then executes $k$ repetitions of \gls{model} training (e.g., via \gls{erm}) and \gls{validation}. 
		Each repetition uses a different fold as the \gls{valset} and the remaining $k-1$ folds 
		as a \gls{trainset}. The final output is the average of the \gls{valerr}s obtained 
		from the $k$ repetitions.},first={$k$-fold cross-validation ($k$-fold CV)},text={$k$-fold CV}}
	
\newglossaryentry{renyidiv}{name={R\'enyi divergence}, 
	description={The R\'enyi divergence\index{R\'enyi divergence} measures the (dis-)similarity 
		between two \gls{probdist}s \cite{RenyiInfo95}.}, 
	first = {R\'enyi divergence}, text = {R\'enyi divergence}} 

\newglossaryentry{nonsmooth}{name={non-smooth},
	description={We\index{non-smooth} refer to a function as non-smooth if it is not 
		\gls{smooth} \cite{nesterov04}.},first={non-smooth},text={non-smooth}}

\newglossaryentry{convex}{name={convex},
	description={A\index{convex} subset $\mathcal{C} \subseteq \mathbb{R}^{\featuredim}$ of the 
		\gls{euclidspace} $\mathbb{R}^{\featuredim}$ is referred to as 
		convex if it contains the line segment between any two points 
		of that set. We define a function as convex if its epigraph is a 
		convex set \cite{BoydConvexBook}.},first={convex},text={convex}}

\newglossaryentry{smooth}{name={smooth},
	description={We\index{smooth} refer to a real-valued function as smooth if it is \gls{differentiable} 
		and its \gls{gradient} is continuous \cite{nesterov04,CvxBubeck2015}. A 
		differentiable function $f(\weights)$ is referred to as $\beta$-smooth if the \gls{gradient} 
		$\nabla f(\weights)$ is Lipschitz continuous with Lipschitz constant $\beta$, i.e., 
		$$\| \nabla f(\weights) - \nabla f(\weights') \| \leq \beta \| \weights - \weights' \|.$$ },first={smooth},text={smooth}}

\newglossaryentry{dataug}{name={data augmentation},
	description={Data augmentation\index{data augmentation} methods add synthetic \gls{datapoint}s 
		to an existing set of \gls{datapoint}s. These synthetic \gls{datapoint}s might be obtained by 
		perturbations (adding noise) or transformations (rotations of images) of the original \gls{datapoint}s. 
		These perturbations and transformations are such that the resulting synthetic \gls{datapoint}s should 
		still have the same \gls{label}. As a case in point, a rotated cat image is still a cat image even if 
		their \gls{feature} vectors (obtained by stacking pixel color intensities) are very different.},first={data augmentation},text={data augmentation}}

\newglossaryentry{localdataset}{name={local dataset},description={The\index{local dataset} concept of a local dataset is 
		in-between the concept of a \gls{datapoint} and a \gls{dataset}. A local dataset consists of several 
		individual \gls{datapoint}s which are characterized by \gls{feature}s and \gls{label}s. In contrast to a 
		single \gls{dataset} used in basic ML methods, a local dataset is also related to other local datasets via different notions 
		of similarities. These similarities might arise from \gls{probmodel}s or communication infrastructure and 
		are encoded in the edges of a \gls{empgraph}.},first={local dataset},text={local dataset}}
	
\newglossaryentry{localmodel}{name={local model},description={Consider\index{local model} a collections 
		of \gls{localdataset}s that are assigned to the nodes of a \gls{empgraph}. A local model $\localmodel{\nodeidx}$ 
		is a \gls{hypospace} assigned to a node $\nodeidx \in \nodes$. Different nodes might be assigned 
		different \gls{hypospace}s, i.e., in general $\localmodel{\nodeidx} \neq \localmodel{\nodeidx'}$ for different 
		nodes $\nodeidx, \nodeidx' \in \nodes$.  },first={local model},text={local model}}
	
\newglossaryentry{mutualinformation}
{name={mutual information},
 description={The\index{mutual information} mutual information $\mutualinformation{\featurevec}{\truelabel}$ 
 	between two \gls{rv}s $\featurevec$, $\truelabel$ defined on the same probability 
 	space is given by \cite{coverthomas} $$\mutualinformation{\featurevec}{\truelabel} \defeq \expect \left\{ \log \frac{p (\featurevec,\truelabel)}{p(\featurevec)p(\truelabel)} \right\}.$$ It is a measure for how well we can estimate $\truelabel$ based 
	solely from $\featurevec$. A large value of $\mutualinformation{\featurevec}{\truelabel}$ indicates that 
	$\truelabel$ can be well predicted solely from $\featurevec$. This \gls{prediction} could be obtained by a 
		\gls{hypothesis} learnt by a \gls{erm}-based ML method. 
	 }, first = {mutual information (MI)}, text={MI} 
}

\newglossaryentry{zerogradientcondition}{name={zero-gradient condition},
	description={Consider\index{zero-gradient condition} the unconstrained 
		optimization problem $\min_{\weights \in \mathbb{R}^{\dimlocalmodel}} f(\weights)$  with 
			a \gls{smooth} and \gls{convex} \gls{objfunc} $f(\weights)$. A necessary and 
			sufficient condition for a vector $\widehat{\weights} \in \mathbb{R}^{\dimlocalmodel}$ 
			to solve this problem is that the \gls{gradient} $\nabla f \big( \widehat{\weights} \big)$ is the zero-vector, 
			$$ \nabla f \big( \widehat{\weights} \big) = \mathbf{0} \Leftrightarrow  f \big( \widehat{\weights} \big) = \min_{\weights \in \mathbb{R}^{\dimlocalmodel}} f(\weights) .$$ }, 
			first={zero-gradient condition},text={zero-gradient condition}}

\newglossaryentry{edgeweight}{name={edge weight},
	description={Each\index{edge weight} edge $\edge{\nodeidx}{\nodeidx'}$ of a \gls{empgraph} is 
		assigned a non-negative \gls{edgeweight}  $\edgeweight_{\nodeidx,\nodeidx'}\geq0$. 
		A zero \gls{edgeweight} $\edgeweight_{\nodeidx,\nodeidx'}=0$ indicates the absence 
		of an edge between nodes $\nodeidx, \nodeidx' \in \nodes$.}, 
	first={edge weight},text={edge weight}}

\newglossaryentry{dataminprinc}{name={data minimization principle},
	description={European\index{data minimization principle} data protection regulation 
		includes a data minimization principle. This principle requires a data controller to 
		limit the collection of personal information to what is directly relevant and necessary 
		to accomplish a specified purpose. The data should be retained only for as long as 
		necessary to fulfill that purpose \cite[Article 5(1)(c)]{GDPR2016} \cite{EURegulation2018}.}, 
	first={data minimization principle},text={data minimization principle}}

\newmdenv[
backgroundcolor=yellow!20,
linecolor=orange,
linewidth=1pt, 
innerleftmargin=5pt,
innerrightmargin=5pt,
innertopmargin=5pt,
innerbottommargin=5pt
]{highlightbox}

\makeglossaries

\begin{document}
\pagenumbering{arabic}

\title{Engineering Trustworthy AI: A Developer Guide for Empirical Risk Minimization}

\author{
	\IEEEauthorblockN{Diana Pfau and Alexander Jung}
	\IEEEauthorblockA{Department of Computer Science, Aalto University, Espoo, Finland}
}



\IEEEpubid{0000--0000/00\$00.00~\copyright~2021 IEEE}

\maketitle
\renewcommand\thefootnote{}
\footnotetext{This work was supported by Research Council of Finland grant nr.\ $363624$, $349965$ and $331197$.}
\renewcommand\thefootnote{\arabic{footnote}} 

\begin{abstract}
AI systems increasingly shape critical decisions across personal and societal domains. 
Indeed, we routinely use AI systems to search for jobs, housing and romantic relationships. 
These systems often use empirical risk minimization (ERM) in order to train a powerful 
predictive model such as a deep neural network. So far, the design of ERM-based method 
prioritizes accuracy over trustworthiness, resulting in biases, opacity, and other adverse effects. 
This paper discusses how key requirements for trustworthy AI can be translated into 
design choices for the components of ERM. We hope to provide actionable guidance 
for building AI systems that meet emerging standards and regulations for trustworthy AI.


\end{abstract}


\section{Introduction}

Artificial intelligence (AI) has become integral to our daily lives, influencing aspects such 
as job searches, housing, and finding new relationships \cite{WangFairRecomm2023,SHARABI2024107973}. 
Most current AI systems employ machine learning (ML) to train personalized models for 
users. These trained models provide tailored \gls{prediction}s on interests like job 
offers, dating, and music videos \cite{Quian2014}. The availability of tailored (personalized) 
\gls{prediction}s is instrumental for many applications. As a point in case, the use of 
personalized diagnosis and treatment can significantly improve healthcare \cite{Sankar:2017aa}.

\subsection{Anecdotal AI Trust Concerns} 
Despite the usefulness of ML applications, there is increasing evidence for 
their potentially harmful effects: 
\begin{itemize} 
	
\item {\bf Impact on Democratic Processes.} 
Social media platforms use ML in the form of recommender systems to select (or suppress) 
information presented to a user \cite{WSJTikTok}. These recommendation systems can (be 
exploited to) amplify sensationalist and divisive content which, in turn, can deepen polarization 
and the fragmentation of public sphere into filter bubbles \cite{Benkler2018,SUNSTEIN:2018aa}. 
There is also evidence for the exploitation of these effects in order to influence core democratic 
processes such as elections \cite{Almond:2022aa,HaoFacbook2021}. 

\item {\bf Autopilot Crashes.} AI-based control of vehicles has been associated with several 
notable accidents. In some instances, the system failed to detect a specific type of obstacle 
(such as emergency vehicles) or misinterpreted road conditions and traffic signs \cite{Moreschi2023,Isidore2021}. 
AI-based autopilots might also reduce driver engagement and, in turn, awareness for 
dangerous situations that require human intervention \cite{Spector}. This case illustrates 
the importance of requiring AI systems to be transparent about their operation and limitations \cite{stempel2024tesla}. 

\item 
	{\bf The Cambridge Analytica Scandal.} The British firm Cambridge Analytica accessed vast 
	amounts of personal data from Facebook without explicit permission, thereby violating privacy 
	rights and data protection regulations \cite{rosenberg2018cambridge,cadwalladr2018cambridge,ico2018investigation}.
	Cambridge Analytica used the data to create detailed psychological profiles, 
	which were then used to micro-target individuals with tailored political ads \cite{grassegger2017data}. 
	The firm was involved in several high-profile political campaigns, including Donald Trump’s 2016 
	presidential campaign and the Leave.EU campaign for Brexit, using data-driven strategies to sway public opinion \cite{howard2016botsstrongerinbrexitcomputational}. The Cambridge Analytica scandal 
	highlights the requirements for trustworthy AI regarding privacy protection and societal wellbeing of \cite{Zuboff2018}. 
	
\item {\bf COMPAS Recidivism Prediction Algorithm.} A study found that the COMPAS 
algorithm, used in the U.S. justice system to predict recidivism, disproportionately 
predicted  African-American and female defendants at higher risk compared to white 
male defendants \cite{Angwin2016,Melissa2019}. This finding raised concerns about 
a potential discriminatory behaviour of the COMPAS algorithm \cite{Dressel:2018aa}. 
	
\item {\bf Uighur minority in China.}
Facial recognition has been used to identify members of the Muslim minority group 
of Uighurs  \cite{mozur2019china,greitens2022china,hrw2018china,amnesty2020china}.
The use of facial recognition technology to target a specific ethnic group 
highlights fundamental concerns about harmful effects or misuse of AI systems. 
Trustworthy AI must adhere to ethical principles and respect human rights including 
privacy and wellbeing on individual as well as on societal level). 
\end{itemize} 
	
\subsection{The Need to Regulate AI}

The use of AI is already regulated by existing legal frameworks. Indeed, any smartphone app 
that uses AI must conform to existing consumer protection law \cite{coppa1998,ftc1914}. 
However, these existing legal frameworks are inadequate for the regulation of 
internet-scale AI systems \cite{HLEGTrustworhtyAI,oecd2019ai,floridi2019five}. 

Existing legal frameworks traditionally emphasize individual harms such as the 
mental well-being of a specific child that uses a AI-powered smartphone app. 
However, the AI systems might be harmful on larger scales such as entire democracies. 
Policy-makers have recognized the need for new legal frameworks to regulate 
AI technology in order to address a significantly larger scale of harmful effects \cite{GDPR2016,Wachter2020,AIAct,XMLISO,HLEGTrustworhtyAI}. 
The regulation of AI systems is particularly important for critical application 
domains such as education \cite{Stoilova2019}, financial services \cite{Fuster2021ssrn} or 
border control \cite{AIAct,Molnar2020}. 


The European Union has formulated {\bf key requirements for trustworthy AI} \cite{HLEGTrustworhtyAI}. 
Among those key requirements are the {\bf robustness},  {\bf \gls{privprot}}, 
{\bf fairness}, and {\bf \gls{explainability}} of AI systems. These requirements closely resemble Australia's 
AI Ethics principles \cite{australia_ai_ethics_2024} as well as the AI principles 
laid out by the Organisation for Economic Co-operation and Development (OECD) \cite{oecd_ai_principles}. 

\begin{figure}[htbp]
	\vspace*{-2mm}
	\begin{center}
		\newcommand{\myarray}{{"compute","data","privacy","robustness","fairness","explainability"}}
		\begin{tikzpicture}[scale=0.90]

			\def\numaxes{6}
			\def\angle{360/\numaxes}
			
			\foreach \i in {1,...,5} {
				\node[] at (\i*\angle:3.4)() {\pgfmathparse{\myarray[\i-1]}\pgfmathresult};
			}
			
			\node[] at (6*\angle:3.7)() {\pgfmathparse{\myarray[5]}\pgfmathresult};
			
			\def\data{{4, 4, 2, 2, 2,1}}
			\def\datanew{{1, 1, 4, 4, 4,3}}
			
			\foreach \i in {1,...,5} {
				\draw (0,0) -- (\i*\angle:3);
			}
			\draw (0,0) -- (6*\angle:2) ; 
			
			\foreach \i in {1,...,\numaxes} {
				\draw (\i*\angle:3*\datanew[\i-1]/5) node[circle,fill,inner sep=1.5pt] (pointnew\i) {};
			}
			\node[blue,left] at (-2.0,-1){some method}; 
			\draw [line width=0.5mm, blue] (pointnew1) -- (pointnew2) -- (pointnew3) -- (pointnew4) -- (pointnew5) -- (pointnew6) --  (pointnew1);
		\end{tikzpicture}
		\vspace*{-4mm}
	\end{center} 
	\caption{Trustworthy AI adds new design criteria for \gls{erm} based methods. Besides small  
		computational complexity and data requirements, these methods 
		must be sufficiently explainable, privacy-friendly, fair and robust. }
	\label{fig_measures_trustworthy_FL} 
	\vspace*{-2mm}
\end{figure}



\section{Empirical Risk Minimization as AI Engine} 
\label{sec_erm_AI_engine} 

Many of the current AI systems are based on machine learning (ML) techniques. 
The goal of ML is to predict some quantity of interest (its \gls{label}) $\truelabel$ from 
low-level measurements (its \gls{feature}s) $\featurevec=\big(\feature_{1},\ldots,\feature_{\nrfeatures}\big)^{T}$. 
The \gls{prediction}s are computed via some \gls{hypothesis} map $\hypothesis$ that reads in the 
\gls{feature}s of a \gls{datapoint} and delivers a \gls{prediction} $\widehat{\truelabel} = \hypothesis(\featurevec)$ 
for its \gls{label}. 

{\bf Model.} The \gls{hypothesis} $\hypothesis$ is learnt or optimized based on the discrepancy 
between previous predictions and observed labels. The space of possible hypothesis
maps, from which a ML method can choose from, is referred to as \gls{hypospace} or model. 

{\bf Loss.} To choose or learn a useful \gls{hypothesis} from a model we need a 
measure for the quality of the \gls{prediction}s obtained from a \gls{hypothesis}. 
To this end, ML methods use \gls{lossfunc}s $\lossfunc{\pair{\featurevec}{\truelabel}}{\hypothesis}$ 
to obtain a quantitative measure for the \gls{prediction} errors. 

{\bf Risk.} The ultimate goal of ML is to learn a \gls{hypothesis} $\learnthypothesis \in \hypospace$ that 
incurs a small \gls{loss} when predicting the \gls{label} of any \gls{datapoint}. We can 
make this informal requirement precise by interpreting \gls{datapoint}s as \gls{realization}s
of \gls{iid} \gls{rv}s with a common \gls{probdist} $p(\featurevec,\truelabel)$. 
This allows to define the expected \gls{loss} or \gls{risk} of a \gls{hypothesis}, 
\begin{equation}
\label{equ_def_risk} 
	\risk{\hypothesis} \defeq \expect\big\{ \lossfunc{\pair{\featurevec}{\truelabel}}{\hypothesis} \big\}. 
\end{equation}

{\bf Data.} Since the underlying \gls{probdist} $p(\featurevec,\truelabel)$ is 
typically unknown, we cannot directly optimize the \gls{risk} \eqref{equ_def_risk}. 
Instead, practical ML methods need to approximate the \gls{risk} from a \gls{dataset}  
\begin{equation}
\label{equ_def_trainset}
	\dataset = \left\{ \pair{\featurevec^{(1)}}{\truelabel^{(1)}},\ldots,\pair{\featurevec^{(\samplesize)}}{\truelabel^{(\samplesize)}} \right\}.
\end{equation} 
The \gls{dataset} is constituted by \gls{datapoint}s, each characterized 
by \gls{feature}s $\featurevec$ and some \gls{label} $\truelabel$. \gls{erm}-based 
methods require a \gls{dataset} to measure the usefulness of a \gls{hypothesis} 
$\hypothesis \in \hypospace$ and, in turn, to train a \gls{model} (learn a 
useful choice for the \gls{modelparams}). 

{\bf Empirical Risk.} Arguably, the most widely used approximation to the risk \eqref{equ_def_risk} 
is the average \gls{loss} or \gls{emprisk}, 
\begin{equation} 
	\label{equ_def_emp_risk} 
	\emprisk{\hypothesis}{\dataset} \defeq (1/\samplesize) \sum_{\sampleidx=1}^{\samplesize} \lossfunc{\pair{\featurevec^{(\sampleidx)}}{\truelabel^{(\sampleidx)}}}{\hypothesis}.
\end{equation}
Much of statistical learning theory revolves around the study of the approximation 
$\emprisk{\hypothesis}{\dataset}  \approx \risk{\hypothesis}$. The approximation 
quality can be studied via different forms of the law of large numbers or concentration 
inequalities \cite{BertsekasProb,BillingsleyProbMeasure,Wain2019,ShalevMLBook}. 

\begin{figure}
\begin{center}
\begin{tikzpicture}
	\begin{axis}[
		axis lines = left,
		xlabel = {\gls{hypothesis} $\hypothesis \in \hypospace$},
		xlabel style={at={(axis description cs:0.5,0.0)}, anchor=north},
		ylabel = {},
		legend pos = north east,
		domain=0:5, 
		samples=100,
		ymin=0, ymax=2,
		xmin=0, xmax=5,
		xtick=\empty,
		ytick=\empty,
		enlargelimits=false,
		clip=false,
		axis on top,
		smooth,
		yshift=0cm
		]
		
		\addplot[blue, thick,,domain=0.5:3] {0.5*(x-2)^2+0.3 }
		node[pos=0, anchor=south] {\gls{risk}}; 

		\addplot[red, thick, dashed, samples=100,domain=2:4] {0.5*(x-3)^2+0.1}
		node[pos=1, anchor=south] {\gls{emprisk}}; 

		
	\end{axis}
\end{tikzpicture}
\end{center}
\caption{\gls{erm} uses the average \gls{loss} incurred on a \gls{trainset} to 
	approximate the \gls{risk} (or expected \gls{loss}). }
\end{figure} 

{\bf Empirical Risk Minimization.} \Gls{erm}-based methods learn a \gls{hypothesis} 
$\learnthypothesis \in \hypospace$ from a \gls{hypospace} (or \gls{model}) 
$\hypospace$ by minimizing the \gls{emprisk} $\emprisk{\hypothesis}{\dataset}$ 
as a proxy for the \gls{risk}, 
\begin{align} 
	\label{equ_def_emp_risk_min} 
	\learnthypothesis & \defeq \argmin_{\hypothesis \in \hypospace} \emprisk{\hypothesis}{\dataset} \nonumber \\ 
	& = \argmin_{\hypothesis \in \hypospace} \sum_{\pair{\featurevec}{\truelabel} \in \dataset} \lossfunc{\pair{\featurevec}{\truelabel}}{\hypothesis}.
\end{align}
We obtain practical ML systems by applying optimization methods to 
solve \eqref{equ_def_emp_risk_min}. Different ML methods are obtained from 
different design choices for \gls{datapoint}s (their \gls{feature}s and \gls{label}), 
the \gls{hypospace} (or model) and \gls{lossfunc} \cite[Ch. 3]{MLBasics}. 

{\bf Design Choices.} From a ML engineering perspective, the design choices in \gls{erm} are mainly guided by \gls{compasp} 
and \gls{statasp} of the resulting optimization problem \eqref{equ_def_emp_risk_min}. 
The \gls{compasp} include the number of arithmetic operations required by a ML method. 
The \gls{statasp} include the generalization error $\risk{\learnthypothesis}- \emprisk{\learnthypothesis}{\dataset}$ 
of the learnt \gls{hypothesis} and its robustness against the presence of \gls{outlier}s in 
the \gls{trainset} \eqref{equ_def_trainset}. 

{\bf Generalization.} While measuring the computational complexity via counting arithmetic operations is 
quite straightforward \cite{calflops}, measuring the generalization error is more challenging. 
Indeed, since we typically do not know the underlying probability distribution of \gls{datapoint}s, we 
can only estimate the generalization performance via a \gls{valset}. The \gls{valset} 
consists of \gls{datapoint}s that have not been used for the \gls{trainset} $\dataset$ 
in \gls{erm} \eqref{equ_def_emp_risk_min}. 
\begin{figure}[htbp]
	\begin{center}
		\begin{tikzpicture}[ycomb,yscale=.5]
			\draw[color=blue,line width=26pt]
			plot coordinates{(0,3)};
			\node [below] at (0,0) {\gls{trainerr}} ; 
			\draw[color=red,line width=26pt]
			plot coordinates{(2.5,5)};
			\node [below] at (2.5,0) {validation error} ; 
			\draw[dashed,line width=2] (-1,2) -- (3.5,2) node[right,text width=5cm]{baseline or \index{benchmark}benchmark \\ (e.g., \gls{bayesrisk},  \\existing ML methods or \\ human performance)};
		\end{tikzpicture}
	\end{center}
	\caption{We can diagnose a ML method by comparing its \gls{trainerr} 
		with its \gls{valerr}. Ideally both are on the same level as a \index{baseline}\gls{baseline} (or 
		benchmark error level).}
	\label{fig_bars_val_sel}
\end{figure}

\begin{highlightbox}
Ensuring trustworthy AI with \gls{erm} requires not only statistical and computational 
optimization but also careful design choices for training data, ML \gls{model}, and \gls{lossfunc}. 
This paper explores how targeted design choices in these three components can meet 
key requirements for trustworthy AI.
\end{highlightbox}

{\bf Regularization.} Consider a \gls{erm}-based ML method using a \gls{hypospace} $\hypospace$ and 
\gls{dataset} $\dataset$ (we assume all \gls{datapoint}s are used for training). 
A key parameter for such a ML method is the ratio $\effdim{\hypospace}/|\dataset|$ 
between the (effective) \gls{model} size $\effdim{\hypospace}$ and the number $|\dataset|$ 
of \gls{datapoint}s.\footnote{Arguably, the most widely used measure for the effective 
	size of a ML model is the \gls{vcdim} \cite{ShalevMLBook}. However, the precise definition of the model size is not 
	relevant for our discussion.} The tendency of the ML method to overfit increases with the 
ratio $\effdim{\hypospace}/|\dataset|$. 

\Gls{regularization} techniques reduce the ratio $\effdim{\hypospace}/|\dataset|$ via three (essentially equivalent) 
approaches: 
\begin{itemize} 
	\item get more \gls{datapoint}s, possibly via \gls{dataug} , 
	\item add penalty term $\regparam \regularizer{\hypothesis}$ to the 
	average \gls{loss} in \gls{erm} \eqref{equ_def_emp_risk}, 
	\item shrink the \gls{hypospace}, e.g., by adding constraints on the \gls{modelparams} 
	such as $\normgeneric{\weights}{2} \leq 10$.
\end{itemize} 
It can be shown that these three perspectives (corresponding to the three 
components \gls{data}, \gls{model} and \gls{loss}) on \gls{regularization} are 
closely related \cite[Ch. 7]{MLBasics}. For example, adding a penalty term $\regparam \regularizer{\hypothesis}$ 
in \gls{erm} \eqref{equ_def_emp_risk} is equivalent to \gls{erm} \eqref{equ_def_emp_risk} 
with a pruned \gls{hypospace} $\hypospace^{(\regparam)} \subseteq \hypospace$. 
Using a larger $\regparam$ typically results in a smaller $\hypospace^{(\regparam)}$ \cite[Ch. 7]{MLBasics}. 
Moreover, adding the penalty term $\regparam \regularizer{\hypothesis}$ is equivalent to 
augmenting the original \gls{trainset} with perturbations of its \gls{datapoint}s (see Fig.\ \ref{fig_equiv_dataaug_penal}). 

\begin{figure}[htbp]
	\begin{center} 
		\begin{tikzpicture}[scale = 1]
			\draw[->, very thick] (0,0.5) -- (7.7,0.5) node[below] {feature $\feature$};       
			\draw[->, very thick] (0.5,0) -- (0.5,4.2) node[above] {label $\truelabel$};   
			
			\draw[color=black, thick, dashed, domain = -0.5: 5.2, variable = \x]  plot ({\x},{\x*0.4 + 2.0}) ;     
			\node at (5.7,4.1) {$\hypothesis(\feature)$};    
			
			\coordinate (l1)   at (1.2, 2.48);
			\coordinate (l2) at (1.4, 2.56);
			\coordinate (l3)   at (1.7,  2.68);
			
			\coordinate (l4)   at (2.2, 2.2*0.4+2.0);
			\coordinate (l5) at (2.4, 2.4*0.4+2.0);
			\coordinate (l6)   at (2.7,  2.7*0.4+2.0);
			
			\coordinate (l7)   at (3.9,  3.9*0.4+2.0);
			\coordinate (l8) at (4.2, 4.2*0.4+2.0);
			\coordinate (l9)   at (4.5,  4.5*0.4+2.0);
			
			\coordinate (n1)   at (1.2, 1.8);
			\coordinate (n2) at (1.4, 1.8);
			\coordinate (n3)   at (1.7,  1.8);
			
			\coordinate (n4)   at (2.2, 3.8);
			\coordinate (n5) at (2.4, 3.8);
			\coordinate (n6)   at (2.7,  3.8);

			\coordinate (n7)   at (3.9, 2.6);
			\coordinate (n8) at (4.2, 2.6);
			\coordinate (n9)   at (4.5,  2.6);
			
			\node at (n1)  [circle,draw,fill=red,minimum size=6pt,scale=0.6, name=c1] {};
			\node at (n2)  [circle,draw,fill=blue,minimum size=6pt, scale=0.6, name=c2] {};
			\node at (n3)  [circle,draw,fill=red,minimum size=6pt,scale=0.6,  name=c3] {};
			\node at (n4)  [circle,draw,fill=red,minimum size=12pt, scale=0.6, name=c4] {};  
			\node at (n5)  [circle,draw,fill=blue,minimum size=12pt,scale=0.6,  name=c5] {};
			\node at (n6)  [circle,draw,fill=red,minimum size=12pt, scale=0.6, name=c6] {};  
			\node at (n7)  [circle,draw,fill=red,minimum size=12pt,scale=0.6,  name=c7] {};
			\node at (n8)  [circle,draw,fill=blue,minimum size=12pt, scale=0.6, name=c8] {};
			\node at (n9)  [circle,draw,fill=red,minimum size=12pt, scale=0.6, name=c9] {};
			
			\draw [<->] ($ (n7) + (0,-0.3) $)  --  ($ (n9) + (0,-0.3) $) node [pos=0.4, below] {$\sqrt{\regparam}$}; ;

			\draw[<->, color=red, thick] (l1) -- (c1);  
			\draw[<->, color=blue, thick] (l2) -- (c2);  
			\draw[<->, color=red, thick] (l3) -- (c3);  
			\draw[<->, color=red, thick] (l4) -- (c4);  
			\draw[<->, color=blue, thick] (l5) -- (c5);  
			\draw[<->, color=red, thick] (l6) -- (c6);  
			\draw[<->, color=red, thick] (l7) -- (c7);  
			\draw[<->, color=blue, thick] (l8) -- (c8);  
			\draw[<->, color=red, thick] (l9) -- (c9);  
			
			\draw[fill=blue] (5.2, 2.7)  circle (0.1cm) node [black,xshift=1.2cm] {\gls{trainset} $\dataset$};
			\draw[fill=red] (5.2, 2.2)  circle (0.1cm) node [black,xshift=1.2cm] {augmentation};
			\node at (4.6,1.2)  [minimum size=20pt, font=\fontsize{11}{0}\selectfont, text=blue] {$\frac{1}{\samplesize} \sum_{\sampleidx=1}^\samplesize \lossfunc{\pair{\featurevec^{(\sampleidx)}}{ \truelabel^{(\sampleidx)}}}{\hypothesis}$};
			\node at (7.4,1.2)  [minimum size=20pt, font=\fontsize{11}{0}\selectfont, text=red] {$+\regparam \regularizer{\hypothesis}$};
		\end{tikzpicture}
		\caption{Equivalence between \gls{dataug} and \gls{loss} penalization. \label{fig_equiv_dataaug_penal} }
	\end{center}
\end{figure} 


\section{Key Requirements for Trustworthy AI}
\label{sec_seven_key_requirements} 

The European Union put forward seven key requirements for trustworthy 
\gls{ai} \cite{HLEGTrustworhtyAI}. These requirements are motivated by 
the EU Charter of fundamental rights as the ultimate legal basis for trustworthy AI \cite{eu_charter_2012}. 
We next list these key requirements for trustworthy AI along with their 
motivation from the perspective of fundamental rights. 
\begin{enumerate} 
	\item {\bf KR1- Human Agency and Oversight  \cite[p.15]{HLEGTrustworhtyAI}.} 
	The requirement of Human agency and oversight is based on the idea of human autonomy, 
	which results from the right to dignity \cite[Article 1]{eu_charter_2012}: 
	Every person regardless of any other characteristics has an inherent, equal 
	and inalienable value. KR1 is also aligned wit the right to liberty \cite[Article 6]{eu_charter_2012} 
	which determines that every person has a right to decide over their own life.
		 
	\item {\bf KR2 -Technical Robustness and Safety \cite[p.16]{HLEGTrustworhtyAI}.} 
   \Gls{erm}-based methods must perform reliably under various conditions, minimizing 
   risks of harm. KR2 aligns with several EU fundamental rights, such as the right to 
   life \cite[Article 2]{eu_charter_2012}, the physical and mental integrity of the person \cite[Article 3]{eu_charter_2012}, 
   and the protection of personal data \cite[Article 8]{eu_charter_2012}. 
   Section \ref{sec_techrobustness_erm} discusses the robustness of \gls{erm}-based AI 
   systems against perturbations of data sources and imperfections of 
   computational infrastructure. 
  		
	\item {\bf KR3 - Privacy and Data Governance  \cite[p.17]{HLEGTrustworhtyAI}.}	
	\gls{erm}-based methods must ensure protection against unauthorized access to - 
	and misuse of - personal \gls{data}. Data and \gls{privprot} are typically implemented 
	as part of a \gls{data} governance framework \cite{Khatri2010}. KR3 aligns with 
	individuals’ rights to privacy and the security of their personal data. 
	
	\item {\bf KR4 - Transparency  \cite[p.18]{HLEGTrustworhtyAI}.} 
	\Gls{transparency} is to enable a person to utilise their right to take action where they 
	believe they have been treated wrongly. This is closely related to the right to 
	data protection. To take actions against a potentially unlawful processing or 
	an unjustified outcome of an \gls{erm}-based AI system, a user has to have 
	enough information to understand how processing has taken place or how 
	a decision was reached.
	
	\item {\bf KR5 - Diversity, Non-discrimination and Fairness  \cite[p.18]{HLEGTrustworhtyAI}.} 
	KR5 is aligned with \cite[Article 21]{eu_charter_2012} which prohibits discrimination 
	based on factors such as race, gender, and religion. AI systems must treat all individuals 
	fairly and inclusively, safeguarding their right to equality. Ensuring KR5 includes 
	quality control for the \gls{dataset} $\dataset$ used in \gls{erm} as well as the usability 
	of interfaces for different user groups. KR5 is ultimately rooted in the inalienable value 
	of all persons.
	
	\item {\bf KR6 - Societal and Environmental Well-Being \cite[p.19]{HLEGTrustworhtyAI}.} 
	KR6 covers the impact of AI systems on environmental and social well-being \cite{Chry2020},\cite[Article 35]{eu_charter_2012}. 
	AI systems should minimize harm to the environment and foster a 
	sustainable development. By doing so, this requirement supports 
	both individual rights and the collective welfare of society.
	
	\item {\bf KR7 - Accountability \cite[p.19]{HLEGTrustworhtyAI}.} 
    KR7 supports fundamental rights to justice, remedy, and \gls{transparency} \cite{eu_charter_2012}. 
	Accountability requires mechanisms to identify, explain, and address potential harm 
	of AI systems. Organisations that operate an AI system are responsible for its direct 
	and indirect effects on the user \cite{ALTAIEU,AIAct}. Developers and deployers must 
	implement measures that allow to explain the aims, motivations, and reasons underlying 
	the behaviour of AI systems. Accountability includes the reporting of data 
	breaches and the possibility of redress \cite{fountaine2019building}. 
	
\end{enumerate} 
The following sections discuss in some detail how the above requirements 
guide the design choices for \gls{data}, \gls{model} and \gls{loss} of \gls{erm} 
(see Section \ref{sec_erm_AI_engine}). As illustrated in Figure \ref{fig_trustworthyAI_design_space}, 
our main goal is to identify regions in the design space for \gls{erm} that enable trustworthy 
AI systems. 
\begin{figure}[htbp]
	\begin{center}
\begin{tikzpicture}
	\pgfmathsetmacro{\xoffset}{0.5}
	\pgfmathsetmacro{\yoffset}{0.5}
	
	\draw[->, line width=1pt, >=stealth] (0,0,0) -- (4,0,0) node[anchor=west] {\gls{data}}; 
	\draw[->, line width=1pt, >=stealth] (0,0,0) -- (0,4,0) node[anchor=south east] {\gls{model}}; 
	\draw[->, line width=1pt, >=stealth] (0,0,0) -- (0,0,4) node[anchor=north] {\gls{loss}}; 
	
    \filldraw[blue, opacity=0.3, smooth, tension=0.8] 
plot coordinates {
	(\xoffset+1, \yoffset+1, 1) 
	(\xoffset+2.5, \yoffset+1, 1.5) 
	(\xoffset+3, \yoffset+2, 2) 
	(\xoffset+2, \yoffset+3.5, 2.5) 
	(\xoffset+1.2, \yoffset+2.8, 3.5) 
	(\xoffset+1, \yoffset+1.5, 3.2) 
	(\xoffset+1, \yoffset+1, 1)
};

\node at (\xoffset+2, \yoffset+2, 3) [anchor=south] {trustworthy AI};
\end{tikzpicture}
\end{center}
\caption{\gls{erm}-based methods are defined by design choices for \gls{data}, \gls{model} and \gls{loss}. 
	This paper discusses design choices that facilitate KRs for trustworthy 
	AI. \label{fig_trustworthyAI_design_space}} 
\end{figure}

\section{KR1 - Human Agency and Oversight} 
\label{sec_kr1} 

\Gls{erm}-based methods must be designed to support user agency, ensuring human oversight, 
and safeguarding fundamental rights (see Section \ref{sec_seven_key_requirements}). 
The \gls{prediction}s delivered by a trained \gls{model} must not result in 
any manipulation or undue influence. We must ensure safeguards to maintain 
human control and the prevention of harmful outcomes. KR1 is closely related 
to fundamental rights such as dignity, freedom, and non-discrimination \cite{eu_charter_2012}.

{\bf Human Agency.} Users should be able to understand, interact with, and challenge decisions 
based on the predictions $\learnthypothesis(\featurevec)$ delivered by a trained model $\learnthypothesis\in \hypospace$. 
Human agency is facilitated by using transparent models (see Section \ref{sec_KR1_model} and Section \ref{sec_kr4}) and 
comprehensive documentation of the training process (e.g., optimization method used to solve \eqref{equ_def_emp_risk}). 
The \gls{erm} design choices for \gls{datapoint}s (their \gls{feature}s and \gls{label})
 and \gls{lossfunc} (see Section \ref{sec_KR1_data} and Section \ref{sec_KR1_loss} must 
 ensure that the trained \gls{model} avoids any manipulation or deception or users 
all of which may threaten individual autonomy \cite{Botes:2023aa,doi:10.1080/21670811.2017.1345645,10.1093/ct/qtab019}. 

{\bf Human Oversight.} We must design \gls{erm}-based methods that do not compromise 
autonomy or cause harm. This can be implemented through various governance models that 
allow for varying degrees of human intervention, from direct involvement in learning cycles 
(monitoring gradient bases methods for solving \eqref{equ_def_emp_risk}) to broader 
oversight of the societal and ethical impacts resulting from the \gls{prediction}s $\hypothesis(\featurevec)$. 

We next discuss how KR1 guides the design choice for \gls{data} (\gls{trainset}), \gls{model} and \gls{loss} 
used in \gls{erm} \eqref{equ_def_emp_risk_min}. 
\subsection{Data} 
\label{sec_KR1_data} 

{\bf Freedom of the Individual.} Ensuring individual freedom demands that individuals, 
especially those at risk of exclusion, have equal access to the benefits and opportunities 
that AI can offer. In this regard, KR1 requires that the \gls{trainset} in \eqref{equ_def_emp_risk} 
is curated with diversity and inclusivity in mind. Biases in \gls{data} collection or labelling 
can disproportionately affect certain groups, potentially limiting their autonomy or 
reinforcing discriminatory patterns. Fair representation of sub-populations in the \gls{dataset} $\dataset$ 
used by \gls{erm} \eqref{equ_def_emp_risk_min} is instrumental for avoiding the 
manipulation of individuals and protecting their mental autonomy and freedom of decision-making.

{\bf Respect for Human Dignity.} Learning personalized \gls{modelparams} for recommender 
systems allows to provide tailored suggestions to users, referred to as micro-targeting. This 
can be useful as it can help users to find suitable contents or products. However, micro-targeting 
can also boost addictive user behaviour or even emotional manipulation of larger user groups \cite{Simchon:2024aa,Kuss:2016aa,Munn:2020aa,NYTMozur2018}. 
KR1 rules out certain design choices for the \gls{label}s of \gls{datapoint}s in order to defuse 
micro-targeting. In particular, we must avoid the mental and psychological characteristics of a user as 
the \gls{label}. KR1 also rules out \gls{lossfunc}s that can be used to train predictors of 
psychological characteristics. 

{\bf Continuous Monitoring.} In its simplest form, \gls{erm}-based methods involve a single 
training phase, i.e., they solve \eqref{equ_def_emp_risk} by some numerical optimization 
method \cite{BoydConvexBook,BertCvxAnalOpt}. Using a single training phase is only useful 
if the data generation is stationary, e.g., if it can be well approximated by an \gls{iidasspt}. 
For many ML applications, this assumption is only realistic if the \gls{trainset} is confined 
to a sufficiently short time period \cite{Dahlhaus2009,Dahlhaus98}. It is then important to 
continuously compute a \gls{valerr} on a timely \gls{valset} which is then used, in turn, to 
diagnose the overall ML system (see \cite[Sec. 6.6]{MLBasics}). Based on the diagnosis, 
the \gls{modelparams} might be updated (re-trained) by using a fresh \gls{trainset} for \gls{erm}.

\subsection{Model} 
 \label{sec_KR1_model}

Human agency and oversight can be facilitated by relying on simple \gls{model}s such as 
\gls{linmodel}s with few \gls{feature}s or \gls{decisiontree}s with small depth. 
It is difficult to state precise criteria for when a model is simple. A more rigorous 
theory of simple models can be developed around quantitative measures for 
their \gls{explainability} (or interpretability). Section \ref{sec_kr4} constructs 
measures for the subjective \gls{explainability} of a trained model $\learnthypothesis \in \hypospace$. 
Roughly speaking, a simple model allows humans to understand how \gls{feature}s 
of a \gls{datapoint} relate to the \gls{prediction} $\hypothesis(\featurevec)$. 

\subsection{Loss}
 \label{sec_KR1_loss}

The choice for the \gls{lossfunc} in \gls{erm} \eqref{equ_def_emp_risk_min} should  
favour hypothesis maps $\hypospace \in \hypospace$ that ensure fundamental rights. 
For example, including a penalty term in the \gls{lossfunc} can force the trained model 
to yield predictions that are invariant across different mental states of the same user. 
We can also explicitly incorporate domain expertise from psychologists to  
penalize predictions that would recommend harmful content to social media users \cite{Lex2021}.   

{\bf Interpretable Loss Function.} To facilitate human oversight, we should use a 
\gls{lossfunc} that can be comprehended by the user. Consider for example a user 
without formal training or education in ML. Here, using, the \gls{zerooneloss} might 
enable human oversight more efficiently compared to using the \gls{logloss} \cite[Sec. 2.3.2]{MLBasics}. 

{\bf Incorporate Human-Centric Objectives.} We can choose a \gls{lossfunc} that includes 
a penalty terms reflecting human-centric values such as fairness (see Section \ref{sec_kr5}) 
or transparency (see Section \ref{sec_kr4}). The idea is to penalize a \gls{hypothesis} $\hypothesis$ 
that delivers \gls{prediction}s $\hypothesis(\featurevec)$ that contradict these goals. 

{\bf Penalizing Unethical Outcomes.} The \gls{lossfunc} in \gls{erm} can be tailored 
to penalize a \gls{prediction} $\hypothesis(\featurevec)$ that would be considered 
unethical. As a case in point, we might assign a very large \gls{loss} value to a prediction 
that results in presenting fake news to a user. 

\section{KR2 - Technical Robustness and Safety} 
\label{sec_techrobustness_erm}

To obtain a practical \gls{erm}-based AI system, we must implement \Gls{erm} \eqref{equ_def_emp_risk} 
by some numerical optimization algorithm that is executed on some computer \cite{OptMLBook}. 
Such an implementation will typically incur a plethora of imperfections, 
ranging from programming errors, quantization noise, power outages, interrupted communication 
links to hardware failures \cite{Sinha:2019aa}.

Assume that we would have a perfect computer that is able to perfectly solve \eqref{equ_def_emp_risk}. 
Still, we must take into account imperfections of the collection process. The \gls{trainset} $\dataset$ 
might be obtained from physical sensors which rarely deliver perfect measurements of 
a physical quantity \cite{LUNDSTROM2020110082}. Moreover, the \gls{trainset} might have 
been intentionally manipulated (poisoned) by an adversary \cite{Kang2020,Jagielski2018}. 

Even if we can rule out any physical measurement errors or data poisoning, it might still be 
useful to consider the \gls{trainset} as being subject to perturbation. Indeed, a key assumption 
of statistical learning theory is that the \gls{trainset} $\dataset$ consists of \gls{iid} samples 
from an underlying probability distribution $p(\pair{\featurevec}{\truelabel})$. 
Thus, we can interpret $\dataset$ as a perturbed representation of $p(\pair{\featurevec}{\truelabel})$. 

It seems natural to require the trained model $\learnthypothesis \in \hypospace$ to be robust 
against perturbations arising from the \gls{iid} sampling process. Indeed, the result of \gls{erm} 
should be a hypothesis with minimum risk, irrespective of the specific realization of the \gls{trainset}. 
For a more detailed analysis of the relation between robustness and generalization of \gls{erm}, 
we refer to \cite{Xu:2012aa,Freiesleben:2023aa} as well as \cite[Sec. 13.2]{ShalevMLBook}. 

To ensure {\bf KR2} we need to understand the effect of perturbations on a \gls{erm}-based 
AI system. These perturbations might affect any of the \gls{erm} components: the \gls{datapoint}s 
in $\dataset$, the model $\hypospace$ or the \gls{loss} function $\loss$. Let us denote 
the perturbed components as $\widetilde{\dataset}$, $\widetilde{\hypospace}$ and $\widetilde{\loss}$. 
The resulting perturbed \gls{erm} is then 
\begin{equation} 
	\label{equ_def_pertrubed_erm}
	\tilde{\hypothesis} = \argmin_{\hypothesis \in \widetilde{\hypospace}} (1/|\widetilde{\dataset}|) \sum_{\pair{\featurevec}{\truelabel} \in \widetilde{\dataset}} \perturbedloss{\pair{\featurevec}{\truelabel}}{\hypothesis}. 
\end{equation} 

The effect of perturbations on optimization problems (such as \eqref{equ_def_emp_risk_min}) 
has been studied extensively in robust optimization literature \cite{RobustOptMLChapter,Ben-Tal:RobustOptimization}. 
By interpreting \gls{erm} as an estimator of (optimal) model parameters allows to use tools 
from robust statistics and signal processing \cite{HuberRobustBook,RobustSigProc}
to study the deviation between \eqref{equ_def_emp_risk_min} and \eqref{equ_def_pertrubed_erm}

The analysis of \eqref{equ_def_pertrubed_erm} is typically based on assuming that the 
perturbed data $\widetilde{\dataset}$, model $\widetilde{\hypospace}$ and \gls{loss} $\widetilde{\loss}$ 
belong to a known uncertainty set $\uncertset$, 
\begin{equation} 
	\big( \widetilde{\dataset}, \widetilde{\hypospace}, \widetilde{\loss} \big) \in \uncertset. 
\end{equation}
Different robustness measures are obtained for different choices for the 
uncertainty set and measures for the deviations between optimization problems. 
For example, the uncertainty set $\uncertset$ might consist of all datasets 
constituted by \gls{datapoint}s within some distance of the \gls{datapoint}s 
in $\dataset$. if we measure the deviation between \eqref{equ_def_emp_risk_min} 
and \eqref{equ_def_pertrubed_erm} in terms of their optimal values, we can 
use basic convex duality to quantify the effect of perturbations \cite[Sec. 5.6]{BoydConvexBook}. 

\subsection{Loss} 

This section discusses specific choices (constructions) for the \gls{lossfunc} in \gls{erm} \eqref{equ_def_emp_risk_min} 
such that its solutions are close to the perturbed \gls{erm} \eqref{equ_def_pertrubed_erm}. In 
particular, we consider uncertainty sets that contain a single choice for model $\hypospace$ 
and \gls{lossfunc} $\loss$ but different perturbed \gls{dataset}s $\widetilde{\dataset}$. 
Thus, throughout this section, we assume \eqref{equ_def_emp_risk_min} and \eqref{equ_def_pertrubed_erm} 
use the same $\hypospace$ and $\loss$. 

{\bf Robust Statistics.} A well-known example for a \gls{loss} function that improves 
robustness of \gls{erm} is the \gls{huberloss}. 
Using the \gls{huberloss} instead of the \gls{sqerrloss} in \eqref{equ_def_emp_risk_min} 
makes the resulting method significantly more robust (or in-sensitive) against 
the presence of outliers in the \gls{trainset} $\dataset$ \cite[Ch. 3]{MLBasics}. Figure \ref{fig_hub_vs_ls} 
depicts a toy \gls{dataset} along with two linear models, one trained by minimizing 
the average \gls{sqerrloss} and another one by minimizing the average \gls{huberloss}. 
\begin{figure} 
	\begin{center}
		\hspace*{-10mm}
\begin{tikzpicture}[scale=0.8]
	\begin{axis}[
		title={},
		xlabel={\gls{feature} $\feature$},
		xlabel style={font=\large},
		ylabel style={at={(axis description cs:0.05,.5)},font=\large},
		ylabel={\gls{label} $\truelabel$},
		legend pos=south east, 
		legend style={font=\large}, 
		xmin=0, xmax=7, 
		ymin = -5, ymax=12, 
		grid=both,
		at={(8cm,0cm)},
		anchor=west,
		width=10cm,
		height=7cm,
		]
		
		\addplot[dashed] table [x=X, y={Linear Regression (with outlier)},col sep=comma] {regression_predictions.csv};
		\addlegendentry{least squares}
		
		\addplot[dashed,mark=o, mark size=1pt,color = red] table [x=X, y={Huber Regression (with outlier)},col sep=comma] {regression_predictions.csv};
		\addlegendentry{Huber}
		
		\addplot[only marks, mark=*, color=blue] coordinates {(1, 1.5) (2, 2.5) (3, 3.5) (4, 4.5) (5, 5.5)};
		\addlegendentry{clean \gls{dataset}}
		
		\addplot[dashed, mark=x, color=orange, line width=2pt, mark size=5pt] coordinates {(6, 10)};
		\addlegendentry{outlier}
		
	\end{axis}
	
\end{tikzpicture}
\end{center}
\caption{\label{fig_hub_vs_ls} Effect of using either squared error or \gls{huberloss} loss on learning the 
	parameters of a linear model. The \gls{modelparams} learnt by minimizing the average \gls{huberloss} 
	seem to be more robust against the presence of an outlier.}
\end{figure}

{\bf Adversarial Loss.} A principled construction of robust loss functions is based on 
replacing \gls{erm} \eqref{equ_def_emp_risk_min} with an adversarial (or worst-case) 
variant  \cite{madry2018towards,Goodfellow2015}
\begin{equation} 
	\label{equ_def_advers_loss_generic}
	\widehat{\hypothesis} = \argmin_{\hypothesis \in \hypospace} \sup_{\widetilde{\dataset} \in \uncertset} \sum_{\pair{\featurevec}{\truelabel} \in \widetilde{\dataset}} \lossfunc{\pair{\featurevec}{\truelabel}}{\hypothesis}. 
\end{equation} 
A rapidly growing body of work studies various instances of \eqref{equ_def_advers_loss_generic} obtained 
for different constructions of the uncertainty set $\uncertset$ \cite{Rice2020,DBLP:conf/iclr/WongRK20,Javanmard2020,JMLR:v20:17-633}. 
We next show how a special case of \eqref{equ_def_advers_loss_generic} is 
equivalent to \eqref{equ_def_emp_risk_min} for a suitable choice $\loss'$ for the \gls{lossfunc}. 

One widely used construction of the uncertainty set in \eqref{equ_def_advers_loss_generic} is to separately 
perturb the features (and potentially also the labels) of \gls{datapoint}s in $\dataset$ \cite{Javanmard2020,Rice2020}. Thus, 
the uncertainty set decomposes into one separate uncertainty set $\uncertset^{\pair{\featurevec}{\truelabel}}$ for each 
data point $\pair{\featurevec}{\truelabel}$ in $\dataset$. The adversarial \gls{erm} \eqref{equ_def_advers_loss_generic} then 
becomes  \cite{Javanmard2020}
\begin{equation}
	\label{equ_def_advers_loss}
	\widehat{\hypothesis} = \argmin_{\hypothesis \in \hypospace} \sum_{\sampleidx=1}^{\samplesize} \underbrace{\sup_{\pair{\widetilde{\featurevec}}{\tilde{\truelabel}} \in \uncertset^{\pair{\featurevec^{(\sampleidx)}}{\truelabel^{(\sampleidx)}}}} \lossfunc{\pair{\widetilde{\featurevec}}{\tilde{\truelabel}}}{\hypothesis}}_{\mbox{ robust loss } L'\big( \pair{\featurevec^{(\sampleidx)}}{\truelabel^{(\sampleidx)}}, \hypothesis \big)}. 
\end{equation} 
Note that the robust \gls{lossfunc} $\loss'$ in \eqref{equ_def_advers_loss} depends on both, the original choice for the \gls{lossfunc} in \eqref{equ_def_emp_risk_min} as well as the uncertainty set $\uncertset$ in \eqref{equ_def_advers_loss}. 

Let us next consider a modification of \eqref{equ_def_advers_loss} where we perturb 
only the \gls{feature}s of \gls{datapoint}s but leaving their \gls{label}s untouched \cite{RobustOptMLChapter}. 
This modification uses an uncertainty set $\uncertset^{(\eta)}$ which is parametrized by a perturbation 
strength $\eta$ and consists of datasets $\widetilde{\dataset} = \pair{\widetilde{\featuremtx}}{\labelvec}$ 
with \gls{featuremtx} 
\begin{align} 
	\label{equ_def_uncert_set_coupled}
	\underbrace{\widetilde{\featuremtx}}_{ \defeq \big(\widetilde{\featurevec}^{(1)},\ldots,\widetilde{\featurevec}^{(\samplesize)}\big)^{T}}& = \underbrace{\featuremtx}_{\defeq \big(\featurevec^{(1)},\ldots,\featurevec^{(\samplesize)}\big)^{T}} + \big( \vu^{(1)},\ldots,\vu^{(\nrfeatures)} \big) \nonumber \\
	&  \mbox{ with } \normgeneric{\vu^{(\featureidx)}}{1} \leq  \eta. 
\end{align}

Carefully note that, in contrast to \eqref{equ_def_advers_loss}, the construction \eqref{equ_def_uncert_set_coupled} 
couples the features of different \gls{datapoint}s in $\dataset$. Thus, instead of maximizing over possible perturbations 
separately for each data point as in \eqref{equ_def_advers_loss}, we need to study the worst-case perturbation 
of the entire dataset: 
\begin{equation}
	\label{equ_def_advers_loss_coupled}
	\widehat{\hypothesis} = \argmin_{\hypothesis \in \hypospace}\sup_{\widetilde{\dataset} \in \uncertset} \sum_{\sampleidx=1}^{\samplesize} \lossfunc{\pair{\widetilde{\featurevec}^{(\sampleidx)}}{{\truelabel}}}{\hypothesis}.
\end{equation} 
Consider \gls{erm} obtained for a \gls{linmodel} and the absolute error loss. 
Here, it can be shown that \eqref{equ_def_advers_loss_coupled} is 
equivalent to \gls{erm} with the robust \gls{loss} $\loss' = \big| \truelabel - \hypothesis(\featurevec) \big|+ \eta \normgeneric{\weights}{1}$ \cite[Thm. 14.9.]{RobustOptMLChapter}. 

Recent work also studies uncertainty sets $\uncertset$ that consist of perturbed datasets 
$\widetilde{\dataset}$ with an empirical distribution $\widetilde{\mathbb{P}}$ close to the 
empirical distribution $\mathbb{P}$ of $\dataset$, 
\begin{equation}
	\label{equ_def_uncert_WS}
	\uncertset^{(\eta)} \defeq \big\{  \widetilde{\dataset}: W\big( \widetilde{\mathbb{P}}, \mathbb{P} \big) \leq \eta \big\}. 
\end{equation}  
Here,  $W\big( \widetilde{\mathbb{P}}, \mathbb{P} \big)$ denotes the Wasserstein 
distance between $\widetilde{\mathbb{P}}$ and $\mathbb{P}$ \cite{JMLR:v20:17-633}. 

Consider the adversarial \gls{erm} \eqref{equ_def_advers_loss_generic} with uncertainty 
set \eqref{equ_def_uncert_WS} and a \gls{lossfunc} $\loss$ that is Lipschitz continuous 
with modulus $\alpha$. It can then be shown that \eqref{equ_def_advers_loss_generic} 
is equivalent to \gls{erm} with a specific robust \gls{lossfunc} \cite[Theorem 4]{JMLR:v20:17-633}. 

For a binary classification, the authors of \cite{Bhattacharjee2022RobustER} study a 
robust \gls{loss} of the form 
\begin{equation}
	\label{equ_def_robust_loss_binary_classif}
	\lossfunc{\pair{\featurevec}{\truelabel}}{\hypothesis} = \begin{cases} 1 & \mbox{ if } \hypothesis(\featurevec') \neq \truelabel \mbox{ for some } \featurevec' \in \mathcal{U}_{\featurevec} \\ 
		 0 & \mbox{ otherwise.} \end{cases}
\end{equation}
Here, $\mathcal{U}_{\featurevec}$ is some robustness region. Note that \eqref{equ_def_robust_loss_binary_classif} 
reduces to the basic \gls{zerooneloss} for the choice $\mathcal{U}_{\featurevec} = \{ \featurevec \}$ \cite{MLBasics}. 


\subsection{Data} 

Instead of choosing a robust \gls{lossfunc} $\loss$ in \gls{erm} \eqref{equ_def_emp_risk_min}, 
we can construct the \gls{trainset} in \eqref{equ_def_emp_risk_min} to make its solutions more 
robust. One widely studied approach is adversarial training, i.e., to include adversarially perturbed 
\gls{datapoint}s in the \gls{trainset} $\dataset$\cite{Goodfellow2015,Wang2023Adversarial,cheng2023adversarial}. 

An opposite approach to adversarial training is to prune a given \gls{dataset} 
using some form of \gls{outlier} detection \cite{Steinhardt2017}. The \gls{trainset} is 
then obtained by the remaining \gls{datapoint}s that have not been declared as \gls{outlier}s. 
However, it can be challenging to distinguish outliers from natural perturbations due to 
the sampling from a true underlying \gls{probdist} \cite{10.1145/324133.324221,pmlr-v202-lu23e}. 

The fundamental limits for outlier removal techniques can be studied using a 
malicious noise model \cite{doi:10.1137/0222052}: 
\begin{align}
	\vz^{(\sampleidx)} & = \begin{cases} \widetilde{\vz}^{(\sampleidx)} & \mbox{ if } b^{(\sampleidx)} = 1 \\  \mathbf{e}^{(\sampleidx)} & \mbox{ otherwise,} \end{cases}  \\ 
	& \mbox{ with } b^{(\sampleidx)} \stackrel{\mbox{\gls{iid}}}{\sim} \mathcal{B}(p_{e}), \widetilde{\vz}^{(\sampleidx)}  \stackrel{\mbox{\gls{iid}}}{\sim} p\pair{\featurevec}{\truelabel}. 
\end{align} 
Here, the outlier $\mathbf{e}^{(\sampleidx)}$ can be chosen arbitrarily (maliciously), even taking 
into account the current state of the optimization method used to solve \eqref{equ_def_emp_risk_min}. 
Consider a \gls{erm} method for binary classification, delivering a \gls{hypothesis} $\learnthypothesis$ 
with expected $0/1$ \gls{loss} $\expect \big\{ \lossfunc{\datapoint}{\learnthypothesis} \big\}$. 
In order to allow for the existence of \gls{erm} method achieving $\expect \big\{ \lossfunc{\datapoint}{\learnthypothesis} \big\} < \varepsilon$, the maximum fraction of outliers that can be tolerated is upper bounded by $\varepsilon/(1+\varepsilon)$ \cite{doi:10.1137/0222052}.

\subsection{Model} 

We can define and measure robustness of ML using different notions of 
continuity of the learnt hypothesis $\learnthypothesis$. Beside 
the basic qualitative notion of continuity we can also use Lipschitz continuity 
to obtain a quantitative measure of robustness \cite{RudinBookPrinciplesMatheAnalysis}. 
Note that Lipschitz continuity requires both, the domain as well as the range of the \gls{hypothesis} 
map $\learnthypothesis$, to be a metric space. 

One obvious way to  ensure robustness of \gls{erm} is to use a \gls{model} $\hypospace$ 
that only contains Lipschitz continuous hypothesis maps $\hypospace$. Recent 
work has shown that \gls{erm} delivers a Lipschitz continuous hypothesis 
if the model $\hypospace$ is sufficiently large \cite{Bubeck2023}. 

Instead of Lipschitz continuity, the authors of \cite{pmlr-v139-leino21a} use the 
concept of local and global robustness for multi-class classification problems. 
Here, \gls{datapoint}s have a label $\truelabel \in \labelspace \defeq \{1,\dots,\nrcategories\}$ 
and the goal is to learn a classifier $\hypothesis(\featurevec) = \big( \hypothesis_{1}(\featurevec),\ldots,\hypothesis_{\nrcluster}(\featurevec) \big)^{T}$ 
which is used to classify a \gls{datapoint} as $\hat{\truelabel} = \argmax_{\clusteridx\in\{1,\ldots,\nrcategories\}} \hypothesis_{\clusteridx}(\featurevec)$.  

A classifier $\hypothesis(\featurevec)$ is then defined as  $\varepsilon$-locally robust at 
feature vector $\featurevec$ if it classifies $\predictedlabel = \predictedlabel'$ for every 
\gls{datapoint} with features $\featurevec'$ such that $\normgeneric{\featurevec - \featurevec'}{2} \leq \varepsilon$ \cite{pmlr-v139-leino21a} . 
Note that if we require this to hold at every $\featurevec$, the classifier must be trivial (delivering the same 
label value for every \gls{datapoint}). To obtain a useful notion of global robustness, the authors of \cite{pmlr-v139-leino21a} 
introduce an auxiliary label value that signals if the classifier fails to be robust locally.

\section{KR3 - Privacy and Data Governance} ``\emph{..privacy, a fundamental right 
	particularly affected by AI systems. Prevention of harm to privacy also necessitates adequate data 
	governance that covers the quality and integrity of the data used...}''\cite[p.17]{HLEGTrustworhtyAI}.

{\bf Data Governance.} Many applications of \gls{erm} involve \gls{datapoint}s generated by human users, 
thus constituting personal data. KR3 emphasizes the protection of personal data 
throughout the entire lifecycle of an ERM-based AI system, from initial data collection 
and \gls{model} training to the final deletion of any personal information. 
Effective data governance practices must ensure data quality control, such as 
verifying factual accuracy and completeness \cite{DatasheetData2021}. 
When handling personal data, special attention to data protection regulations 
is essential \gls{gdpr}. This often involves appointing a data protection officer 
and to conduct a data protection impact assessment \cite{DPC_DPIA}.  

{\bf Measuring Privacy Leakage.} Ensuring \gls{privprot} for an \gls{erm}-based system 
requires some means to quantify its privacy leakage. To this end, it is useful to think of an 
\gls{erm}-based method as a map $\algomap$: An \gls{erm}-based method $\algomap$ 
reads in the \gls{trainset} $\dataset$, solves \eqref{equ_def_emp_risk}, and delivers some 
output $\algomap(\dataset)$. The output could be the learnt \gls{modelparams} $\widehat{\weights}$ 
or the prediction $\learnthypothesis(\featurevec)$ obtained for a specific \gls{datapoint} 
with \gls{feature}s $\featurevec$. 

{\bf \Gls{privprot} requires non-invertibility.} To implement means of 
\gls{privprot}, we need to clarify what parts of a \gls{datapoint} 
are considered private or sensitive information. To fix ideas, consider \gls{datapoint}s 
representing humans. Each \gls{datapoint} is characterized by \gls{feature}s $\featurevec$, 
potentially a \gls{label} $\truelabel$ and a \gls{sensattr} $\sensattr$ (e.g., a recent medical diagnosis). 
For a \gls{erm}-based method $\algomap$, \gls{privprot} means that it should be 
impossible to infer, from the output $\algomap(\dataset)$, any of the \gls{sensattr}s
$\sensattr$ in $\dataset$. Mathematically, \gls{privprot} requires non-invertibility 
of the map $\algomap(\dataset)$. In general, just making $\algomap(\dataset)$ non-invertible 
is typically insufficient for \gls{privprot}. We need to make $\algomap(\dataset)$ 
sufficiently non-invertible. 


{\bf \Gls{diffpriv}.} One widely used approach to make a \gls{erm}-based method sufficiently 
non-invertible is introduce some randomness or noise. Examples for such randomness include the 
adding of noise to the output and the selection of a random subset of $\dataset$. 
The map $\algomap$ then becomes stochastic and, in turn, the output $\algomap(\dataset)$ is 
then  characterized by a \gls{probdist} ${\rm Prob} \big\{ \algomap(\dataset) \in \mathcal{S} \}$ for all sets $\mathcal{S}$ within a 
well-defined collection of measurable sets \cite{BillingsleyProbMeasure}. 

\Gls{diffpriv} measures  the non-invertibility of a stochastic algorithm $\algomap$ via 
the similarity of the \gls{probdist}s obtained for two \gls{dataset}s  $\dataset,\dataset'$ that 
are considered neighbouring or adjacent \cite{NISTDiffPriv2023,AlgoFoundDP}. Typically, 
we consider $\dataset'$ to be adjacent to $\dataset$ if it is obtained by modifying the 
\gls{feature}s or \gls{label} of a single \gls{datapoint} in $\dataset$. In general, the notion 
of neighbouring \gls{dataset}s is a design choice used in the formal definition of \gls{diffpriv}. 
\begin{definition}
	\label{equ_def_dp}
	(from \cite{AlgoFoundDP}) A \gls{erm}-based method $\algomap$ is $(\varepsilon,\delta)$-\gls{diffpriv} if 
	for any measurable set $\mathcal{S}$ and any two neighbouring \gls{dataset}s $\dataset, \dataset'$, 
	\begin{equation} 
		\label{equ_def_dp_prob}
		{\rm Prob} \big\{ \algomap(\dataset) \in \mathcal{S} \} \leq \exp(\varepsilon) 	{\rm Prob} \big\{ \algomap(\dataset') \in \mathcal{S} \} + \delta. 
	\end{equation} 
\end{definition}

\subsection{Data} 
\label{sec_kr3_data} 

One simple way to implement \gls{privprot} in \gls{erm}-based methods 
is by careful selection of the \gls{feature}s used to characterize \gls{datapoint}s \cite{Zohra2024,Sartor2020GDPR_AI}. 
The idea is to use only \gls{feature}s that are relevant for the learning task but 
at the same time do not convey too much information about any \gls{sensattr}. 

There is an inherent trade-off between \gls{privprot} and resulting statistical 
accuracy. Indeed, we trivially obtain perfect \gls{privprot} by not using any 
property of a \gls{datapoint} as their \gls{feature}s. Note, however, this extreme 
case of maximum \gls{privprot} comes at the cost of a lower quality of the 
\gls{prediction}s delivered by (the \gls{hypothesis} learnt from) \gls{erm}. 

{\bf Private Feature Learning.} In general, it is difficult to manually identify 
\gls{feature}s that strike a good balance between \gls{privprot} and predictive 
accuracy.\footnote{Think of \gls{datapoint}s being images, each characterized by millions 
	of pixel colour intensities as their raw \gls{feature}s.} We could then try learn, 
	in a data-driven fashion, a \gls{feature} map $\featuremapvec: \mathbb{R}^{\dimlocalmodel} \rightarrow \mathbb{R}^{\dimlocalmodel'}$. 
The map $\featuremapvec$ is learnt such that the new \gls{feature}s $\vz = \featuremapvec(\featurevec) \in \mathbb{R}^{\dimlocalmodel'}$ 
do not allow to infer (accurately) the private attribute $\privattr$ while still allowing 
to predict the \gls{label} $\truelabel$ of a \gls{datapoint}. 

We next discuss two specific approaches to private \gls{featlearn}. These two 
approaches differ in how they measure the predictability of $\privattr$ and 
$\truelabel$. Both measures are based on a simple \gls{probmodel} for 
the \gls{datapoint}s in $\dataset$, interpreting them as \gls{realization}s of \gls{iid} \gls{rv}s. 
The first approach, referred to as the \gls{privfunnel}, measures predicability 
of the $\privattr$ using \gls{mutualinformation}. The second approach uses the 
minimum achievable (by linear maps) expected \gls{sqerrloss} as measure for 
predicability.

{\bf Privacy Funnel.}  The \gls{mutualinformation} $\mutualinformation{\privattr}{\featuremapvec(\featurevec)}$ 
can be used as a measure for the predicability of $\privattr$ from $\featuremapvec(\featurevec)$. 
A small value of $\mutualinformation{\privattr}{\featuremapvec(\featurevec)}$ 
indicates that it is difficult to predict the private attribute $\privattr$ solely from $\featuremapvec(\featurevec)$, i.e., a 
high level of \gls{privprot}.\footnote{The relation between \gls{mutualinformation}-based 
	privacy measures 
	and \gls{diffpriv} has been studied in some detail \cite{DPasMIDP2016}.} Similarly, we can 
	use the \gls{mutualinformation} $\mutualinformation{\truelabel}{\featuremapvec(\featurevec)}$ 
to measure the predicability of the label $\truelabel$ from $\featuremapvec(\featurevec)$. A large value 
$\mutualinformation{\truelabel}{\featuremapvec(\featurevec)}$ indicates that $\featuremapvec(\featurevec)$ 
allows to accurately predict $\truelabel$ (which is of course preferable). 

It seems natural to use a \gls{feature} map $\featuremapvec(\featurevec)$ that optimally 
balances a small $\mutualinformation{s}{\featuremapvec(\featurevec)}$ (stronger \gls{privprot}) 
with a sufficiently large $\mutualinformation{\truelabel}{\featuremapvec(\featurevec)}$ (allowing 
to accurately predict $\truelabel$). The mathematically precise formulation of this 
plan is known as the privacy funnel \cite[Eq. (2)]{PrivacyFunnel}, 
\begin{equation}
	\label{equ_def_privacy_funnel}
	\min_{\featuremapvec(\cdot)}  \mutualinformation{s}{\featuremapvec(\featurevec)} \mbox{ such that } \mutualinformation{\truelabel}{\featuremapvec(\featurevec)}\geq R.
\end{equation} 
Figure \ref{fig_illustrate_priv_funnel} qualitatively illustrates the solution of \eqref{equ_def_privacy_funnel} 
for varying threshold $R$.
\begin{figure}
	\begin{center} 
		\begin{tikzpicture}[yscale=1]
			\begin{axis}[
				xlabel={$\mutualinformation{\truelabel}{\featuremapvec(\featurevec)}$},
				ylabel={$\mutualinformation{\privattr}{\featuremapvec(\featurevec)}$},
				thick,
				ymax=4,
				ymin=0,
				xmin=0,
				xmax=7,
				domain=0:7,
				ylabel near ticks,
				xlabel near ticks,
				axis lines=left, 
				xtick=\empty, 
				ytick=\empty, 
				]
				\addplot[smooth, blue, mark=none] {4*exp(-0.3*(7-x))-4*exp(-0.3*(7))} node[pos=0.24, anchor=north west] {\hspace*{3mm}};
				
			\end{axis}
			\vspace*{-4mm}
		\end{tikzpicture}
		\vspace*{-3mm}
	\end{center} 
	\caption{\label{fig_illustrate_priv_funnel} The solutions of the privacy funnel \eqref{equ_def_privacy_funnel} 
		trace out (for varying constraint $R$ in \eqref{equ_def_privacy_funnel}) a curve in the plane spanned by the 
		values of $\mutualinformation{\privattr}{\featuremapvec(\featurevec)}$ (measuring the \gls{privleakage}) and $\mutualinformation{\truelabel}{\featuremapvec(\featurevec)}$ 
		(measuring the usefulness of the transformed \gls{feature}s for predicting the \gls{label}).}
\end{figure}

{\bf Private Linear Feature Learning.} 
The privacy funnel \eqref{equ_def_privacy_funnel} uses the \gls{mutualinformation} $\mutualinformation{\privattr}{\featuremapvec(\featurevec)}$ 
to quantify the privacy leakage of a \gls{featuremap} $\featuremapvec(\featurevec)$. An 
alternative measure for the privacy leakage is the minimum reconstruction error $\privattr - \hat{\privattr}$. 
The reconstruction $\hat{\privattr}$ is obtained by applying a map $r(\cdot)$ to the 
transformed \gls{feature}s $\featuremapvec(\featurevec)$. If the joint \gls{probdist} $p(\privattr,\featurevec)$ 
is a \gls{mvndist} and the $\featuremapvec(\cdot)$ is a linear map (of the form $\featuremapvec(\featurevec) \defeq \mathbf{F} \featurevec$ with some matrix $\mathbf{F}$), then the optimal reconstruction map $r(\cdot)$ 
is again linear \cite{LC}. 

\begin{figure}[htbp]
	\begin{center}
		\begin{tikzpicture}[scale = 1.1]
			\node at (4.3,2) (mp) {\includegraphics[width=24mm]{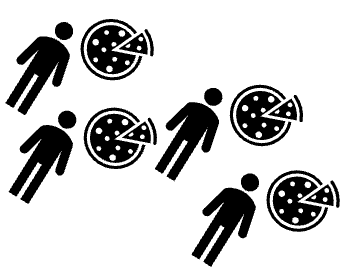}};
			\node at (1.0,1.0)(fa) {\includegraphics[width=24mm]{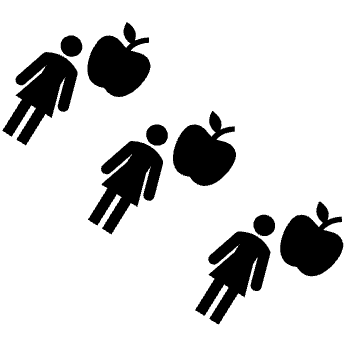}};
			\node at (3.5,0.0) (fp){\includegraphics[width=24mm]{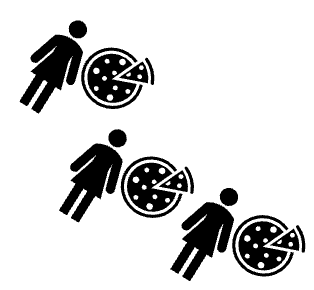}};
			\node at (2,3) (ma){\includegraphics[width=24mm]{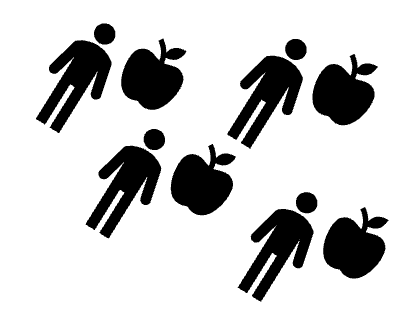}};
			\draw[<->, line width=1mm, color=green!50!black] (3.0,3.9) -- (4.4,3.0) node[above=9mm](y) {\hspace*{15mm} food preference $\truelabel$}; 
			\draw[<-, line width=1mm, color=green!50!black] (-0.5,0.7) -- (0.9,-0.2) node[above=0mm](y1) {\hspace*{0mm} $\mathbf{f}$}; 
			\draw[<->, line width=1mm, color=red] (4.7,0.5) -- (5.7,1.9) node[below=6mm](s) {\hspace*{20mm} gender $\privattr$}; 
			\draw[->, very thick,color=blue!60!black] (0,0.3) -- (6.5,0.3) node[right](x) {$\feature_1$};       
			\draw[->, very thick,color=blue!60!black] (0.3,0) -- (0.3,4.2) node[above] {$\feature_2$};   
		\end{tikzpicture}
		\vspace*{-3mm}
	\end{center}
	\vspace*{-5mm}
	\caption{A toy \gls{dataset} $\dataset$ whose \gls{datapoint}s represent customers, each 
		characterized by \gls{feature}s $\featurevec = \big(\feature_{1},\feature_{2}\big)^{T}$. 
		These raw \gls{feature}s carry information about a \gls{sensattr} $\privattr$ (gender) and 
		the \gls{label} $\truelabel$ (food preference) of a person. The \gls{scatterplot} suggests 
		that we can find a linear \gls{feature} transformation $\mathbf{F} \defeq \mathbf{f}^{T} \in \mathbb{R}^{1 \times 2}$ 
		resulting in a new \gls{feature} $z \defeq \mathbf{F} \featurevec$ that does not allow to predict $\privattr$, 
		while still allowing to predict $\truelabel$. \label{fig_pp_feature_learning_lin}}
\end{figure}

We would like to find the linear \gls{featuremap} $\featuremapvec(\featurevec) \defeq \mathbf{F} \featurevec$ such 
that for any linear reconstruction map $\mathbf{r}$ (resulting in $\hat{s} \defeq \mathbf{r}^{T} \mathbf{F} \featurevec$) 
the expected squared error $\expect \{ (s - \hat{s})^2 \}$ is large. The minimal 
expected \gls{sqerrloss} 
\begin{equation} 
	\varepsilon(\mathbf{F}) \defeq \min_{\mathbf{r} \in \mathbb{R}^{\dimlocalmodel'}} \expect \{ (s - \mathbf{r}^{T} \mathbf{F} \featurevec)^2 \}
\end{equation}
measures the level of \gls{privprot} offered by the new \gls{feature}s $\vz = \mathbf{F} \featurevec$. 
The larger the value $\varepsilon(\mathbf{F})$, the more \gls{privprot} is offered. 
It can be shown that $\varepsilon(\mathbf{F})$ is maximized by any matrix $\mathbf{F}$ whose 
rows are orthogonal to the cross-covariance vector $\mathbf{c}_{\featurevec,s} \defeq \expect \{ \featurevec s \}$, i.e., 
whenever $\mathbf{F}\mathbf{c}_{\featurevec,s} = \mathbf{0}$. One specific choice 
for $\mathbf{F}$ that satisfies this orthogonality condition is 
\begin{equation} 
	\label{equ_pp_linear_feature_map} 
	\mathbf{F}=\mathbf{I} - (1/\normgeneric{\mathbf{c}_{\featurevec,s}}{2}^{2}) \mathbf{c}_{\featurevec,s} \mathbf{c}_{\featurevec,s}^{T}. 
\end{equation} 
Figure \ref{fig_pp_feature_learning_lin} illustrates a \gls{dataset} for which we want 
to find a linear \gls{featuremap} $\mathbf{F}$ such that the new \gls{feature}s $\vz = \mathbf{F} \featurevec$ 
do not allow to accurately predict a \gls{sensattr}.   

{\bf Sufficient Statistics.} So far, we have discussed \gls{privprot} in the 
sense of not allowing to predict \gls{sensattr}s. In some applications, it might 
not be clear what a \gls{sensattr}s is. Still we would like to minimize any 
potential privacy leakage. To implement such a \gls{dataminprinc} we can use the 
concept of a sufficient statistic \cite{Pitman1936,RKHSExpFamIT2012}. To this end, 
we assume that \gls{datapoint}s are obtained as \gls{iid} samples from a \gls{probdist} 
$\prob{\featurevec;\weights}$ which is parametrized by \gls{modelparams} $\weights$. 

\gls{erm}-based methods can be interpreted as methods for estimating the true 
underlying $\weights$ of the \gls{probdist} $\prob{\featurevec;\weights}$. 
A statistic $\vz = \featuremapvec(\featurevec)$, with some map $\featuremapvec(\cdot)$, 
is sufficient for the parameter $\weights$ if the conditional \gls{probdist} of $\featurevec$, 
given the statistic $\vz = \featuremapvec(\featurevec)$, does not depend on the \gls{modelparams} $\weights$. 

Whenever we have identified a sufficient statistic for the \gls{probmodel} $\prob{\featurevec;\weights}$, 
we can safely discard the original raw \gls{feature}s and instead use the sufficient statistic $\vz = \featuremapvec(\featurevec)$ 
as the new \gls{feature}s. Of particular interest are sufficient statistics that are minimal in the sense 
that any other sufficient statistic is determined from known the value of a minimal sufficient statistic \cite{LC}.

\subsection{Model} 
\label{sec_kr3_model}

The \gls{featlearn} techniques from the above Section \ref{sec_kr3_data} 
can also be implemented as a design choice for the \gls{model} used in \gls{erm}. 
Indeed, we can think of linear feature learning map as being a pre-processing step 
within a \gls{hypothesis} map. 

Remember that \gls{privprot} of an \gls{erm}-based method $\algomap(\dataset)$ 
is determined by its non-invertibility. Let us next illustrate the impact of the choice for 
the \gls{model} $\hypospace$ on the non-invertibility of $\algomap(\dataset)$. 
Figure \ref{fig_dec_regions_privprot} depicts a toy \gls{dataset} $\dataset$ along 
with the \gls{decisionregion}s of the \gls{hypothesis} $\learnthypothesis$ 
learnt with \gls{erm} \eqref{equ_def_emp_risk_min} using a \gls{decisiontree} 
\gls{model} $\hypospace$ \cite[Ch. 3]{MLBasics}. 

Note that one of the \gls{decisionregion}s depicted in Figure \ref{fig_dec_regions_privprot} 
contains a single \gls{datapoint}, denoted $\pair{\featurevec^{(1)}}{\truelabel^{(1)}}$, from $\dataset$. 
Thus, if we have a sufficiently accurate estimate for the \gls{feature}s $\featurevec^{(1)}$, we can infer the
\gls{label} $\truelabel^{(1)}$ by observing the \gls{prediction}s delivered by $\learnthypothesis$ 
for \gls{feature}s near-by $\featurevec^{(1)}$. To avoid such a \gls{model} inversion attack, we 
should use a more shallow \gls{decisiontree} \gls{model} such that each resulting \gls{decisionregion} 
contains a minimum number of \gls{datapoint}s from $\dataset$.

\begin{figure}[htbp]
\begin{tikzpicture}
	\begin{axis}[
		width=8cm,
		height=8cm,
		axis lines=middle,
		xlabel={$x_1$},
		ylabel={$x_2$},
		xmin=-0.1, xmax=10.1,
		ymin=-0.1, ymax=10.1,
		legend pos=north east
		]
	\addplot[only marks, mark=*, mark size=3pt, blue] coordinates {(2,2) (4,8) (8,4)};
	\addplot[only marks, mark=square*, mark size=3pt, red] coordinates {(5,6) (7,9) (6,1)};
	
	\node at (axis cs: 2.8,2) [anchor=south east] {$\pair{\featurevec^{(1)}}{\truelabel^{(1)}}$}; 
	

	\draw[dashed, red, thick] (axis cs: 3,0) -- (axis cs: 3,10); 
	\draw[dashed, red, thick] (axis cs: 0,6.5) -- (axis cs: 10,6.5); 
	\end{axis}
\end{tikzpicture}
\caption{\Gls{scatterplot} of a \gls{dataset} $\dataset$ along with the \gls{decisionboundary} of a \gls{decisiontree} $\learnthypothesis$ 
	trained via \gls{erm} on $\dataset$. One of the \gls{decisionregion}s contains a single \gls{datapoint} from the 
	\gls{trainset} which could allow an adversary to infer the \gls{label} $\truelabel^{(1)}$ from the \gls{prediction}s $\learnthypothesis(\featurevec)$ obtained for \gls{feature}s near-by $\featurevec^{(1)}$. \label{fig_dec_regions_privprot}}
\end{figure}

\subsection{Loss} 
\label{sec_kr3_loss} 
We can also ensure \gls{privprot} in \gls{erm}-based AI systems via suitable 
design choice for the \gls{lossfunc} $\loss$ \eqref{equ_def_emp_risk_min}. 
As a (not very useful) extreme case, consider a constant \gls{lossfunc} $\lossfunc{\pair{\featurevec}{\truelabel}}{\weights} = 0$. 
Here, the \gls{hypothesis} learnt by \gls{erm} \eqref{equ_def_emp_risk_min} is 
totally unrelated to the \gls{datapoint}s in the \gls{trainset} $\dataset$ and, in turn, 
does not carry any information about them (in particular, their \gls{sensattr}s). 
This maximal \gls{privprot} comes at the cost of learning a useless \gls{hypothesis} 
in general. 

A less trivial construction for a privacy-friendly \gls{lossfunc} is studied in \cite{DPERM}. 
Given the \gls{lossfunc} of a potentially non-private \gls{erm}-based method, we simply 
add a random linear function to the objective function in \eqref{equ_def_emp_risk_min}.  
The authors of \cite{DPERM} then study \gls{diffpriv} guarantees \eqref{equ_def_dp_prob} 
guaranteed by using the randomly perturbed \gls{erm}.

\section{KR4 - Transparency} 
\label{sec_kr4} 

According to \cite{HLEGTrustworhtyAI}, this key requirement encompasses \gls{transparency} 
of elements relevant to an AI system. KR4 requires \gls{erm}-based methods to provide \gls{explanation}s 
for their \gls{prediction}s. Instead of constructing explicit \gls{explanation}s, \gls{erm}-based 
methods should utilize \gls{model}s that are intrinsically interpretable \cite{rudin2019stop,Zhang:2024aa}. 
	
KR4 also mandates that users be informed when interacting with an automated 
system, e.g., through notifications such as 'You are now conversing with a chatbot'. 
In addition, \gls{erm}-based methods should be transparent about their capabilities 
and limitations, including quantitative measures of prediction uncertainty.

{\bf Traceability.} The design choices (and underlying business models) for a \gls{erm}-based AI systems 
must be documented. This includes the source for the (\gls{datapoint}s in the) 
\gls{trainset}, the \gls{model}, the \gls{lossfunc} used in \eqref{equ_def_emp_risk} \cite{DatasheetData2021}. 
Moreover, the documentation should also cover the details of the optimization method 
used to solve \eqref{equ_def_emp_risk}. This documentation might include the recording of the 
current \gls{modelparams} along with a time-stamp (``logging''). 

{\bf Communication.} The user interface of an AI system must clearly indicate if it delivers responses 
based on automated data processing such as \gls{erm}. AI systems also need to communicate the 
capabilities and limitations to their end users (e.g., of a digital health app running on a smartphone). 
For example, we can indicate a measure of uncertainty about the \gls{prediction}s delivered by the 
trained \gls{model}. Such an uncertainty measure can be obtained naturally from \gls{probmodel} for 
the data, e.g., the (estimated) conditional variance of the label $\truelabel$, given the \gls{feature}s 
$\featurevec$ of a random \gls{datapoint}. Another example for an uncertainty measure is the \gls{valerr} 
of a trained \gls{model} $\learnthypothesis \in \hypospace$. 

{\bf \Gls{explainability}.} Another core aspect of \gls{transparency} is the \gls{explainability} 
of an AI system. In what follows, we will discuss how specific design choices can facilitate 
the \gls{explainability} of \gls{erm}-based methods. To this end, we need a precise 
definition or quantitative measure for the \gls{explainability} of \gls{erm}. There is a variety 
of approaches to constructing numeric measures for \gls{explainability} of \gls{erm} based 
methods \cite{Schwalbe:2024aa}. One recent line of work revolves around the notion of simulatability \cite{doshivelez2017rigorousscienceinterpretablemachine,hase-bansal-2020-evaluating,Chen2018,Colin:2022aa,JunXML2020,Zhang:2024aa}. A key challenge in meeting KR4 is the subjective nature of \gls{explainability}, 
as the clarity of \gls{explanation}s can vary depending on the user's perspective 
\cite{JunXML2020,Zhang:2024aa}.\footnote{As a case in point, a \gls{linmodel} for predicting 
	a disease based on several bio-physical measurements might be explainable for a medical expert. 
	However, it might not be explainable to an elementary school student.}

{\bf Simulatability.} It seems natural to consider an \gls{erm}-based method explainable 
to a specific user if they can anticipate (or predict) the predictions delivered by the trained \gls{model} $\learnthypothesis \in \hypospace$ \cite{hase-bansal-2020-evaluating,doshivelez2017rigorousscienceinterpretablemachine,fel:hal-03473101}. 
Consider some \gls{testset} $\testset$ that consists of unlabeled \gls{datapoint}s, 
each characterized by some features $\featurevec$. We further assume that we 
have access to the labels $\user(\featurevec)$ predicted by a user \cite{10.1145/3287560.3287596}. 

{\bf Objectivity vs. Subjectivity.} We can measure the (lack of) \gls{explainability} of a 
trained \gls{model} $\learnthypothesis \in \hypospace$ via the discrepancy 
between its \gls{prediction}s $\learnthypothesis(\featurevec)$ and the user 
\gls{prediction}s $\user(\featurevec)$. This results in a subjective 
\gls{explainability} as it is based on the (subjective) \gls{prediction}s $\user(\featurevec)$ 
provided by a specific user. This approach also allows for different levels of 
objectivity (or subjectivity) by using increasingly large user groups to aggregate 
the user predictions for the \gls{datapoint}s. Roughly speaking, instead of having 
a user prediction from a single user, such as the co-author \emph{A. Jung} 
of this work, we instead aggregate the user predictions from a larger 
group of users such as \emph{Austrian males}. Manually curated (labelled) benchmark 
\gls{dataset}s are another special case where the user group is large and 
composed of recognized domain experts \cite{10.1145/3287560.3287596}. 

\subsection{Data} 
\label{sec_kr4_data} 

{\bf Datasheets for Datasets.} The authors of \cite{DatasheetData2021} propose 
a documentation principle for datasets, similar to product data sheets. 
In particular, each dataset should be accompanied by a data sheet that 
describes the collection process and intended use. This helps to ensure 
that biases and limitations are documented.

{\bf Data Augmentation for Simulatability.} To ensure simulatability of the 
hypothesis $\learnthypothesis \in \hypospace$ learnt by \gls{erm} \eqref{equ_def_emp_risk_min} 
we can include pseudo-labeled \gls{datapoint}s $\widetilde{\datapoint}$ in the \gls{trainset} $\dataset$. 
Such a pseudo-labeled \gls{datapoint} $\widetilde{\datapoint}=\pair{\featurevec}{\user(\featurevec)}$ 
is obtained by having a user provide a label $\user(\featurevec)$ for a test \gls{datapoint} with 
features $\featurevec \in \testset$. The \gls{testset} can be obtained by collecting new raw data 
or by systematic modifications of \gls{datapoint}s in the original \gls{trainset}. For example, 
the modification can amount to constructing counterfactual examples by removing or changing 
important features \cite{hase-bansal-2020-evaluating}. 

\begin{figure}[htbp]
	\begin{center} 
\begin{tikzpicture}[scale = 1]
	\draw[->, very thick] (0,0.5) -- (7.7,0.5) node[below, xshift=-1cm] {$\feature$};       
	\draw[->, very thick] (0.5,0) -- (0.5,4.2) node[above] {$\truelabel$};   
	
	\draw[color=black, thick, dashed, domain = -0.5: 7.2, variable = \x]  plot ({\x},{\x*0.4 + 2.0}) ;     
	\node at (6.7,4.3) {$\hypothesis(\feature)$};    
	
	\coordinate (l1)   at (1.2, 2.48);
	\coordinate (l2) at (1.4, 2.56);
	\coordinate (l3)   at (1.7,  2.68);
	
	\coordinate (l4)   at (2.2, 2.2*0.4+2.0);
	\coordinate (l5) at (2.4, 2.4*0.4+2.0);
	\coordinate (l6)   at (2.7,  2.7*0.4+2.0);
	
	\coordinate (l7)   at (3.9,  3.9*0.4+2.0);
	\coordinate (l8) at (4.2, 4.2*0.4+2.0);
	\coordinate (l9)   at (4.5,  4.5*0.4+2.0);
	
	\coordinate (n1)   at (1.2, 1.75); 
	\coordinate (n2)   at (1.4, 1.85); 
	\coordinate (n3)   at (1.7,  1.9); 
	
	\coordinate (n4)   at (2.2, 3.9); 
	\coordinate (n5)   at (2.4, 3.75); 
	\coordinate (n6)   at (2.7,  3.7); 
	
	\coordinate (n7)   at (5.7, 2.6);
	\coordinate (n8)   at (5.9, 2.6);
	\coordinate (n9)   at (6.2,  2.6);
	
	\node at (n1)  [circle,draw,fill=blue,minimum size=6pt,scale=0.6, name=c1] {};
	\node at (n2)  [circle,draw,fill=blue,minimum size=6pt, scale=0.6, name=c2] {};
	\node at (n3)  [circle,draw,fill=blue,minimum size=6pt,scale=0.6,  name=c3] {};
	\node at (n4)  [circle,draw,fill=blue,minimum size=12pt, scale=0.6, name=c4] {};  
	\node at (n5)  [circle,draw,fill=blue,minimum size=12pt,scale=0.6,  name=c5] {};
	\node at (n6)  [circle,draw,fill=blue,minimum size=12pt, scale=0.6, name=c6] {};  
	\node at (n7)  [circle,draw,fill=red,minimum size=12pt,scale=0.6,  name=c7] {};
	\node at (n8)  [circle,draw,fill=red,minimum size=12pt, scale=0.6, name=c8] {};
	\node at (n9)  [circle,draw,fill=red,minimum size=12pt, scale=0.6, name=c9] {};
	
	\draw[<->, color=blue, thick] (1.2, 2.48) -- (1.2, 1.75);  
	\draw[<->, color=blue, thick] (1.4, 2.56) -- (1.4, 1.85);  
	\draw[<->, color=blue, thick] (1.7, 2.68) -- (1.7, 1.9);  
	\draw[<->, color=blue, thick] (2.2, 2.2*0.4+2.0) -- (2.2, 3.9);  
	\draw[<->, color=blue, thick] (2.4, 2.4*0.4+2.0) -- (2.4, 3.75);  
	\draw[<->, color=blue, thick] (2.7, 2.7*0.4+2.0) -- (2.7, 3.7);  
	\draw[<->, color=red, thick] (5.7, 5.7*0.4+2.0) -- (5.7, 2.6);  
	\draw[<->, color=red, thick] (5.9, 5.9*0.4+2.0) -- (5.9, 2.6);  
	\draw[<->, color=red, thick] (6.2, 6.2*0.4+2.0) -- (6.2, 2.6);  
	
	\draw[fill=blue] (4.2, 1.7)  circle (0.1cm) node [black,xshift=1.3cm] {\gls{trainset} $\dataset$};
	\draw[fill=red] (4.2, 1.2)  circle (0.1cm) node [black,xshift=1.9cm] {pseudo-labeled test set};
	
\end{tikzpicture}

		\caption{We can improve simulatability (or subjective explainability) of \gls{erm} by 
			augmenting the \gls{trainset} with pseudo-labeled \gls{datapoint}s. These are obtained 
			from having the user predict labels of \gls{datapoint}s in a \gls{testset}.  \label{fig_aug_simulatability} }
	\end{center}
\end{figure}

\subsection{Model} 
\label{sec_kr4_model} 

{\bf Model Cards.} Similar to datasheets for \gls{dataset}s, model cards provide 
\gls{transparency} about the performance of trained \gls{model}s across 
different demographic groups \cite{10.1145/3287560.3287596}. This helps 
to identify fairness-related issues in \gls{erm}-based methods.

{\bf ``Simple'' Models.} One way to ensure \gls{explainability} of \gls{erm} \eqref{equ_def_emp_risk} 
is to choose a \gls{model} $\hypospace$ that only contains hypothesis 
maps that are simulatable. However, this choice must take into account the 
specific user (knowledge) and the construction of test set over which we 
compare user predictions with model predictions. For example, a \gls{linmodel} 
might be considered explainable only if the underlying \gls{featurespace} has 
small dimension and for users that have basic understanding of linear functions. 

{\bf Constructing Explanations.} Methods for explainable AI not only differ in how they measure 
\gls{explainability} but also in the form of \gls{explanation}s \cite{JunXML2020}. One widely 
used form of \gls{explanation} is to list the most important \gls{feature}s of a \gls{datapoint} \cite{Ribeiro2016}. 
Another form of \gls{explanation} is to use heat-maps that indicate the relative importance of 
image pixels \cite{GradCamPaper}. Case-based reasoning uses specific \gls{datapoint}s 
from the \gls{trainset} as an \gls{explanation} \cite{Aamodt:1994aa}. In general, an \gls{explanation} 
is some function $\explanation(\featurevec)$ of the features of a \gls{datapoint}. This \gls{explanation} 
is delivered along with the \gls{prediction} to the user. Formally, this corresponds to 
using a \gls{hypothesis} map $\hypothesis$ with structured output $\hypothesis(\featurevec) = \big( \explanation(\featurevec), \predictedlabel \big)^{T}$. 

\begin{figure} 
	\begin{center}
\begin{tikzpicture}
	\fill[white] (0, 0) rectangle (5, -4); 
	

	\node[draw, rounded corners, fill=white, anchor=north west, text width=3.0cm, align=left] at (0.5, -0.8) { 
		The lecture was bad.
	};
	
	\draw[->, thick] (4, -1) -- (4.5, -1) node[midway, above] {$\hypothesis(\feature)$};
	
	\node[anchor=north west] at (4.8, 0) {\gls{prediction} $\predictedlabel\!=\!\mbox{``negative''}$
	};
	 \node[anchor=east] at (6.0, -1.5) {$\explanation(\featurevec) = $};
	\node[anchor=north west, draw, rounded corners, fill=white, text width=2.3cm, align=left] at (6.0, -1) {
	The lecture was \tikz[baseline]{\node[fill=yellow, anchor=base] {bad};}. 
	};
	\vspace*{-2cm}
\end{tikzpicture}
\end{center}
	\vspace*{-2cm}
\caption{We can ensure \gls{explainability} of \gls{erm}-based methods 
	by augmenting the \gls{prediction} delivered by the trained \gls{model} 
	with some \gls{explanation}. \label{fig_model_explanation_along_prediction}}
\end{figure}

\subsection{Loss}
\label{sec_kr4_loss} 

The augmentation of the \gls{trainset} in \gls{erm} with pseudo-labeled examples 
(see Section \ref{sec_kr4_data}) is equivalent to including the penalty term 
$\regularizer{\hypothesis} = \sum_{\featurevec \in \testset} \lossfunc{\pair{\featurevec}{\user(\featurevec)}}{\hypothesis}$ 
in the \gls{lossfunc} used by \eqref{equ_def_emp_risk} (see Figure \ref{fig_equiv_dataaug_penal}). 
Instead of using an explicit \gls{testset}, the authors of \cite{Zhang:2024aa} use a simple probabilistic 
model for the \gls{datapoint}s and user signal $p(\featurevec,\user)$ which allows 
to construct a penalty term via the expected loss 
\begin{equation}
	\nonumber
	\regularizer{\hypothesis} = \expect \bigg\{  \lossfunc{\pair{\featurevec}{\user(\featurevec)}}{\hypothesis} \bigg\}.
\end{equation}

\section{KR5 - Diversity, Non-discrimination and Fairness} 
\label{sec_kr5}

``\emph{...we must enable inclusion and diversity throughout the entire AI 
	system’s life cycle...this also entails ensuring equal access through inclusive 
	design processes as well as equal treatment.}''\cite[p.18]{HLEGTrustworhtyAI}.

Consider an AI application that uses \gls{datapoint}s representing humans. Each \gls{datapoint} is 
characterized by \gls{feature}s $\featurevec$ and a \gls{sensattr} $\sensattr$. The \gls{sensattr} 
typically depends on the raw \gls{feature}s of a \gls{datapoint}, $\sensattr = \sensattr(\featurevec)$ with 
some map $\sensattr(\cdot)$. Examples for a \gls{sensattr} $\sensattr$ include ethnicity, 
age, gender or religion.\footnote{The definition of the \gls{sensattr} $\sensattr$ is a design 
	choice that varies by application. For instance, religion might be a \gls{sensattr} on a job 
	application platform, but it could be a relevant \gls{feature} in a diet planning app.}

{\bf Individual Fairness (Disparate Treatment).} Roughly speaking, a fair \gls{erm}-based method 
should learn a $\learnthypothesis \in \hypospace$ that does not put inappropriate weight on the 
\gls{sensattr}. To makes this fairness notion precise, we need a measure $d\big(\featurevec,\featurevec'\big)$ 
for the similarity between \gls{datapoint}s, with features $\featurevec, \featurevec'$, that maximally 
ignores their \gls{sensattr}s \cite{dwork2012fairness,barocas2019fairml}. A fair \gls{classifier} 
should deliver the same \gls{prediction}s for sufficiently similar \gls{datapoint}s, 
\begin{align}
		\label{equ_def_fair_insensitive_attribute}
	\widehat{\hypothesis}\big( \featurevec \big)  & =	\widehat{\hypothesis}\big( \featurevec' \big)  \nonumber \\ 
	& \mbox{ whenever } d\big(\featurevec,\featurevec'\big) \mbox{ is sufficiently small.} 
\end{align}
Here, $d\big(\featurevec,\featurevec'\big)$ denotes a quantitive measure for the similarity 
between two \gls{datapoint}s with \gls{feature}s $\featurevec, \featurevec'$, respectively. 
The fairness requirement \eqref{equ_def_fair_insensitive_attribute} seems natural 
in order to prevent \emph{disparate treatment} \cite{Pessach2022}.

{\bf Example: Job Platform.} Consider a job platform that uses \gls{erm} to learn a hypothesis $\learnthypothesis$
 for predicting if a given user is suitable for a specific job opening. Each user is characterized by 
 \gls{feature}s $\featurevec=\big(\feature_{1},\ldots,\feature_{\nrfeatures}\big)$ with its first entry $\feature_{1}$ being the age of the user. Thus, 
 the \gls{sensattr} is $\sensattr=\feature_{1}$. Fairness might require  that the prediction $\learnthypothesis(\featurevec)$ 
 does not depend at all on the age of the user \cite{TutorGroup2023}. We could ensure this by using 
 a classifier satisfying \eqref{equ_def_fair_insensitive_attribute} with a metric $d\big(\featurevec,\featurevec'\big)$ that does 
 not depend on $\feature_{1}$ . However, the requirement \eqref{equ_def_fair_insensitive_attribute} is 
insufficient when the \gls{sensattr} $\sensattr = \feature_{1}$ can be inferred (predicted) 
from the values of the remaining \gls{feature}s $\feature_{2},\ldots,\feature_{\nrfeatures}$ \cite{hardt2016equality,10.1145/1401890.1401959}. 

ML literature has proposed and studied a variety of quantitative measures for the fairness 
of a trained model $\learnthypothesis \in \hypospace$. In what follows we briefly survey 
some of these measures in the context of binary classification where the learn hypothesis 
is used to deliver a predicted label $\predictedlabel \in \{ 0,1\}$. 

{\bf Group Fairness (Disparate Impact).} Besides the individual fairness constraint 
\eqref{equ_def_fair_insensitive_attribute}, another flavour of fairness is to require 
a trained model to have similar performance across sub-populations \cite{FairERM2018,berk2017fairness,kleinberg2017tradeoffs,chouldechova2017fair,hardt2016equality,dwork2012fairness}. 
For example, we might require identical conditional risk for subsets of \gls{datapoint}s 
with \gls{sensattr} value $\sensattr^{(1)}$ and $\sensattr^{(2)}$, respectively, 
\begin{equation}
	\label{equ_def_fair_loss_cond_same}
	\expect \big\{ \lossfunc{\pair{\featurevec}{\truelabel}}{\widehat{\hypothesis}} \!\big|\! \sensattr\!=\!\sensattr^{(1)}  \big\}\!=\!  
		\expect \big\{ \lossfunc{\pair{\featurevec}{\truelabel}}{\widehat{\hypothesis}} \!\big|\! \sensattr\!=\!\sensattr^{(2)}  \big\}. 
\end{equation}  
Imposing \eqref{equ_def_fair_loss_cond_same} requires the learnt \gls{hypothesis} 
to have the same performance (expected \gls{loss}) over sub-populations of \gls{datapoint}s 
that have a common \gls{sensattr} $\sensattr$ (e.g., ``males'' and ``females''). The 
fairness requirement \eqref{equ_def_fair_loss_cond_same} is closely related to the notion 
of \emph{disparate impact} \cite{Pessach2022}.

\begin{figure}[htbp]
	\begin{center} 
		\begin{tikzpicture}[scale = 1]
			\draw[->, very thick] (0,0.5) -- (7.7,0.5) node[below, xshift=-1cm] {\gls{feature}s $\feature$};       
			\draw[->, very thick] (0.5,0) -- (0.5,4.2) node[above] {credit score $\truelabel$};   
			
			\draw[color=black, thick, dashed, domain = -0.5: 5.2, variable = \x]  plot ({\x},{\x*0.4 + 2.0}) ;     
			\node at (5.7,4.1) {$\hypothesis(\feature)$};    
			
			\coordinate (l1)   at (1.2, 2.48);
			\coordinate (l2) at (1.4, 2.56);
			\coordinate (l3)   at (1.7,  2.68);
			
			\coordinate (l4)   at (2.2, 2.2*0.4+2.0);
			\coordinate (l5) at (2.4, 2.4*0.4+2.0);
			\coordinate (l6)   at (2.7,  2.7*0.4+2.0);
			
			\coordinate (l7)   at (3.9,  3.9*0.4+2.0);
			\coordinate (l8) at (4.2, 4.2*0.4+2.0);
			\coordinate (l9)   at (4.5,  4.5*0.4+2.0);
			
			\coordinate (n1)   at (1.2, 1.8);
			\coordinate (n2) at (1.4, 1.8);
			\coordinate (n3)   at (1.7,  1.8);
			
			\coordinate (n4)   at (2.2, 3.8);
			\coordinate (n5) at (2.4, 3.8);
			\coordinate (n6)   at (2.7,  3.8);

			\coordinate (n7)   at (3.9, 2.6);
			\coordinate (n8) at (4.2, 2.6);
			\coordinate (n9)   at (4.5,  2.6);
			
			\node at (n1)  [circle,draw,fill=red,minimum size=6pt,scale=0.6, name=c1] {};
			\node at (n2)  [circle,draw,fill=blue,minimum size=6pt, scale=0.6, name=c2] {};
			\node at (n3)  [circle,draw,fill=red,minimum size=6pt,scale=0.6,  name=c3] {};
			\node at (n4)  [circle,draw,fill=red,minimum size=12pt, scale=0.6, name=c4] {};  
			\node at (n5)  [circle,draw,fill=blue,minimum size=12pt,scale=0.6,  name=c5] {};
			\node at (n6)  [circle,draw,fill=red,minimum size=12pt, scale=0.6, name=c6] {};  
			\node at (n7)  [circle,draw,fill=red,minimum size=12pt,scale=0.6,  name=c7] {};
			\node at (n8)  [circle,draw,fill=blue,minimum size=12pt, scale=0.6, name=c8] {};
			\node at (n9)  [circle,draw,fill=red,minimum size=12pt, scale=0.6, name=c9] {};
			

			\draw[<->, color=red, thick] (l1) -- (c1);  
			\draw[<->, color=blue, thick] (l2) -- (c2);  
			\draw[<->, color=red, thick] (l3) -- (c3);  
			\draw[<->, color=red, thick] (l4) -- (c4);  
			\draw[<->, color=blue, thick] (l5) -- (c5);  
			\draw[<->, color=red, thick] (l6) -- (c6);  .
			\draw[<->, color=red, thick] (l7) -- (c7);  
			\draw[<->, color=blue, thick] (l8) -- (c8);  
			\draw[<->, color=red, thick] (l9) -- (c9);  
			
			\draw[fill=blue] (5.2, 1.7)  circle (0.1cm) node [black,xshift=1.4cm] {original \gls{dataset}};
			\draw[fill=red] (5.2, 1.2)  circle (0.1cm) node [black,xshift=1.5cm] {modified gender};
			
		\end{tikzpicture}
		\caption{We can improve fairness of a ML method by augmenting the \gls{trainset} 
			using perturbations of an irrelevant \gls{feature}. For example, in a credit 
			scoring application, we might change the gender of a person while keeping 
			the remaining \gls{feature}s fixed. \label{fig_fairness_aug} }
	\end{center}
\end{figure} 

\subsection{Data}
\label{sec_kr5_data} 

The \gls{trainset} $\dataset$ used in \gls{erm} should be carefully selected to not enforce 
existing discrimination. In a health-care application, there might be significantly more 
training data for patients of a specific gender, resulting in models that perform best for 
that specific gender at the cost of worse performance for the minority \cite[Sec. 3.3.]{NISTDiffPriv2023}. 

Fairness is also important for ML methods used to determine credit score and, in turn, if a loan 
should be granted or not \cite{KOZODOI20221083}. Here, we must ensure that ML methods do 
not discriminate customers based on ethnicity or race. To this end, we could augment \gls{datapoint}s 
via modifying any \gls{feature}s that mainly reflect the ethnicity or race of a customer (see Figure \ref{fig_fairness_aug}). 

{\bf Data augmentation} for enforcing fairness of \gls{erm} is also studied in \cite{10.1145/3531146.3534644}. 
The data augmentation strategy in this paper involves replacing individuals in video 
frames with new individuals while maintaining the original motion \cite{MarkRCNN2017,HRNet2021}. 
The two-step process involves (i) tracking and segmenting the target person in the 
video and (ii) replacing the person with another individual by transforming key-points and poses..

{\bf Fair Data Collection.} The fairness of \gls{erm}-based methods includes  
the data collection process \cite{Longpre2024Data}. The \gls{datapoint}s used 
in the \gls{trainset} $\dataset$ of \gls{erm} \eqref{equ_def_emp_risk_min} must be 
gathered in a way that is aligns with fundamental rights and regulations like \gls{gdpr} \cite{GDPR2016,eu_charter_2012}. 
The data collection should be transparent and representative, avoiding biased sampling 
and improper consent procedures \cite{GoodRx2023ConsumerReports,Criteo2023EDPB}.

\subsection{Model} 
\label{sec_kr5_model} 

We can ensure fairness of the learnt hypothesis $\learnthypothesis \in \hypospace$ by using a 
model $\hypospace$ that only includes hypothesis maps satisfying fairness constraints such 
as (variations) of \eqref{equ_def_fair_insensitive_attribute}. One example for such a constraint is 
to require each hypothesis $\hypothesis$ to be Lipschitz continuous \cite{dwork2012fairness}. 
The idea is to require $\learnthypothesis$ to deliver similar predictions for \gls{datapoint}s 
that are similar in a non-discriminatory sense. A key challenge for the practical 
use of this requirement is to find a useful choice for the metric underlying the 
Lipschitz condition \cite{dwork2012fairness}.

\subsection{Loss} 
\label{sec_kr5_loss} 

{\bf Fairness via Regularization.} Fairness constraints of the form \eqref{equ_def_fair_loss_cond_same} 
can be included in the \gls{loss} of \gls{erm}. By Lagrangian duality \cite[Ch. 5]{BoydConvexBook}, the 
constraints can be translated into a penalty term that is added ot the \gls{erm} \gls{objfunc} \cite{JMLR:v20:18-262,agarwal2018reductions,FairERM2018,hardt2016equality}. 
Adding such a fairness penalty term can be interpreted as a form of \gls{regularization} 
(see Section \ref{sec_erm_AI_engine} and Figure \ref{fig_equiv_dataaug_penal}). 

{\bf Fairness via Sample Weighting.} Instead of adding a penalty term to the \gls{lossfunc} in \gls{erm}, 
we can also ensure fairness by sample weighting \cite{Kamiran:2012aa}. The idea is 
to scale the \gls{loss} incurred on a \gls{datapoint} based on the relative frequency of 
its \gls{sensattr} in the \gls{trainset} $\dataset$. Magnifying the \gls{loss} incurred for 
\gls{datapoint}s from a minority ensures that under-represented groups have a larger 
influence on the solution of \eqref{equ_def_emp_risk_min}. 

\section{KR6 - Societal and Environmental Well-Being} ``\emph{...Sustainability and
	ecological responsibility of AI systems should be encouraged, and research should be 
	fostered into AI solutions addressing areas of global concern, such as for instance the 
	Sustainable Development Goals.}''\cite[p.19]{HLEGTrustworhtyAI}.
	
So far, we discussed KRs that focused on the effect for \gls{erm}-based methods on 
individual users. In contrast, KR6 key requirement revolves around the wider impact 
of an \gls{erm}-based method on the level of societies and natural environments. 

{\bf Society and Democracy.} Design choices for \gls{erm} should also consider the effect of 
(predictions delivered by) a trained model $\learnthypothesis \in \hypospace$ on society 
at large. The predictions $\learnthypothesis(\featurevec)$ could not only harm the mental 
health of individual users but also affect core democratic processes such as policy-making 
or elections. As a case in point, social media apps train personalized \gls{model}s $\learnthypothesis$ 
to recommend (or select) content delivered to its users. The resulting tailored filtering of content 
can boost polarization and, in the extreme case, social unrest \cite{Goncalves-Sa:2024aa}. 

{\bf Environment.} \gls{erm}-based AI systems need to solve the optimization problem \eqref{equ_def_emp_risk} 
using some computational methods. The implementation of these methods in physical 
hardware requires energy which is typically provided in the form of electricity \cite{CostsLLM}.  
Given the increasing energy requirement by AI systems, it is crucial to use environmental-friendly 
means of energy production \cite{GreenCom}. Design choices for \gls{erm} should minimize 
the energy demand, as well as demand for cooling water  \cite{li2023makingaithirstyuncovering}, 
of the resulting AI system. These demands not only depend on the computational work required 
to solve \gls{erm} \eqref{equ_def_emp_risk_min} but also on the data collection strategies \cite{Fredriksson2020}.

\section{KR7 - Accountability} ``\emph{...mechanisms be put in place to ensure responsibility 
	and accountability for AI systems and their outcomes, both before and after their development, 
	deployment and use.}'' \cite[p. 19]{HLEGTrustworhtyAI}. 
	
{\bf Policy and Governance Approaches.} Organizations such as the OECD have been 
working on governance structures for \gls{erm}-based AI systems. This includes 
formalizing auditing procedures and ensuring that developers are held to both 
ethical standards and legal requirements \cite{OECDAcountReport}. 

{\bf Frameworks for Answerability.} AI developers and operators must be able to 
justify their actions and decisions. This involves both \gls{transparency} (see Section \ref{sec_kr4}) 
and oversight (see Section \ref{sec_kr1}) mechanisms which are especially important 
in high-stakes domains \cite{Novelli:2024aa}. The justification of \gls{erm} design choices 
also requires a solid understanding of the inherent trade-offs between design criteria such 
as \gls{explainability} and \gls{acc} \cite{Jaotombo:2023aa,Zhang:2024aa}.

{\bf Regular Audits and Third-Party Reviews.}
Periodic reviews of AI systems by independent auditors help ensure and validate  
accountability. To this end, independent external teams (``red teams'') should 
stress-test the \gls{erm}-based system for vulnerabilities and biases that might 
undermine accountability \cite{OECDAcountReport,2022arXiv220203286P}.

\bibliographystyle{IEEEBib}
\bibliography{/Users/junga1/adictml/Literature}
 
\end{document}